\documentclass[preprint,11pt]{article}
\makeatletter
\def\blfootnote{\gdef\@thefnmark{}\@footnotetext}
\makeatother
\usepackage{fullpage}
\ifdefined\usebigfont

\usepackage{times}
\usepackage[fontsize=13pt]{scrextend}
\usepackage[left=1.56in,right=1.56in,top=1.74in,bottom=1.74in]{geometry}
\else
\fi

\usepackage{amssymb,amsfonts,amsmath,amsthm,amscd,dsfont,mathrsfs}
\usepackage{graphicx,float,psfrag,epsfig,amssymb}
\usepackage[usenames,dvipsnames,svgnames,table]{xcolor}
\definecolor{darkgreen}{rgb}{0.0,0,0.9}
\RequirePackage{algorithm}
\usepackage[pagebackref,letterpaper=true,colorlinks=true,pdfpagemode=none,citecolor=OliveGreen,linkcolor=BrickRed,urlcolor=BrickRed,pdfstartview=FitH]{hyperref}
\usepackage{wrapfig}
\usepackage{relsize}
\usepackage{color}
\usepackage{pict2e}
\usepackage[tight]{subfigure}
\usepackage{caption}
\usepackage{nameref}
\usepackage{makecell}
\usepackage[font={small}]{caption} 
\usepackage{float}
\usepackage{bbm}
\usepackage{tikz}
\usepackage{multirow}
\usepackage{adjustbox}
\usepackage{csvsimple,longtable,booktabs}
\usepackage{graphicx}
\usepackage{cases}
\usepackage{mathtools}
\usepackage{dcolumn,booktabs}
\newcolumntype{d}[1]{D{.}{.}{#1}}

\usepackage{enumitem}
\usepackage{multirow}
\usepackage{lscape}
\usepackage{gensymb}
\usepackage{natbib}

\let\chapter\section
\usepackage{hyperref}

\usepackage{amssymb}
\usepackage{mathtools}
\usepackage{accents}

\usepackage{amsmath}
\usepackage{amsthm}
\usepackage{graphicx,psfrag,epsf}
\usepackage{enumerate}
\usepackage{natbib}
\usepackage[noend]{algorithmic} 

\usepackage{graphicx}
\usepackage{comment}

\newcommand\inner[2]{\langle #1, #2 \rangle}
\usepackage[english]{babel}

\usepackage{comment}

\newcommand{\adel}[1]{\textcolor{red}{~Adel:~#1}}

\usepackage[capitalize,noabbrev]{cleveref}

\theoremstyle{plain}
\newtheorem{theorem}{Theorem}[section]
\newtheorem{proposition}[theorem]{Proposition}
\newtheorem{lemma}[theorem]{Lemma}
\newtheorem{corollary}[theorem]{Corollary}
\theoremstyle{definition}

\theoremstyle{remark}

\usepackage{bm}
\newcommand{\ex}{\mathbb{E}}
\newcommand{\thetawa}{\theta}
\newcommand{\repardist}{\Phi}
\newcommand{\repardens}{\phi}
\newcommand{\noisedist}{\Phi}

\newcommand{\Thetawa}{\Theta}
\newcommand{\thetaap}{\lambda}

\newcommand{\pl}{\omega}
\newcommand{\KL}{{\sf KL}}
\newcommand*\samethanks[1][\value{footnote}]{\footnotemark[#1]}

\title{Structured Dynamic Pricing: Optimal Regret in a Global Shrinkage Model}

\author{%
Rashmi Ranjan Bhuyan\thanks{Data Science and Operations Department, Marshall School of Business, University of Southern California} 
\quad\,
Adel Javanmard\samethanks[1] 
\quad\,
Sungchul Kim\thanks{Adobe Research}\\
Gourab Mukherjee\samethanks[1] 
\quad\,
Ryan A. Rossi\samethanks[2]
\quad\,
Tong Yu\samethanks[2]
\quad\,
Handong Zhao\samethanks[2]
}

\begin{document}

\maketitle

\blfootnote{A. Javanmard is partially supported by the Sloan Research Fellowship in mathematics, Adobe Data Science
Faculty Research Awards and the NSF CAREER Award DMS-1844481. G. Mukherjee is partially supported by an Adobe Data Science
Faculty Research Award.}

\begin{abstract}
 We consider dynamic pricing strategies in a streamed longitudinal data set-up where the objective is to maximize, over time, the cumulative profit across a large number of customer segments. We consider a dynamic model with the consumers' preferences as well as price sensitivity varying over time. Building on the well-known finding that consumers sharing similar characteristics act in similar ways, we consider a global shrinkage structure, which assumes that the consumers' preferences across the different segments can be well approximated by a spatial autoregressive (SAR) model. 
In such a streamed longitudinal set-up, we measure the performance of a dynamic pricing policy via regret, which is the expected revenue loss compared to a clairvoyant that knows the sequence of model parameters in advance. We propose a pricing policy based on penalized stochastic gradient descent (PSGD) and explicitly characterize its regret as functions of time, the temporal variability
in the model parameters as well as the strength of the auto-correlation network structure spanning the varied customer segments. Our regret analysis results not only demonstrate asymptotic optimality of the proposed policy but also show that for policy planning it is essential to incorporate available structural information as policies based on unshrunken models are highly sub-optimal in the aforementioned set-up. We conduct simulation experiments across a wide range of regimes as well as real-world networks based studies and report encouraging performance for our proposed method. 
\end{abstract}
\section{Introduction}\label{sec:intro}
Due to the ubiquitous reach of digital marketing, dynamic pricing settings are extensively studied by firms that sell a significant fraction of their inventories over online marketplaces and through digital advertisements (see \citealp{cohen2016feature,keskin2016chasing,ban2017personalized,bimpikis2019spatial,javanmard2017perishability,leme2018contextual,javanmard2019dynamic} and the references therein). As such it is a vibrant topic of research in online machine-learning \citep{zhou2019learning,cesa2015regret}, operations research \citep{golrezaei2017dynamic,cheung2017dynamic}, information \citep{cui2021informational}, marketing \citep{schwartz2017customer,choi2020online} and management sciences \citep{farias2010dynamic,broder2012dynamic,den2013simultaneously}. For fuller references see Sec~1.3.  

In this work, we study the problem of a firm selling a product to customers who arrive over time. The firm has the opportunity to set different prices not only over time $t$ but also for different customer segments $l=1,\ldots,L$. We consider setting the prices across these customer segments in a dynamic manner such that the expected cumulative revenue, aggregated over the customer segments as well as time is maximized. As a motivational example, consider the digital marketing problem \citep{liu2019decade} where an advertisement of a product priced at $p_{lt}$ is shown to $n_{lt}$ 
customers in segment $l$ at time $t$. Let $y_{ltk}, k=1,\ldots,n_{lt}$ denote the binary variables corresponding to conversion based on the advertisement, i.e.,  $y_{ltk}=1$ if the $k$-th advertisement in the $l$-th segment at time $t$ led to a purchase, and $y_{ltk}=0$ otherwise. Often in these problems, the firm also has the opportunity to access other covariates $\bm{x}_{lt}$ such as demographic information for the customer segment $l$ at time $t$. 

As time $t$ progresses, the goal is to explore and set prices $p_{lt}$ optimally based on the current covariates $x_{lt}$ as well as on the previous customer responses $\{y_{lsk}:1\leq s < t\}$ and their associated prices and covariate information. The goal is to optimize the cumulative revenue

\begin{align}\label{eq.1}
\sum_{t=1}^T\sum_{l=1}^L y_{lt}\, p_{lt} \text{ where, } y_{lt}=\sum_{k=1}^{n_l}y_{ltk}.
\end{align}
\subsection{Streamed Longitudinal Probit Set-up}

 Demand heterogeneity \citep{bimpikis2019spatial,chintagunta2002market} is traditionally tackled by segmenting consumers who have similar purchasing propensity as well as similar responses to price changes. Though truly homogeneous segments of consumers do not exist, the approximation provides a reasonable interface to design differential pricing strategies that optimally target each customer segments. Modern online trading platforms, marketplaces and lead generation systems, facilitate implementing price differential strategies across a wide range of segments. Often advertisers have access to the geographical location of the consumer and these segments based on zip-codes of the consumers \citep{train2009discrete}. Another popular choice is segmenting customers based on the different marketing channels by which they were approached \citep{berman2018planning}.   
In accordance with these modern applications, we consider the number of segments $L$ to be large. In the existing literature \citep{javanmard2017perishability}, the overall revenue in \eqref{eq.1} is optimized for probability models based on rational choice theory, which assumes that consumers are rational and make choices that maximize their utility.

Let the utility function $U_{ltk}$ for the $k$th customer in the $l$th segment at time $t$ be given by the following additive model:

\begin{align}\label{eq:2}
U_{ltk}=\alpha_{lt}+\beta_{t}\,p_{lt}+\bm{x}_{lt}' \bm{\mu}_{t} + \sigma Z_{ltk}
\end{align}
where $k=1,\ldots,n_{lt}$; $\alpha$s 
are the preferences of the customers that vary across both time and segments; $\beta$s are the price-sensitivities of the customer. Vectors $\bm{\mu}$ are coefficients corresponding to the non-priced covariates and vary over time but invariant across segments. $Z$s are independent and identically from Gaussian distribution with mean $0$ and has variance $1$. Based on rational choice theory, if $U_{ltk}>0$ then a sale occurs, i.e., $Y_{ltk}=1$; else, $Y_{ltk}=0$. Let $Y_{lt}=\sum_{k=1}^{n_l}Y_{ltk}$ be the count of sales for segment $l$ at time $t$. Then,
$$Y_{lt}\sim \text{Binomial}(n_{lt},q_{lt})\,,$$
where $q_{lt}=\Phi(\sigma^{-1}(\alpha_{lt}+\beta_{t} p_{lt}+\bm{x}_{lt}' \bm{\mu}_{t}))$ and  
$\Phi$ is the cumulative distribution function of standard Gaussian distribution. Note, that $q_{lt}$ is only a function of the model parameters but also depends on the price. We use capital letters for random variables, small letters for the values a random variable takes, and boldface letters for vectors and matrices.

The joint log-likelihood across all segments at time $t$ is given by 

\begin{align}\label{eq:2.1}
\ell_t(\bm{\thetaap})=\sum_{l=1}^L y_{lt} \log q_{lt} + (n_{lt}-y_{lt})\log (1-q_{lt}),
\end{align}
where $\bm{\thetaap}=\{\bm{\thetaap}_t:=(\bm{\alpha}_t,\beta_t,\bm{\mu}_t): t=1,\ldots,T\}$ and $\bm{\alpha}_t=(\alpha_{1t},\ldots,\alpha_{Lt})$. 
The expected revenue from segment $l$ subjected to price $p_{lt}$ at time $t$ is,
\begin{align*}
    \text{Rev}(\bm{\thetaap},l,t,p_{lt})&= p_{lt}\,\ex_{\bm{\thetaap}}(Y_{lt}) = n_{lt}\, p_{lt}\, q_{lt}\\
    &=n_{lt}p_{lt}\Phi\big(\sigma^{-1}(\bm{\alpha}_{lt}+\beta_{t} p_{lt}+\bm{x}_{lt}'\bm{\mu}_{t})\big),
\end{align*}
and the goal is to maximize the cumulative revenue
$$\text{Rev}(\bm{\thetaap},\bm{p})=\sum_{l=1}^L\sum_{t=1}^T\text{Rev}(\bm{\thetaap},l,t,p_{lt})$$
over the prices $\bm{p}=\{p_{lt}: 1\leq l \leq L, 1\leq t \leq T\}$ that the firm can set. 
Conditioned on the parameters $\bm{\thetaap}$, maximizing $\text{Rev}(\bm{\thetaap},\bm{p})$ decouples into separate maximization of the revenue of each segment at each time point. The first-order condition for optimal price $p_{lt}^*$ conditional on the set of parameters is  
\begin{align}\label{eq:2.2}
   p_{lt}^*  = -\sigma \,\beta_{t}^{-1} \frac {\Phi(\sigma^{-1}(\alpha_{lt}+{\beta}_{t}\,p_{lt}^*+\bm{x}_{lt}' \bm{\mu}_{t}))}{ \phi(\sigma^{-1}(\alpha_{lt}+{\beta}_{t}\,p_{lt}^*+\bm{x}_{lt}' \bm{\mu}_{t}))}~.
\end{align}
As $p_{lt}^*$ depends on the unknown model parameters $\bm{\thetaap}_t$, we call this the oracle price and $\text{Rev}(\bm{\thetaap},\bm{p}^*)$ imposes the highest theoretically achievable upper bound on the revenue. For any other pricing policy $\bm{p}$ we define its regret over the oracle strategy as: 
$$\mathcal{R}(\bm{\thetaap},\bm{p}) = \sum_{l=1}^L\sum_{t=1}^T \mathcal{R}_{lt}(\bm{\thetaap},\bm{p}), \text{ where }$$ $$\mathcal{R}_{lt}(\bm{\thetaap},\bm{p})=\text{Rev}(\bm{\thetaap},l,t,p^*_{lt})-\text{Rev}(\bm{\thetaap},l,t,\hat p_{lt}).$$

\subsection{SAR based global Shrinkage Structures}

We impose the following regularity condition on the temporal changes in the price sensitivity and the covariate effects:

\begin{align}\label{eq:3}
\sum_{t=1}^{T-1}|\beta_{t+1}-\beta_t|\leq C_\beta^* \text{ and } \sum_{t=1}^{T-1}\|\bm{\mu}_{t+1}-\bm{\mu}_t\|_2\leq C_\mu^*.
\end{align}
Unlike the price sensitivity, the preference coefficients $\bm{\alpha}_t$ however greatly depends on the state of the current inventory and can highly fluctuate over time. However, it is well known that people who are close to each other in some networks often reflect highly corrected preferences \citep{bradlow2005spatial, ma2015latent}. Spatial models provide a natural way to model this correlation between different units of analysis based on their contiguity in a network \citep{banerjee2014hierarchical,gelfand2010handbook,lesage2004family}.
Geographic closeness is a proxy for many socio-demographic variables like income, education, wealth and property values, which are also related to consumer purchase behavior, and has been the primary 
focus of a large number of existing pricing models \citep{yang2003modeling,jank2005understanding,bimpikis2019spatial}. Networks based on non-geographic 
metrics can also capture preference similarities among customers \citep{karmakar2021}. Consider the following \textit{Spatially Autoregressive} (SAR) structure (see ch. 6 of \citealp{anselin2013spatial} and ch. 2 of \citealp{banerjee2014hierarchical}) on the $\alpha_{lt}$:
\begin{align}\label{eq:4}
 \alpha_{lt}= \rho_t \sum_{j=1}^L w_{lj} \alpha_{jt} + \tau \epsilon_{lt},
 \end{align}
where $w_{lj}\geq 0$, $\epsilon_{lt}$ are i.i.d $N(0,1)$ and $\tau >0$. In its most basic form, \eqref{eq:4} imposes a global hierarchical structure with the auto-correlation parameter $\rho_t$ regulating the level of global spillovers (and hence connectedness) among the units. Relation \eqref{eq:4} implies having the following hierarchical prior on the preference parameters: 
\begin{align}\label{eq:5}
\bm{\alpha}_t\sim N_L(\bm{0},\,\tau^2\, (\bm I-\rho_t \bm W)^{-2})\,.
\end{align}
We consider the network structure and its associated contiguity matrix $\bm W$ to be invariant over time. SAR models such as above have been very successful in assimilating spatial network information in real-world datasets \citep{manski1993identification,  anselin2013spatial}. In \cite{bramoulle2009identification}, SAR is used in modeling recreational services consumption by secondary school students, whereas \cite{hsieh2016social} used SAR to incorporate friendship networks of high school students in predicting their academic performances. In \cite{zhou2017estimating}, SAR is used to model user activity on social media regarding transportation services in China. We allow the auto-correlation parameter $\rho_t$ to vary over time while satisfying the regularity condition 
\begin{align}\label{eq:6}
\sum_{t=1}^{T-1}|\rho_{t+1}-\rho_t|\leq C_\rho^*.
\end{align}
\begin{align}\label{eq:7}
\mathcal{B}(\bm{\thetawa},\bm{p})=\ex_{\bm{\lambda}_{\alpha}}\{\mathcal{R}(\bm{\thetaap},\bm{p})\},
\end{align}
where the expectation is over the distribution of $\alpha_t$ governed by \eqref{eq:5}. Let $\Theta$ be a set of parameter $\bm{\thetawa}$ satisfying \eqref{eq:3} and \eqref{eq:6}. For this set of parameters, we consider developing dynamic pricing strategies $\bm{p}$ that minimize the Bayes regret in \eqref{eq:7}.  

\newcommand{\bigo}{\mathcal{O}}

\subsection{Our Contributions and Related Work}

We develop a Projected Stochastic Gradient Descent (PSGD) algorithm based on the logarithm of the marginal likelihood $\ell_t(\bm{\thetawa})=\log\{\ex_{\bm{\theta}_\alpha}\{\exp \ell_t(\bm{\thetaap})\}\}$ which is the convolution of the likelihood in \eqref{eq:2.1} with the prior in \eqref{eq:5}. We show that the proposed algorithm controls the Bayes regret at the order of $\bigo(\sqrt{T})$. We also show that for any data-driven pricing strategy the Bayes regret can not be of the lower order of $\bigo(\sqrt{T})$. Thus, as $T \to \infty$, the proposed algorithm is asymptotically rate-optimal. Our main result, Theorem~\ref{thm.1} is provided in Section 3. 

An important attribute of Theorem~\ref{thm.1} is that, we provide an explicit characterization of the Bayes regret of the proposed PSGD algorithm in terms of not only time $T$ but also as functions of the model parameters and the underlying heterogeneity (difference in the $n_{lt}$) in the data.  We show how the regret of the proposed algorithm depends on temporal variability in the model parameters as well as on the strength of correlation among the segments. Our upper-bound on the regret of the prescribed method (see~\eqref{eq:23}) depends on the spectral radius of the SAR structure in \eqref{eq:7}. It is sensitive to the magnitude of the autocorrelation parameter and greatly contracts as the correlation increases. 

In Corollary~\ref{cor.1}, we show that any unshrunken pricing policy that does not borrow strength across the customer segments is highly sub-optimal with respect to the proposed strategy. This is in accordance with classical statistical shrinkage theory results \citep{fourdrinier2018shrinkage} that are based on non-dynamic set-ups. To see the connections consider the penalized likelihood criterion:

\begin{align}\label{eq:8} 
\text{PL}(\bm{\thetaap};\pl)=\sum_{t=1}^T \big\{\ell_t(\bm{\thetaap}) + \,\pl \|(\bm I-\rho_t \bm W)\bm{\alpha}_t\|_2^2 \,\big\}.
\end{align}

Running a vanilla stochastic gradient descent (with projection on $\Thetawa$) based on this penalized criteria is asymptotically equivalent to applying the proposed PSGD algorithm on the marginal log-likelihood. However, the same algorithm based on the unpenalized likelihood $\text{PL}(\bm{\thetaap};0)$ will have higher estimation error in the estimates of $\alpha_t$ when $L$ is large, which would in turn yield a significantly higher regret. In this context, it is crucial for any decent pricing policy to shrink its $\bm{\alpha}_t$ estimates towards the ellipsoids $\{\bm{\alpha}_t:\|(\bm I-\rho_t \bm W)\bm{\alpha}_t\|_2\leq s_{\omega}\}$. Figure~1 shows the schematic for this essential shrinkage effect on the $\bm{\alpha}_t$. The rigorous mathematical proof is provided in Corollary~\ref{cor.1}.

Our research is connected to and builds on recent works in statistical shrinkage theory, online machine learning and econometrics theory on demand modeling. Next, we list the relevant literature in these research and also briefly mention our contributions.  
\begin{figure*}[t]
            \centering
            \includegraphics[width=0.8\textwidth]{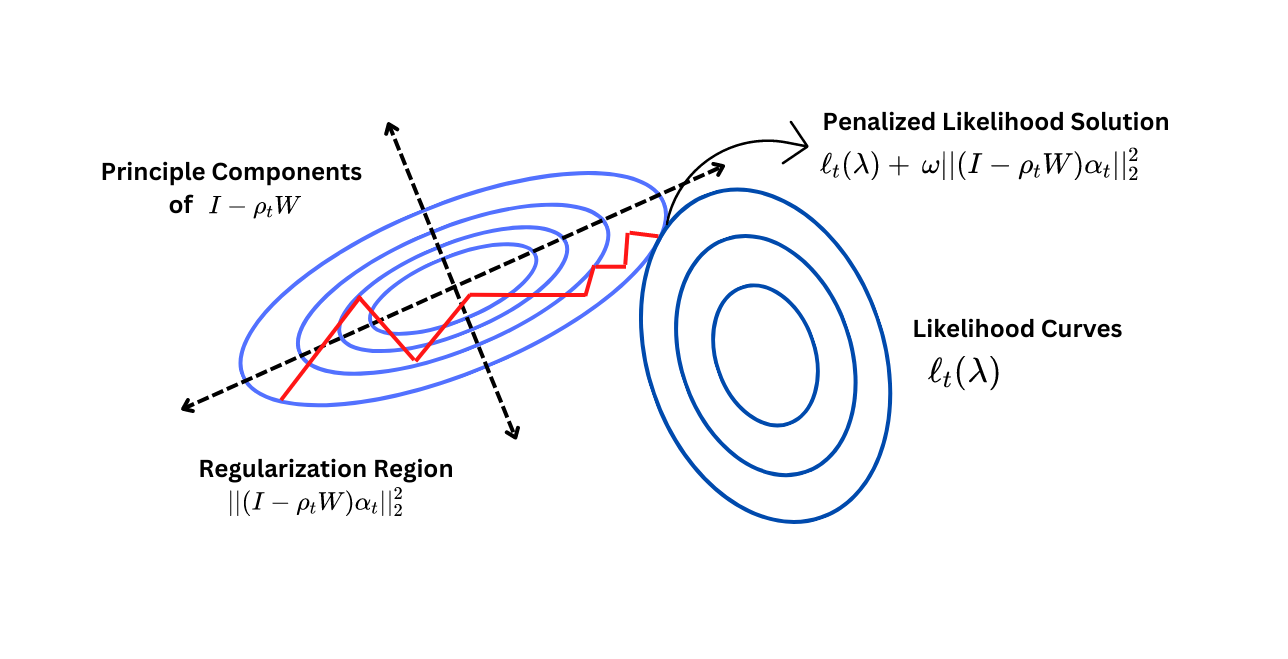}
            \vspace{-1cm}
            \caption{Schematic on the working principle of proposed PSGD. The red path are the PSGD updates on the marginal likelihood with penalty.\\[-7ex]  
            }
            \label{fig:pl}
        \end{figure*}
\begin{description}[leftmargin=*]

\item [1) Dynamic Pricing with Online Learning] There exists a growing body of research on dynamic pricing with learning~\citep{den2015dynamic,farias2010dynamic, harrison2012bayesian, cesa2015regret, ferreira2016online, cheung2017dynamic}. The classical formulations of this problem~\cite{broder2012dynamic, den2013simultaneously,besbes2009dynamic} consider parametric model for the demand-price curve, which is unknown and the learner aims to learn, via exploration-exploitation of prices, while aiming to obtain a low regret in revenue. These works focus on non-contextual settings (no features for customers), and are relevant to applications where a seller is offering an unlimited supply of a single product to the market. Recently, there was significant interest in contextual-models, which use the customers and products attributes to model willingness-to-pay of the buyers for the products, potentially in a heterogeneous
way and offer personalized pricing~\citep{leme2018contextual, cohen2016feature,ban2017personalized,javanmard2019dynamic,lobel2016multidimensional,GolrezaeiJM21}. In addition, some of the recent works in this area~\citep{javanmard2017perishability,keskin2017chasing} aim to model the temporal behavior of buyers, by considering time-dependent demand models. 

Closer to our analysis is the notion of dynamic regret, which has been used in online convex optimization to evaluate the performance of a learner against a dynamic target~\cite{zinkevich2003online,yang2016tracking,jadbabaie2015online,besbes2015non}. However, the general framework of online optimization does not directly apply to our setting, since in the former framework, after taking an action each step the learner observes the incurred loss (or some first order information on it), which can be used in next rounds. In contrast, in our setting the firm only observes the count of sales at each segment, and not the utility of customers. Our work is the first to propose and analyze a dynamic contextual demand model, which accounts for temporal behavior of consumers as well as the network effect among them via spatially autoregressive structure.

\item [2) Dynamic Hierarchical modeling.] Hierarchical modeling provides an effective tool for pooling information across similar units and is one of the most popular approaches for modeling large and complex data sets \citep{fourdrinier2018shrinkage,banerjee2014hierarchical,kou2017optimal}. Here, \eqref{eq:5} imposes a dynamic hierarchical structure on the customer preference coefficients that are linked through a time-invariant non-exchangeable network structure in the second-level prior on \eqref{eq:2}. Recent applications of hierarchical modeling to consumer responses in digital products \cite{banerjeejoint2022,mukhopadhyay2022,banerjee2021improved} have been very successful in analyzing structured longitudinal data-sets abet in a non-dynamic set-up.  Here, we provide an extensive characterization on the operational characteristics of PSGD in a streamed longitudinal set-up and thereby provide theoretical support for the popular PSGD approach for hierarchical modeling in dynamic set-ups.    

\item [3) Modeling Demand Heterogeneity.] Spatial models are very popular in operations management and information sciences to capture non-stationarity in demand \citep{karmakar2021,bimpikis2019spatial,jank2005understanding}. In \eqref{eq:2}-\eqref{eq:5} we have a dynamic spatial model that is governed by changes in the auto-correlation parameter. An important feature of our contributions is that we show PSGD is able to track the variation of the auto-correlation parameter over time and yield asymptotically rate-optimal regret in a dynamic spatial model. 

\item [4) Shrinkage prediction under heterogeneity.] 
It is now commonplace to use notions of shrinkage to improve predictive performances of algorithms in multi-parametric set-ups \citep{hastie2009elements,efron2021computer}.  
Recent results in \citep{xie2012sure,tan2015improved,weinstein2018group,brown2018empirical}  have brought to light new shrinkage phenomena in heteroscedastic models. Here, we study shrinkage prediction in a heteroscedastic dynamic set-up; $n_{lt}$ -- the number of customers in segment $l$ approached at time $t$ can greatly vary over $l$ and $t$. This is an important aspect of our model. It entertains high imbalance across the design matrix but greatly increases applicability. Particularly, in ``pull" marketing systems unlike ``push" systems \citep{peter2000preface} the firm has no control on the number of customers who visits the site/store and does a price check. Thus, $n_{lt}$ will be large in some zip-codes/demographics and quite low in others. Due to the SAR structure in \eqref{eq:5} it is possible to learn the preferences $\alpha_{lt}$ with high precision even in segments with very low $n_{lt}$s. In Section 3 and 4, we illustrate the impact of the heterogeneous $n_{lt}$s on the regret bounds. 
\item [5) Spatial models and applications.]
Spatial models provide a natural way to model the correlation between different units of analysis based on how close they are in a similarity space. Spatial models based on correlated customer preferences have been successfully employed in marketing and economics to model real-world sales data with high predictive accuracy. \cite{jank2005understanding} used a spatially correlated preference model to model consumer choices of two product forms of a book --print or PDF. \cite{yang2003modeling} estimate a binary choice model akin to ours in which consumer preferences for a vehicle's country-of-origin (Japanese/nonJapanese) are spatially correlated based on the distance and demographic similarity between consumers. \cite{ma2015latent} models consumers' decision of whether or not to purchase a callback ringtone. In several non-marketing data applications also, spatial autoregressive (SAR) models have been very successful in assimilating spatial network information in real-world datasets \citep{manski1993identification,  anselin2013spatial}. \cite{bramoulle2009identification} used SAR to model consumption of recreational services such as participation in artistic, sports and social activities by secondary school students, whereas \cite{hsieh2016social} used SAR to incorporate the friendship networks of high school students to predict academic performance. \cite{zhou2017estimating} used SAR to model user activity on social media regarding transportation services in China. Based on this existing literature  which shows that SAR can well capture the correlation among customers and users in economic and social real-world data, we feel that the proposed model will be good for real-world applications in dynamic set-up. 
\end{description}
\section{Proposed PSGD Algorithm}\label{sec.2}

\subsection{Assumptions}
We make some assumptions on the covariate and the parameter space to simplify the presentation of our results. The covariates are normalized such that $\|\bm{x}_{lt}\| \leq 1$. Similarly the parameters $\bm\mu_t$ are such that $\|\bm\mu_t\| \leq C_{\bm\mu}$  where $C_{\bm\mu}$ is a known constant. This gives a ball  of radius $C_{\bm\mu}$ in which the parameters reside. We can even allow the parameter to belong in any convex set $\Theta_\mu$. The results would then depend on the size of the parameter space up to a constant factor. 

Based on \eqref{eq:2}, we also assume that the price sensitivity $\beta_t$ should be negative i.e. an increment in price decreases the utility of the product for the consumer. We also make an assumption on the lower and the upper bound on the magnitude of price sensitivity $c_\beta \leq |\beta_t| \leq C_\beta$. These restrictions inherently create the restricted space $\Theta_{\bm\mu}$ and $\Theta_\beta$ for our model parameters. 

We also make two key assumptions on the SAR structure and auto-correlation parameter.
\begin{description}
\item [\texttt{Assumption 2.1.}] $W$ is a symmetric, PSD kernel e.g. RBF kernel.
\end{description} 

{Since $\bm W$ denotes a distance matrix across the $L$ segments, it is natural that $W$ is a symmetric matrix. It is also worth noting that the common choices of kernels in non-parametric estimation are PSD (see \cite{Tsybakov:2008:INE:1522486}, Section 1.2).}

\begin{description}
\item [\texttt{Assumption 2.2.}] The interaction parameter $\rho_t$ for all time periods is positive and uniformly bounded away from the reciprocal of the maximum eigenvalue of the interaction matrix i.e. $\exists \, \varepsilon \geq 0$, such that $\rho_t \leq ({1 - \varepsilon})/\omega^* $, where $\omega^*$ is the largest eigenvalue of the known interaction matrix $\bm W$.
\end{description} 

{The spatial autoregressive structure of the preference coefficients $\alpha_t$ in equation \eqref{eq:4} with the Gaussian noises implies the joint Gaussian nature of the preference coefficient with variance $(\bm I - \rho_t \bm W)^{-2}$ in \eqref{eq:5}. Since covariance matrices are always positive semi definite, so $\bm I - \rho_t \bm W$ is positive, hence, $\rho_t \leq 1/\omega^*$. With this assumption, we imply the inequality to be strict. Otherwise, the model becomes degenerate (i.e. covariance of Gaussian distribution becomes rank-deficient) and in that case one can work with the lower-dimensional space where the SAR covariance is full-rank.}

\subsection{Reparameterization}\label{sec.reparameter}

The hierarchical prior in \eqref{eq:5} can be used to write the explicit value of $\alpha_{lt}$ in terms of prior hyperparameters $\rho_t$, $\tau$ and standard multivariate normal noise $\epsilon$ as 
$
\alpha_{lt} = \tau\inner{\bm e_l}{(\bm I - \rho_t \bm W)^{-1}\epsilon},$
where $\bm e_l$ denotes the $l^{th}$ basis vector.
Substituting $\alpha_{lt}$ in the utility model , we can rewrite \eqref{eq:2} in terms of $\bm\thetawa$ as 
\begin{align}\label{eq:11.1}
U_{ltk}= \beta_{t}\,p_{lt}+\bm{x}_{lt}' \bm{\mu}_{t} + \sigma Z_{ltk}+ \tau\inner{\bm e_l}{(\bm I - \rho_t \bm W)^{-1}\epsilon}.
\end{align}
This produces a marginal model, where the utility for each segment $l$ can be described as a normal distribution with variance $V^2_{lt} = \| (\bm I - \rho_t \bm W)^{-1} \bm e_l\|^2\tau^2 + \sigma^2$. Next, normalizing the utility to have unit variance we consider the following reparameterized utility model
\begin{align}\label{eq:12.1}
  \Tilde{U}_{ltk}= b_{lt}\,p_{lt}+\bm{x}_{lt}' \bm m_{lt} + Z_{ltk},
\end{align}
where, $b_{lt} = \beta_{t}/V_{lt}$ and $\bm m_{lt} = \bm\mu_t/V_{lt}$. We use this marginal utility model for describing our policy.

\subsection{Optimal Pricing}\label{sec:2.3}

%

Since the noises in \eqref{eq:2.2} are distributed as standard Gaussian, it follows that the optimal price $p_{lt}^*$ is the solution to the equation:
\begin{align}\label{eq:15}
     p_{lt}^*  = -{b}_{lt}^{-1}\frac{\repardist\left({b}_{lt}\,p_{lt}^* +\bm{x}_{lt}' {\bm m}_{lt}\right)}{\repardens \left({b}_{lt}\,p_{lt}^* +\bm{x}_{lt}' {\bm m}_{lt}\right)}~.
\end{align}
The optimality condition in \eqref{eq:15} can be restructured as 
$\varphi(-{b}_{lt}\,p_{lt}^* - \bm{x}_{lt}' {\bm m}_{lt}) + \bm{x}_{lt}' {\bm m}_{lt} = 0$
 where $\varphi(v) = v - {\Phi(-v)}/{\phi(v)}$ is the virtual valuation function \citep{myerson1981optimal}. With the use of the valuation function, we can explicitly describe the optimal price $p_{lt}^*$ as a function of the utility model parameters 
\begin{align}\label{eq:16}
    p_{lt}^* := g(b_{lt}, \bm m_{lt}) =  - \frac{\varphi^{-1}(-\bm{x}_{lt}' {\bm m}_{lt}) + \bm{x}_{lt}' {\bm m}_{lt}}{b_{lt}} ~.
 \end{align}

\begin{proposition}\label{prop.1}
Consider the definition of marginal variance $V^2_{lt} = \| (\bm I - \rho_t \bm W)^{-1} \bm e_l\|^2\tau^2 + \sigma^2$. Under Assumptions 2.1 and 2.2, the variance  $V^2_{lt}$  satisfies
$$c_V \leq V_{lt} \leq C_V,$$
where $c^2_V = \tau^2 + \sigma^2$ and $C^2_V = \tau^2/\varepsilon^2  + \sigma^2$. Additionally, the optimal prices satisfy $p^*_{lt}\le M$, where $M = {c_\beta}^{-1}C_V({C_\mu c_V^{-1} - 0.5\repardens(0)})$.
 \end{proposition}

 
\subsection{Proposed Pricing Policy} 
We propose a pricing policy based on a projected stochastic gradient descent on the loss function described in \eqref{eq:14.1}. With the PSGD, we aim to estimate the reparameterized parameter set $(b_{lt},\bm m_{lt})$ for every segment.

\begin{algorithm}[tb]
   \caption{PSGD based Dynamic Pricing Policy}
   \label{alg.1}
   \label{alg:DP-SGD}
\begin{algorithmic}
   \STATE \textbf{Data } $\bm{W}$ : known segment structure
\STATE \textbf{Initialize} $p_{l1} = c$, , $\hat{b}_{l1} \in \Theta_{b}$ and $\hat{\bm m}_{1t} \in \Theta_{m}$ $\forall l$
\FOR{$t=1, 2, \dots$}
\STATE \textbf{Data} $y_{lt}$, $\bm{x}_{l,t+1}$ : Longitudinal data stream
\STATE 1. Compute the gradient $\mathcal{L}_{lt}'(\hat{\bm m}_{lt},\hat{b}_{lt})$  \eqref{eq:15.1} for each segment $l$
\STATE 2. Update parameters by moving in the opposite direction of gradient with step size $\eta_t$ and then projecting onto the restricted space \eqref{eq:19}
\begin{align*}
    \hat{b}_{l,t+1} = \Pi_{\Theta_{b}}(\hat{b}_{lt} - \eta_t \nabla \mathcal{L}_{lt}^{b}); \quad \quad 
    \hat{\bm m}_{l,t+1} = \Pi_{\Theta_{ m}}(\hat{\bm m}_{lt} - \eta_t \nabla \mathcal{L}_{lt}^{\bm m} ) 
\end{align*}
\STATE 3. Set price $p_{l,t+1}$  using the optimal pricing function
$p_{l,t+1}  = g(\hat{b}_{l, t+1}, \hat{\bm m}_{l,t+1})$
\ENDFOR
\end{algorithmic}
\end{algorithm}

Based on the assumptions $\|\bm\mu_t\| \leq C_{\bm\mu}$ and $|\beta_t| \leq C_\beta$ we first define $\Theta_b := \{\beta/c_V:\beta \in \Theta_\beta\}$ and $\Theta_{m} := \{\bm\mu/c_V:\bm\mu \in \Theta_\mu\}$, the restricted space of the new parameters $b, \bm m$. These are natural extensions to the assumptions since $b_{lt} = \beta_t/V_{lt}$ and $c_V$ is the lower bound on $V_{lt}$. 

Next, define the loss function as the negative of the log-likelihood function for the utility model \eqref{eq:12.1}.

\begin{align}\label{eq:14.1}
\mathcal{L}_t(\bm{\thetaap})= - \sum_{l=1}^L y_{lt} \log q_{lt} + (n_{lt}-y_{lt})\log (1-q_{lt}),
\end{align}

with $q_{lt} = \repardist(b_{lt}\,p_{lt}+\bm{x}_{lt}' \bm m_{lt})$. Each summand in the loss \eqref{eq:14.1} is the loss for a specific segment $l$. Let $\mathcal{L}_{lt}$ denote these losses, $$\mathcal{L}_{lt} = - y_{lt} \log q_{lt} - (n_{lt}-y_{lt})\log (1-q_{lt}).$$ 
 We compute the gradient of loss functions: $\mathcal{L}_{lt}'(b_{lt},\bm m_{lt}) = (\nabla \mathcal{L}_{lt}^{(b)}, \nabla \mathcal{L}_{lt}^{(m)})$ for each segment as
 \begin{align}\label{eq:15.1}
      \nabla \mathcal{L}_{lt}^{(b)} = - y_{lt} \frac{\repardens(u_{lt}^0)}{\repardist(u_{lt}^0)}+ (n_{lt}- y_{lt})\frac{\repardens(-u_{lt}^0)}{\repardist(-u_{lt}^0)}p_{lt},\\
      \nabla \mathcal{L}_{lt}^{(m)} = - y_{lt} \frac{\repardens(u_{lt}^0)}{\repardist(u_{lt}^0)}+ (n_{lt}- y_{lt})\frac{\repardens(-u_{lt}^0)}{\repardist(-u_{lt}^0)}\bm x_{lt}.
 \end{align}
At time point $t$, we move in the opposite direction of the gradient with step size $\eta_t$. The resultant estimates are then projected onto the restricted space $\Theta_b,\Theta_{\bm m}$  based on the assumptions on the size of the parameters to get the successive estimates: 
 %
\begin{align}
    \hat{b}_{l,t+1} &= \Pi_{\Theta_{b}}(\hat{b}_{lt} - \eta_t \nabla \mathcal{L}_{lt}^{b}), \\
    \hat{\bm m}_{l,t+1} &= \Pi_{\Theta_{ m}}(\hat{\bm m}_{lt} - \eta_t \nabla \mathcal{L}_{lt}^{\bm m} ),\label{eq:19}
\end{align}
where, $\Pi_{\Theta_b}(.)$ and $\Pi_{\Theta_m}(.)$ are the projection functions on to the convex set $\Theta_b$ and $\Theta_m$ respectively. The policy finally uses these estimated parameters and the optimal pricing function $g(\cdot, \cdot)$ defined in \eqref{eq:16} to set the price for the next period. The method is summarized in Algorithm~\ref{alg.1}.

\section{Theoretical Results}\label{sec.3}

In this section, we provide the bounds on the regret for the dynamic pricing policy we employ. We show that under regularity conditions on the temporal nature of the parameters $\beta$, $\bm \mu$ and $\rho$, the regret as defined in \eqref{eq:7} has order square root of the time horizon $T$. We also show the optimality of the bound by showing that no policy can achieve a worst-case regret better than the same rate.

\subsection{Upper Bound on Regret of the Proposed Algorithm}

We present an upper bound on the regret of the proposed pricing policy in terms of the reparameterized model parameters in \eqref{eq:12.1}. Later, through lemma~\ref{lemma.1}, we create a link between the actual parameters and the reparameterized ones. Finally, we show that when the step sizes $\eta_t \propto 1/\sqrt{t}$, we achieve $\mathcal{O}(\sqrt{T})$ regret.
\begin{theorem}\label{thm.1}
For any $\bm{\theta} \in \Theta$ defined below \eqref{eq:7}, the regret of our proposed policy satisfies: 
\begin{align}
\mathcal{B}(\bm{\thetawa},\bm{p})\leq \mathcal{R}_1 + \mathcal{R}_2 + \mathcal{R}_3 + \mathcal{R}_4 + \mathcal{O}(\log T), \text{where},\label{eq.thm.1}
\end{align}
\begin{align*}
&\mathcal{R}_1 = C_1 \sum_{t = 1}^T \sum_{l = 1}^L \eta_{t}^{-1}|b_{l,t+1} - b_{l,t}|, \\
&\mathcal{R}_2 = C_2 \sum_{t = 1}^T \sum_{l = 1}^L \eta_{t}^{-1}\|\bm m_{l,t+1} - \bm m_{l,t}\|_2, \\
&\mathcal{R}_3 = C_3 \sum_{t = 1}^T \sum_{l = 1}^L \eta_{t}   {n_{lt}^2} \leq C_3 \sum_{t = 1}^T \eta_t n_{t}^2, \text{and}, \\ 
&\mathcal{R}_4 = C_4~ \eta_{T+1}^{-1} \, L .
\end{align*}
   $C_1$, $C_2$, $C_3$ and $C_4$ are constants independent of $T$, $n$, $L$ and the model parameters. 
\end{theorem}
The detailed proof of the  theorem is presented in the Appendix. We present a brief overview of its proof in Section 4. We next concentrate on further explaining the terms on the right side of \eqref{eq.thm.1}. 
We simplify the first two terms $\mathcal{R}_1$, $\mathcal{R}_2$ in theorem \ref{thm.1} and provide a key lemma \ref{lemma.1}. For ease of notation we define  $\delta_{t\nu} = \|\nu_{t+1} - \nu_t\|$ for any parameter $\nu$. This helps us transform our regret from the reparameterized quantities $b$, $\bm m$ to the original parameters $\beta$, $\bm \mu$ and $\rho$. 

\begin{lemma}\label{lemma.1}
    Let $\omega_*$ be the smallest eigenvalue of $\bm W$, then under Assumptions 2.1 and 2.2, the variation across the parameters in the utility model \eqref{eq:12.1}, can be bounded as 
    \begin{align}
       |b_{l,t+1} - b_{l,t}| \leq {\tau}^{-1}({1 - \rho_t\omega_*}){\delta_{t\beta}} + C_5 {\delta_{t\rho}}~,\\
       \|\bm m_{l,t+1} - \bm m_{l,t}\|_2  \leq {\tau}^{-1}({1 - \rho_t\omega_*}){\delta_{t\mu}} + C_5 {\delta_{t\rho}}.
    \end{align}
\end{lemma}

Next, we demonstrate the implications of the above result in a simplified setup with any network structure $\bm W$. The goal is to understand the effect of the network structure and the auto-correlation $(\rho_t)$ on the upper bound of the regret in \eqref{eq.thm.1}. We provide the following corollary that explicitly shows the relation of regret with the auto-correlation parameter. 

\begin{corollary}\label{cor.2}
If $\eta_t \propto 1/\sqrt{t}$, and $\rho_t = \rho$ for all $t$, then the dynamic pricing policy based on Algorithm \ref{alg.1} has regret

\begin{align}\label{eq:23}
   \mathcal{B}(\bm{\thetawa},\bm{p})\leq  {C}_6 {\tau}^{-1}({1 - \rho\omega_*})\sum_{t = 1}^T\sqrt{t}({\delta_{t\beta} + \delta_{t\mu}}) + \mathcal{O}(\sqrt{T}).
\end{align}
\end{corollary}

Corollary \ref{cor.2} shows that the regret has two parts, one with order $\sqrt{T}$, while the other part depends on the temporal nature of price sensitivity and customer preferences. The regret occurred in this part depends on the strength of the network inversely, i.e, higher the strength of the network (higher the $\rho$) lower the regret and vice-versa.    

Note that if $\rho_t$ was varying across time, we can extend the bound in Corollary~\ref{cor.2}. Assume that $\rho_* = \min_t \rho_t$, then the bound on regret can be modified as

\begin{align}
   \mathcal{B}(\bm{\thetawa},\bm{p})\leq  &{C}_6 {\tau}^{-1}({1 - \rho_{*}\omega_*})\sum_{t = 1}^T\sqrt{t}({\delta_{t\beta} + \delta_{t\mu}})  + C_7 \sum_{t = 1}^T \sqrt{t}\delta_{t\rho} + \mathcal{O}(\sqrt{T}).\label{eq:23.1}
\end{align}

where $C_6$ and $C_7$ are constants. The bound above behaves similarly to Corollary~\ref{cor.2} if the temporal changes across auto-correlation is small.

\subsection{Lower Bound on Regret of Any Data-driven Policy}

We show that the bound in corollary \ref{cor.2} is indeed tight in terms of dependence on the time horizon. In the next theorem we show that there exists parameters in the space $\Thetawa$ such that under the demand model with these parameters, the regret  of any policy is of the order at least $\sqrt{T}$. The detailed proof is provided in the Appendix. 

\begin{theorem}\label{thm.2}
    Consider the utility model \eqref{eq:12.1} and let $N_T:=\sum_{t=1}^T \sum_{l=1}^L n_{lt}$ be the total number of costumers across all segment and times up to $T$. For any fixed graph $\bm W$, the worst-case risk of any data driven pricing policy $\hat{\bm p}$ satisfies

    \[\min_{\hat{\bm p}}\max_{\bm\theta \in \Theta}\mathcal{B}(\bm{\thetawa},\bm{p}) \geq C_8 \sqrt{T} (1+\log(N_T/T))\,,\]

    for some constant $C_8$.
    In particular, if $n_{lt}\ge 1$ for all $l, t$, we have

    \[\min_{\hat{\bm p}}\max_{\bm\theta \in \Theta}\mathcal{B}(\bm{\thetawa},\bm{p}) \geq C_8 \sqrt{T} (1+\log(L))\,.\]
\end{theorem}

Theorem \ref{thm.2} along with \eqref{eq:23.1} implies that our pricing policy in algorithm~\ref{alg.1}, is optimal, if the temporal changes across the price sensitivity, customer preferences and auto-correlation is of the order $\sqrt{T}$, i.e. if $\sum_{t = 1}^T\sqrt{t}(\delta_{t\beta} + \delta_{t\mu} + \delta_{t\rho}) = \mathcal{O}(\sqrt{T})$, then our policy is order optimal.

\subsection{Sub-optimality of Unshrunken Pricing Policies}

Next, we consider unshrunken policies that do not incorporate the structure \eqref{eq:5} on the $\bm{\alpha}_t$s.
Such unshrunken policies suffer from severe noise accumulation in estimating $\bm{\alpha}_t$ as free parameters at every time point. The following result whose proof is provided in Section \ref{append.cor.1} of the appendix shows that the Bayes regret from any unshrunken pricing policies based on the unpenalized likelihood is highly sub-optimal as compared to the proposed strategy $\bm p$. Consider a parametric space $\bar{\Theta}$ such that any $\bm \theta \in \bar{\Theta}$ satisfies that $\sum_{t = 1}^T \sqrt{t}\delta_{t\beta}$, $\sum_{t = 1}^T \sqrt{t}\delta_{t\mu}$ and $\sum_{t = 1}^T \sqrt{t}\delta_{t\rho}$ are $\mathcal{O}(\sqrt{T})$. The following result shows sub-optimality of unshrunken pricing policies over $\bar{\Theta}$.


\begin{lemma}\label{cor.1}
For any $\bm\theta \in \bar{\Theta}$, the regret of any data-driven policy $\bm{p}_U$ based on the unpenalized likelihood in \eqref{eq:8} satisfies
$$\mathcal{B}(\bm{\theta},\bm{p}_U)\big/ \mathcal{B}(\bm{\theta},\bm{p}) = \Omega(\sqrt{T}).$$
\end{lemma}

\section{Outline of Proofs and Overview of Techniques}

For the detailed proofs of all the results, we refer to the appendix. In this section,  we delve into the intuition and the intermediate steps used in proving the two main results in Section~\ref{sec.3}. 

\subsection{Proof Sketch of Theorem \ref{thm.1}}
The crucial idea is to bound the revenue loss (regret) with the parameters in the model. To achieve that consider the revenue function with the utility model \eqref{eq:12.1} 
\begin{align}\label{eq:Rev-mod}
    \text{Rev}_{lt}(p_{lt}) = n_{lt}p_{lt}\repardist(b_{lt}p_{lt} + \bm x_{lt}'\bm m_{lt}).
\end{align}

The revenue loss using our policy is then the difference between $\text{Rev}_{lt}(p_{lt}^*)$ and $\text{Rev}_{lt}(p_{lt})$ where $p_{lt}^*$ is the optimal price for the model true parameters and $p_{lt}$ is the price posted with the dynamic pricing policy. 

\begin{proposition}\label{prop.2}
    There exists a constant $C_9$ such that the regret of our policy on segment $l$ at time $t$ can be bounded as 
    \begin{align}\label{eq:25}
        \mathcal{R}_{lt} = {\rm Rev}_{lt}(p_{lt}^*) - {\rm Rev}_{lt}(p_{lt}) \leq C_9 n_{lt}(p_{lt} - p_{lt}^*)^2\,,
    \end{align}
    where $p_{lt}$ is our posted price and $p_{lt}^*$ is the optimal price that maximizes the revenue under known parameters.
\end{proposition}

We simplify the regret bound term $(p_{lt} - p_{lt}^*)^2$ on the right hand side of \eqref{eq:25} in our next lemma. The idea is to use the optimal pricing function $g(\cdot,\cdot)$ defined in section~\ref{sec:2.3}. The prices $p_{lt}^*$ and $p_{lt}$ can then be defined as $p_{lt}^* = g(b_{lt}, \bm m_{lt})$, the optimal price based on the true parameters and $p_{lt} = g(\hat{b}_{lt}, \hat{\bm m}_{lt})$, the optimal price with respect to the estimated parameters that our proposed policy posts. The lemma then hinges on the fact that the function $g(\cdot,\cdot)$ defined in \eqref{eq:16} is Lipschitz. 

\begin{lemma}\label{lemma.2}
 For model \eqref{eq:12.1}, under the true parameters $b_{lt}, \bm{m}_{lt}$ and the output $\hat{b}_{lt}, \hat{\bm{m}}_{lt}$ from our PSGD pricing policy, the following holds true: 
 \begin{equation}
     (p_{lt} - p_{lt}^*)^2 \leq C_{10} \langle\bm{x}_{lt}, \bm m_{lt} - \hat{\bm m}_{lt}\rangle^2 + C_{10} p_{lt}^2(b_{lt} - \hat{b}_{lt})^2
 \end{equation}
 for some constant $C_{10}>0$. 
\end{lemma}

The $\mathcal{R}_{lt}$ terms in \eqref{eq:25} are the building blocks for our total regret $\mathcal{B}(\bm{\thetawa},\bm{p})$ as in \eqref{eq:7} where $\mathcal{B}(\bm{\thetawa},\bm{p}) = \sum_{t = 1}^T\sum_{l = 1}^L\mathcal{R}_{lt}$. The above two lemmas relate the revenue regret occurred by the policy with the estimation error of the parameters in the model \eqref{eq:12.1}. The final step involves creating a link between this estimation error and the temporal nature of the parameters to achieve the regret bound as in Theorem~\ref{thm.1}.

\subsection{Proof Sketch of Theorem \ref{thm.2}}
For the lower bound, we want to find worst case scenarios in terms of parameters. In this case, we use the idea of ``uninformative prices" \citep{broder2012dynamic}. These are prices where the purchase probability curves for all different sets of parameters intersect. Such prices do not reveal any information about the parameters since  all the purchase curves contain the point. These uninformative prices become an issue when they are also the optimal prices for some set of parameters. If a policy wants to learn the parameters fast, they need to do exploration away from these uninformative prices. But during the process of exploration it chooses parameters farther from the actual parameters and thus increases regret.

The general idea of such proofs is to create a setting where these optimal prices are indeed uninformative. In the proof, we show existence of such parameters and their corresponding optimal ``uninformative prices". Calling these parameters $\gamma_0$, the proof hinges on two relations, one showing that learning the utility model parameters closely is expensive in terms of regret: 
\begin{align}\label{eq:KL-LB}
\text{Reg}_T^{\pi,\gamma_0} \geq \frac{C_{12}}{(\gamma_0 - \gamma)^2}\KL\left(f_T^{\pi, \gamma_0} ; f_T^{\pi, \gamma}\right)
\end{align}
where $f_t^{\pi, \gamma}$ is the density of purchases for all consumers until time $t$, provided that the policy $\pi$ is employed. An interpretation of the KL-divergence $\KL\left(f_t^{\pi, \gamma_0} ; f_t^{\pi, \gamma}\right)$ is the certainty level of the policy $\pi$ about the true model parameters $\gamma_0$, over some other counterfactual parameter $\gamma$. So the above bound implies that increasing certainty about the underlying model is costly. 

The next bound shows that if the policy can not differentiate between two parameters that are ``close" to each other (in other words to increase its confidence in one), then again it incurs a large regret. Specifically, if $\gamma_1 = \gamma_0 + 1/(4T^{1/4})$,
\begin{align}\label{eq:KL-LB2}
\text{Reg}_T^{\pi,\gamma_0} + \text{Reg}_T^{\pi,\gamma_1} \geq C_{13}\sqrt{T}e^{-\KL\left(f_T^{\pi, \gamma_0} ; f_T^{\pi, \gamma_1}\right)}.
\end{align}

Intuitively, the first equation shows that exploitation is necessary (choosing the optimal parameter $\gamma_0$ to have small KL-divergence) and the second one asks for exploration to stay away from uninformative prices, so as to gain information about the model parameters and increase the certainty about it, as measured by KL-divergence. 

\section{Numerical Experiments}\label{sec:num_exp}

We study the performance of the proposed algorithm using numerical experiments based on synthetic as well as real-world based networks. We consider a wide range of regimes with varying (a) temporal variation of the model parameters (b) strength and nature of the network (c) sampling heterogeneity across sectors, and (d) noise distributions.

\textbf{Set-up 1.} Consider $L=10$ segments and a time-invariant sampling policy with different sampling rates across two segment groups: for any $t\geq 1$, $n_{lt}=50$ for $l=1,\ldots,5$ and $n_{lt}=200$ for $l=6,\ldots,10$. We use a time-invariant network $\bm W$ that was generated using radial basis function (RBF) kernel of width one on independent standard Gaussian feature vectors, drawn from input space $\mathbb{R}^{10}$. 
 We use bivariate covariates $\bm x_{lt}$ generated from standard exponential distribution and set $\tau=1,\sigma=1$ in \eqref{eq:2}-\eqref{eq:4}. The price sensitivity and the customer preferences are assumed as $\beta_1 = -0.4$ and $\bm\mu_1 = (0.1, 0.15)$. With change in time the parameters change as follows,
$\beta_{t+1} = \beta_t + \delta_{t\beta}; \quad \bm\mu_{t+1} = \bm\mu_t + \bm\delta_{t\mu}$,
where $\delta_{t\beta} = t^{-b}\tilde{Z_t}/(10|\tilde{Z_t}|)$ and $\bm\delta_{t\mu} = t^{-b}\bar{\bm Z_t}/(10\|\bar{\bm Z_t}\|)$ where $\tilde{Z_t}$ and $\bar{\bm Z_t}$ are standard Gaussian random variables of dimension $1$ and $2$ respectively.

\begin{figure}[!htb]
    $\qquad$
    {\includegraphics[width=0.4\textwidth]{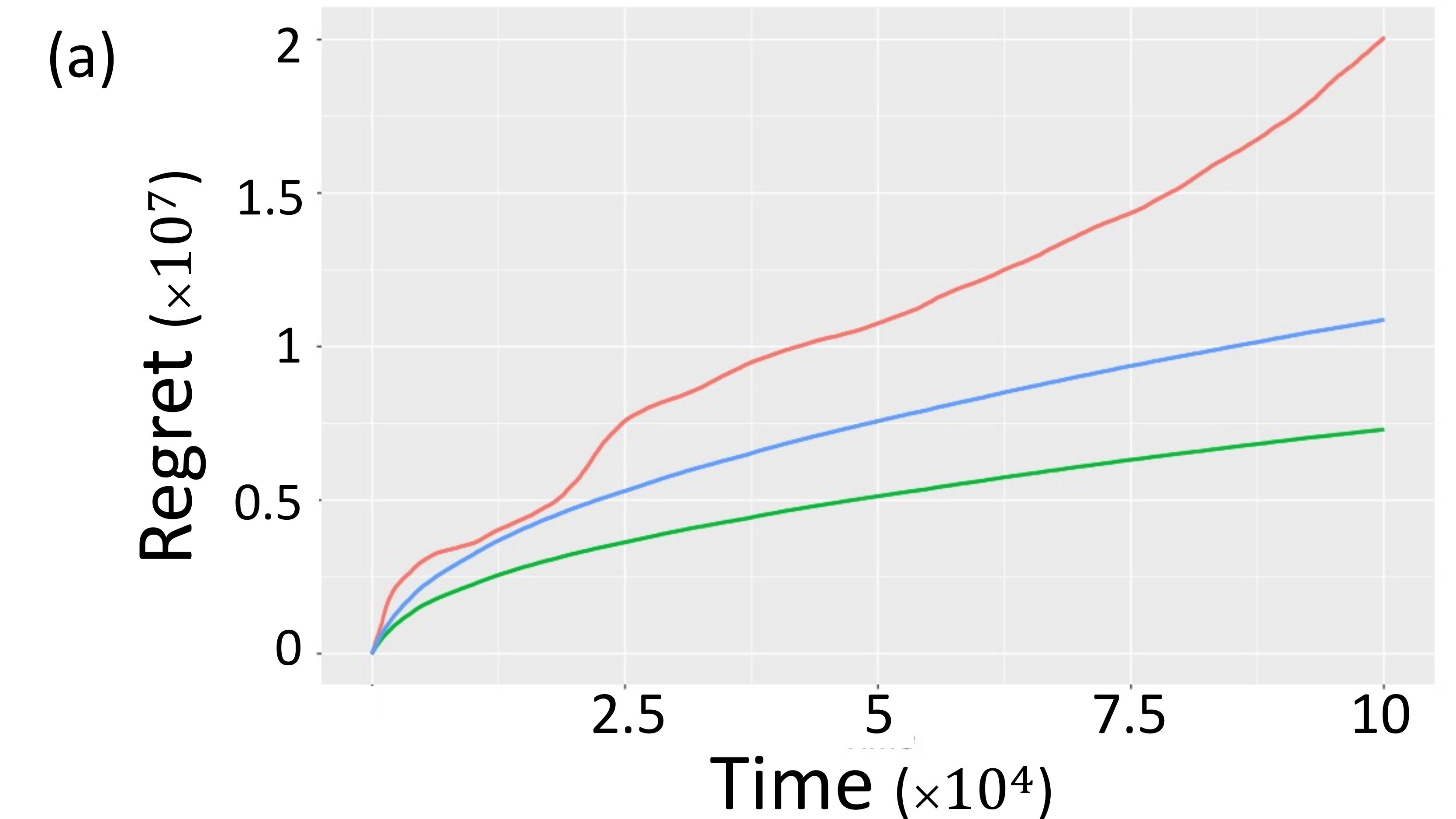}}$\qquad$
    {\includegraphics[width=0.4\textwidth]{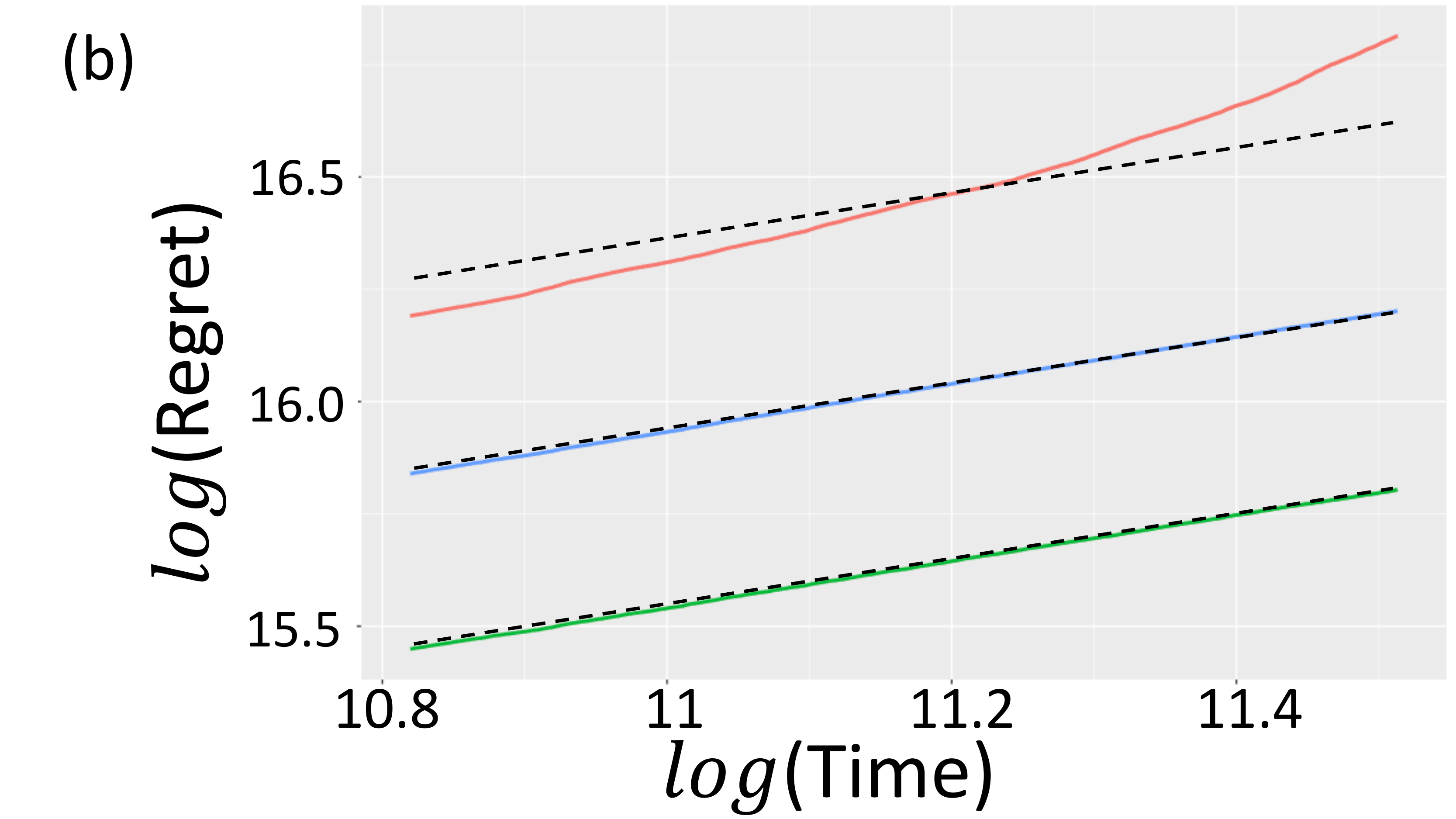}} 
    \caption{Regret plots of the proposed policy in Set-up 1 as $b$ which governs the shift in the model parameters over time changes. The plots for $b=0.5, 1, \infty$ are in red, green and blue respectively. In panels (a) Regret in original scale (b) log(Regret) vs log(T). The dotted lines in panel (b) are the best fitted line with slope 0.5.}
    \label{fig:regret_b_vary}
\end{figure}

We set $\rho_t$ to $0.5$ for all $t\geq 1$ and consider three cases, $b=0.5, 1, \infty$, for the temporal variations across $\beta_t$ and $\bm \mu_t$. Note that the case of $b=\infty$ corresponds to the scenario where the parameters do not change over time.  Following from Corollary~\ref{cor.2}, the regret for the two cases of $b = 1, \infty$ should be of order $O(\sqrt{T})$, while the regret for $b = 0.5$ should be $\mathcal{O}(T)$. In Figure~\ref{fig:regret_b_vary}, we plot the regret (cumulative revenue lost to the oracle policy) over time for the three cases. From the figures it is evident that when $b = 0.5$, the regret from the proposed method eventually grows linearly where as in the other two cases its is controlled at $\mathcal{O}(\sqrt{T})$.

\begin{figure}[!htb]
    $\qquad$
    {\includegraphics[width=0.4\textwidth]{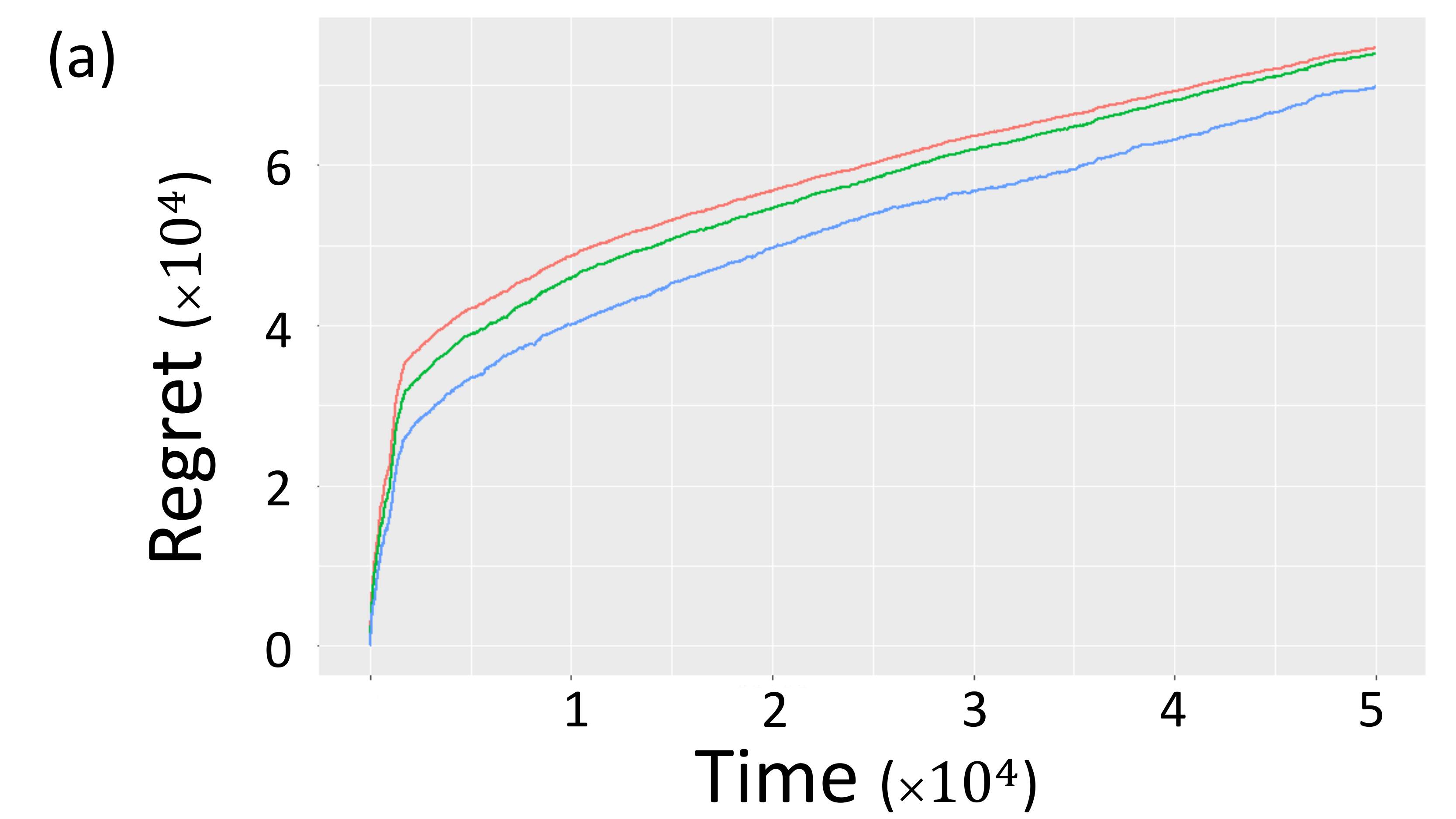}}$\qquad$
    {\includegraphics[width=0.4\textwidth]{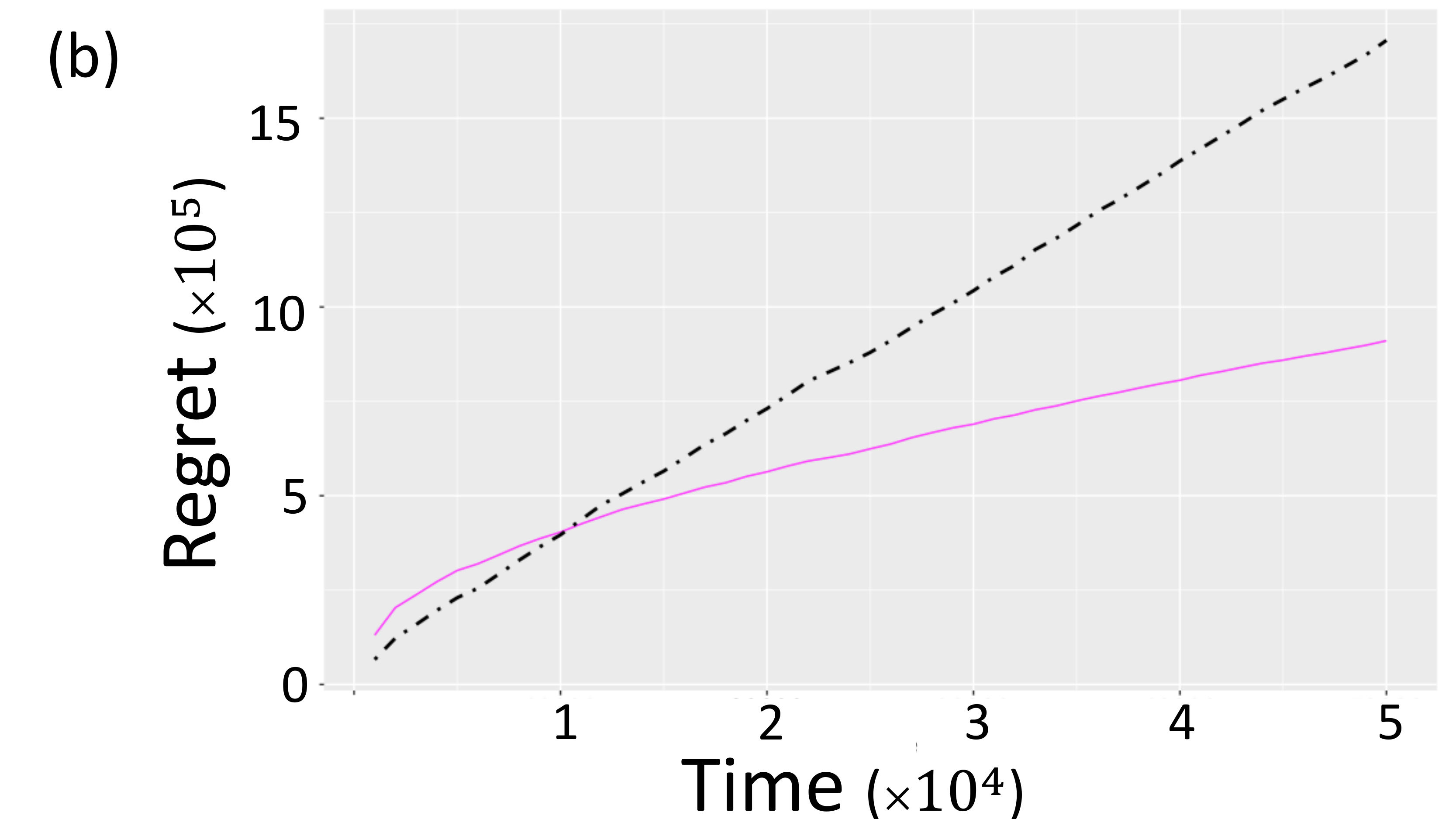}}
    \caption{Regret plots for Set-up 2 (a) Regret of proposed policy for $\rho = 0.5 \text{ (blue)}, 0.3 \text{ (green)}, 0.1 \text{ (red)}$ varies (b) Regret of proposed policy (continuous line) and an unshrunken policy (dotted line) for $\rho=0.5$.}
    \label{fig:rho_vary_unshrunken}
\end{figure}

\textbf{Set-up 2.} Here, we aim to study the performance of the proposed method as the strength of the network varies due to change in auto-correlation parameter. We consider a simpler setting than set-up 1 with $L=4$ segments with a homogeneous sampling rate $n_{lt}=50$ for all $l,t$. We consider $b=1, \rho_t=\rho$ for all $t\geq 1$ and vary $\rho = 0.1, 0.3$ and $0.5$ across 3 experiments. Note that, the first two terms $\mathcal{R}_1$ and $\mathcal{R}_2$ in Theorem~\ref{thm.1} depend on $\rho$ while the third term $\mathcal{R}_3$ increases with $n_{lt}$. This means that as $n_{lt}$s increase, $\mathcal{R}_3$ grows very large and the effect of $\mathcal{R}_1$ and $\mathcal{R}_2$ (effect of $\rho$) on the regret is significantly less. 
Hence, to see the effect of $\rho_t$, we consider moderate $n_{lt}$s here. 
We plot the regret in Figure~\ref{fig:rho_vary_unshrunken} (a) and see a significant improvement in terms of regret as $\rho$ increases gradually. In Figure~\ref{fig:rho_vary_unshrunken} (b) we fix $\rho=0.5$ and compare the regret of our policy with the unshrunken policy based on \eqref{eq:8}. We see that the regret of the unshrunken policy has linear trend and is quite sub-optimal as $T$ exceeds $20000$. 

\begin{figure}[!htb]
\centering
{\includegraphics[width=0.7\textwidth]{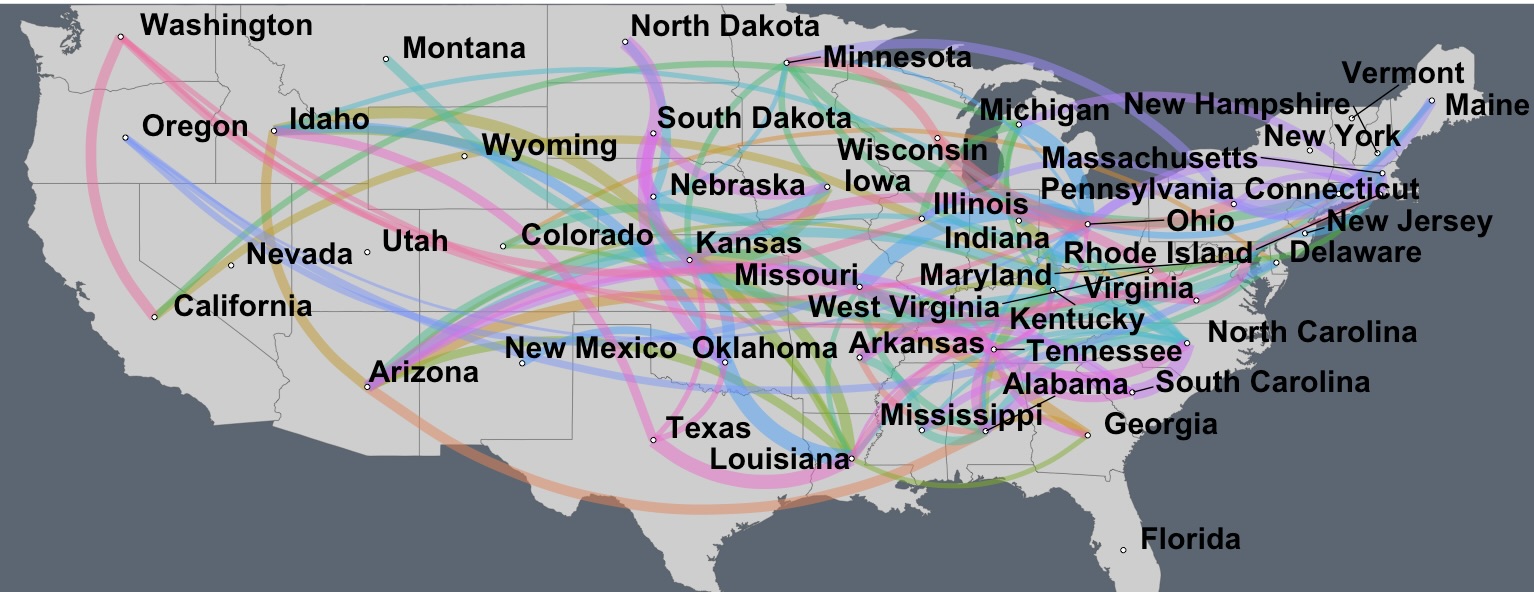}}
{\includegraphics[width=0.8\textwidth]{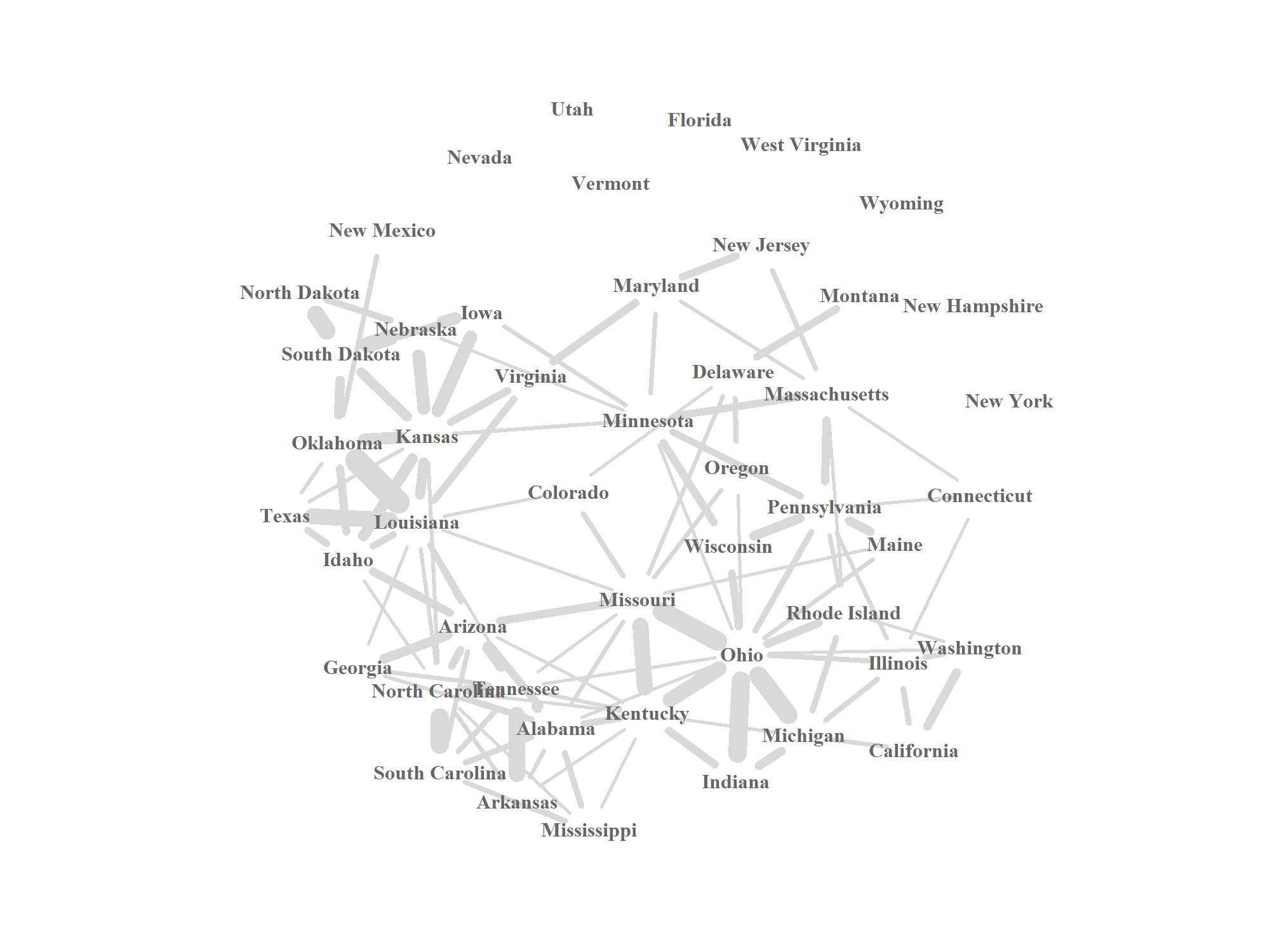}}

    \caption{A network on $L=48$ US states barring Hawaii and Alaska. This network is based on similarity between states across $15$ demographic and economic variables and is thresholded at $0.05$. The network is used in experiments for set-ups 3 and 4.}
    \label{fig:map_network}
\end{figure}

\textbf{Set-up 3.}
In this set-up, unlike set-up 1 and 2, we do not consider a randomly generated synthetic network but consider a real-network based on US census data \cite{census2008statistical}. Here, we consider $L=48$ segments. Each segments constitute a US state.  For ease of analysis and presentation, we remove Hawaii and Alaska from the analysis. We choose 15 demographic and socio-economic variables such as percentage of residents in the age group 5-65, average income, unemployement rates, etc for making the network matrix $\bm{W}$ among the $L$ segments.  We use an RBF kernel of width two and threshold the resultant network at $0.05$ level, i.e., edges with weight less than $0.05$ are deleted from the network. Figure~\ref{fig:map_network} shows the network.  

\begin{figure}[!htb]
\centering
{\includegraphics[width=0.6\textwidth,trim={0 0 0 0},clip]{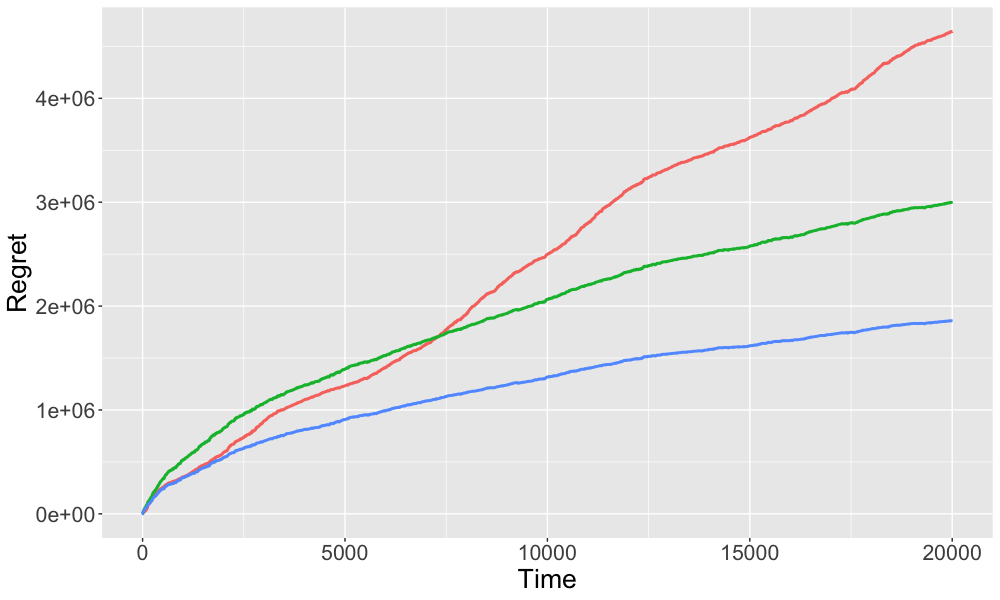}}
    \caption{Regret of the proposed policy in Set-up $3$ for different values of b.}
    \label{fig:b_vary_map}
\end{figure}

We generate the covariates $\bm{x}_{lt}$ using standard exponential distribution. We apply model (1)--(2) in the main paper in the context of conversion from leads. Consider the problem where at time $t$, the firm purchases leads $n_{lt}$ for segment $l$ from other lead generation companies and shows pricing $p_{lt}$ which lead to conversion $Y_{lt}$. We consider that the number of leads $n_{lt}$ is fixed over time $t$. Let $n_t=n$. For this set-up, we vary $n=1000, 2500, 5000, 10000, 20000$. We set 
$$n_{lt} \propto \text{Population size}_{l}\times \text{Median Income}_{l},$$
i.e., the number of leads in each state is proportional to the population of the state as well as the median disposable income per household in the state.  Similar to the analysis in Section~\ref{sec:num_exp}, we consider the three cases , $b=0.5, 1, \infty$, for the temporal variations across $\beta_t$ and $\bm \mu_t$.  In Figure~\ref{fig:b_vary_map}, we plot the regret for the three cases. We see that the results perfectly match the results in Set-up 1 where we had a random network: with $b = 0.5$, the regret grows linearly, whereas with $b = 1$ and $b = \infty$, the regret is $\mathcal{O}(\sqrt{T})$.

\textbf{Set-up 4.} In the previous setting we assumed that the number of leads in each segment is proportional to the population and the income levels. But for most firms it is highly unlikely that they have good penetration and market share in all the US states. 
We consider the scenario where the firm has a stronger customer base in some states compared to others. Thus, there will be difference in leads across states. Particularly, in states $l$ where the firm has low penetration, $n_{lt}$s will be very low. An interesting attribute of this exercise is that the presence of the network structure in (1)-(8) makes the preferences correlated and the preference coefficients from states with low $n_{lt}$s can also be efficiently learnt by the prescribed method by leveraging the information from states with high penetration.

\begin{table}[htbp]
  \centering
  \caption{Performance of the prescribed method relative to Unshrunken policy in Set-up 4 as $\upsilon_1(1-\upsilon_1)$ varies across columns. 
  (negative implies worse performance)\\}
        \begin{tabular}{|c|r|rrr|}
    \toprule
    \textbf{n} & \multicolumn{1}{c|}{\textbf{T}} & 
    \multicolumn{1}{c|}{\textbf{0.7}} & \multicolumn{1}{c|}{\textbf{0.8}} & \multicolumn{1}{c|}{\textbf{0.9}} \\
    \midrule
    \multirow{4}[0]{*}{\textbf{1000}} & \textbf{100} & -1.3\% & -4.4\% & -0.8\% \\
          & \textbf{500} & 3.2\% & -0.8\% & -0.8\% \\
          & \textbf{1000} & 15.0\% & 13.9\% & 12.5\% \\
          & \textbf{5000} & 50.5\% & 50.5\% & 48.5\% \\\midrule
    \multirow{4}[0]{*}{\textbf{2500}} & \textbf{100} & -2.9\% & -5.0\% & -5.3\% \\
          & \textbf{500} & 2.2\% & 0.8\% & -1.4\% \\
          & \textbf{1000} & 17.1\% & 14.8\% & 11.8\% \\
          & \textbf{5000} & 51.2\% & 50.3\% & 48.2\% \\\midrule
    \multirow{4}[0]{*}{\textbf{5000}} & \textbf{100} & -2.2\% & -5.4\% & -3.1\% \\
          & \textbf{500} & 5.1\% & 1.7\% & -0.1\% \\
          & \textbf{1000} & 18.0\% & 15.0\% & 12.0\% \\
          & \textbf{5000} & 52.3\% & 49.8\% & 48.1\% \\\midrule
    \multirow{4}[0]{*}{\textbf{10000}} & \textbf{100} & -2.4\% & -4.9\% & -3.8\% \\
          & \textbf{500} & 3.0\% & 1.1\% & -1.4\% \\
          & \textbf{1000} & 16.3\% & 15.0\% & 10.8\% \\
          & \textbf{5000} & 51.3\% & 50.3\% & 47.5\% \\\midrule
    \multirow{4}[1]{*}{\textbf{20000}} & \textbf{100} & -1.9\% & -4.9\% & -4.4\% \\
          & \textbf{500} & 4.5\% & 1.1\% & -1.5\% \\
          & \textbf{1000} & 17.3\% & 15.0\% & 10.9\% \\
          & \textbf{5000} & 51.8\% & 50.3\% & 47.7\% \\
    \bottomrule
    \end{tabular}%
  \label{tab:rel_reg_all}%
\end{table}%

To study this through numerical experiments, we create an imbalanced design. We divide the states into two groups $L_1$ and $L_2$ of equal sizes. Here, $n_{lt}$ are not only proportional to population and income as in Set-up 3 but states in $L_1$ are given more weightage that those in $L_2$, i.e., $\upsilon_1=\sum_{l \in L_1} n_{lt} (\sum_{l \in L_2} n_{lt})^{-1}>1$. With the same network structure as in set-up 3, we compare the regret of the proposed policy compared to an unshrunken policy at three different levels of $\upsilon_1$. The first level is $0.7:0.3$. The second and third level of imbalance is $0.8:0.2$ and $0.9:0.1$. 

We compute the cumulative regret for all three imbalanced cases for different values of total number of customers $n_t=n$ and time horizon $T$.
We report the relative regret (in terms of percentage) by our proposed policy against the cannonical unshrunken policy in Table~\ref{tab:rel_reg_all}. Relative to the unshrunken policy, the performance of our PSGD based policy is always observed to be superior for moderately large $T$. The relative performance at higher imbalances is still very high. 

We also report the performance of our policy in these imbalanced design compared to an unshrunken policy in a balanced design regime. We compute the regret of the unshrunken policy in the balanced setting as $n_t$ varies and compare with the efficacy of the prescribed PSGD based policy in the imbalanced setting in Table~\ref{tab:bal_comp}. We see that with this imbalanced design our policy still outperforms the unshrunken policy in a balanced design regime.

\begin{table}[htbp]
  \centering
  \caption{Performance of the prescribed method in imbalanced designs relative to Unshrunken policy in balanced design in Set-up 4 as $\upsilon_1(1-\upsilon_1)$ varies across columns. 
  (negative implies worse performance)\\}
    \begin{tabular}{|c|r|rrr|}
    \toprule
    \textbf{n} & \multicolumn{1}{c|}{\textbf{T}} & \multicolumn{1}{c|}{\textbf{0.7}} & \multicolumn{1}{c|}{\textbf{0.8}} & \multicolumn{1}{c|}{\textbf{0.9}} \\\midrule
    \multirow{4}[0]{*}{\textbf{1000}} & \textbf{100} & -6.6\% & -7.6\% & -16.3\% \\
          & \textbf{500} & -1.9\% & -9.2\% & -22.0\% \\
          & \textbf{1000} & 10.5\% & 6.1\% & -6.4\% \\
          & \textbf{5000} & 47.9\% & 44.9\% & 35.8\% \\\midrule
    \multirow{4}[0]{*}{\textbf{2500}} & \textbf{100} & -8.3\% & -18.2\% & -35.9\% \\
          & \textbf{500} & -2.9\% & -12.8\% & -32.1\% \\
          & \textbf{1000} & 12.8\% & 4.1\% & -13.6\% \\
          & \textbf{5000} & 48.7\% & 43.4\% & 31.8\% \\\midrule
    \multirow{4}[0]{*}{\textbf{5000}} & \textbf{100} & -7.6\% & -20.1\% & -40.1\% \\
          & \textbf{500} & 0.1\% & -19.1\% & -36.9\% \\
          & \textbf{1000} & 13.7\% & -2.4\% & -19.1\% \\
          & \textbf{5000} & 49.8\% & 38.4\% & 28.8\% \\\midrule
    \multirow{4}[0]{*}{\textbf{10000}} & \textbf{100} & -7.8\% & -128.6\% & -36.1\% \\
          & \textbf{500} & -2.1\% & -127.6\% & -33.7\% \\
          & \textbf{1000} & 11.9\% & -95.8\% & -17.2\% \\
          & \textbf{5000} & 48.8\% & -16.8\% & 29.9\% \\\midrule
    \multirow{4}[1]{*}{\textbf{20000}} & \textbf{100} & -7.2\% & -16.5\% & -39.3\% \\
          & \textbf{500} & -0.5\% & -16.6\% & -36.4\% \\
          & \textbf{1000} & 13.0\% & -0.1\% & -19.0\% \\
          & \textbf{5000} & 49.3\% & 40.3\% & 29.1\% \\
    \bottomrule
    \end{tabular}%
  \label{tab:bal_comp}%
\end{table}%

\begin{figure}[!htb]
\centering
{\includegraphics[width=0.7\textwidth]{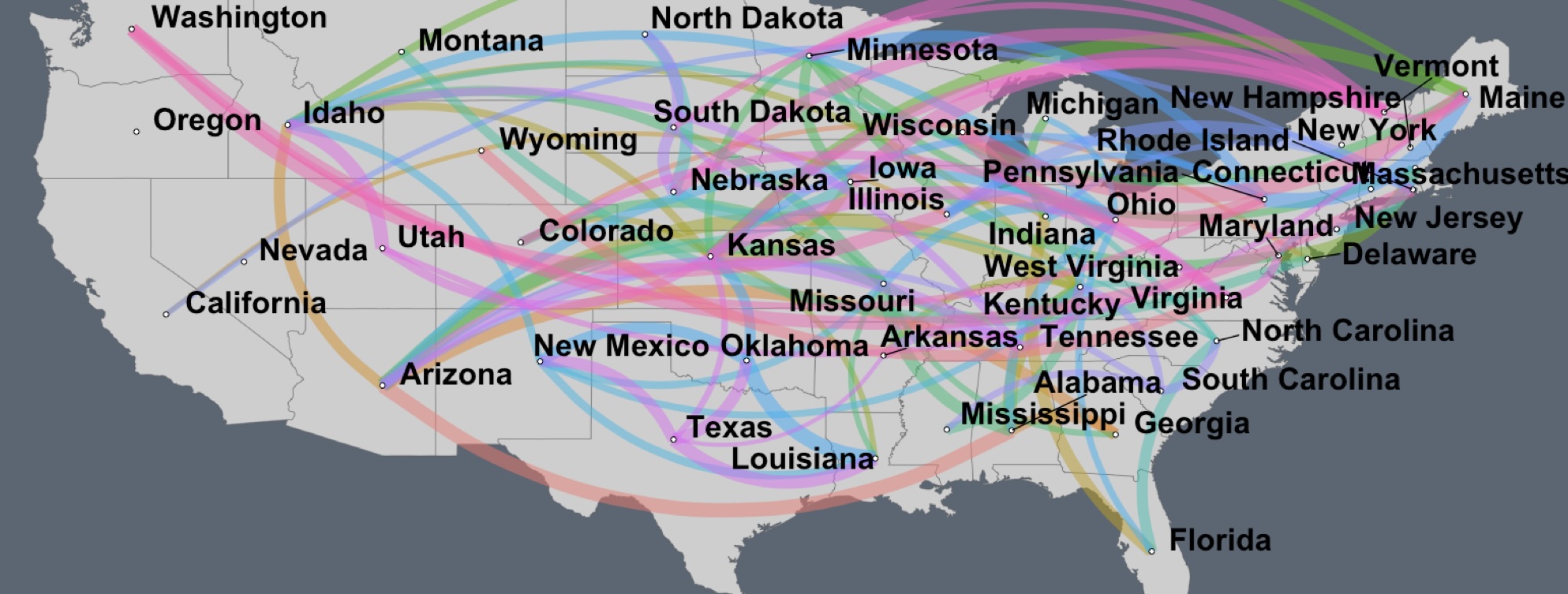}}
{\includegraphics[width=\textwidth]{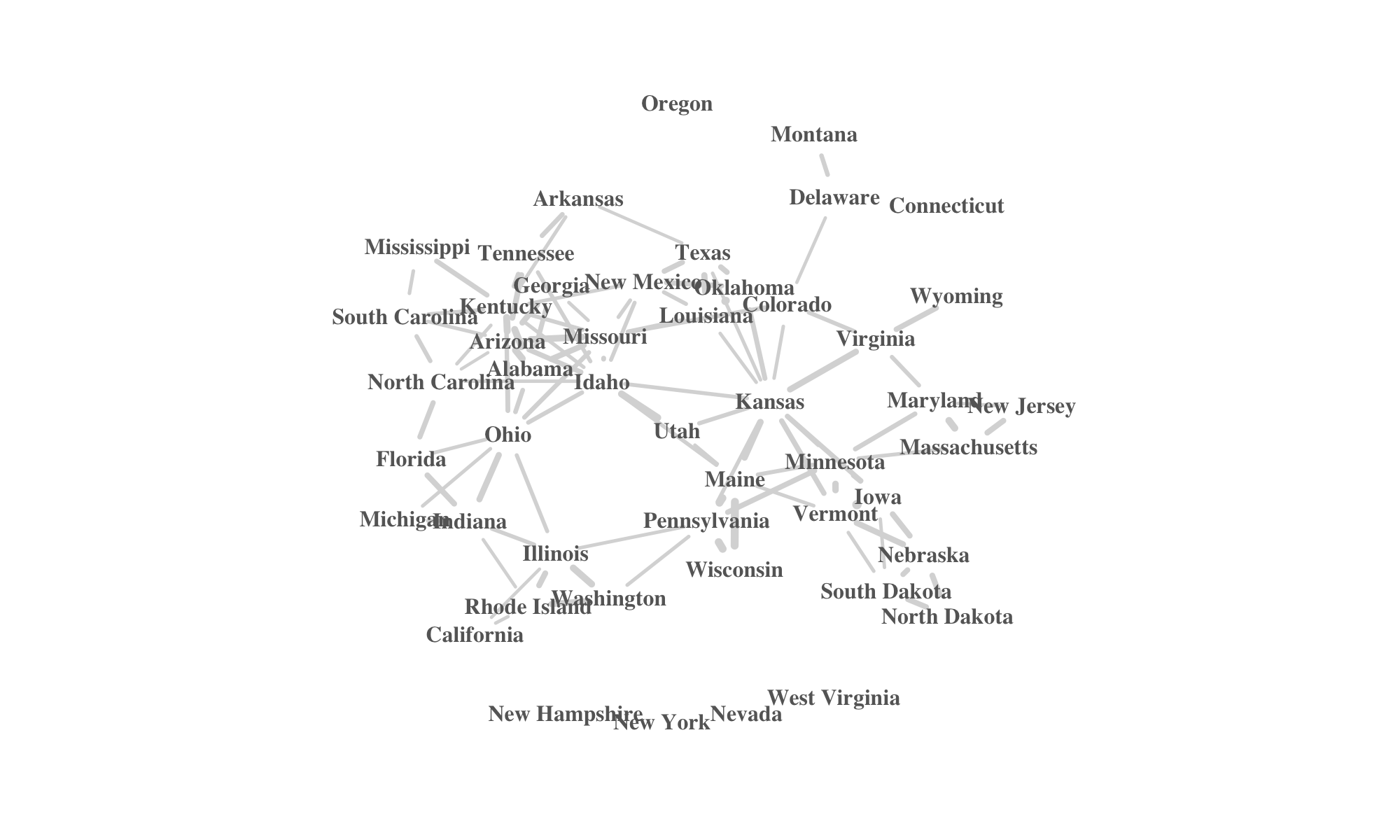}}

    \caption{A network on 48 US states (barring Hawaii and Alaska) based on demographic variables. It is used in set-up 5.}
    \label{fig:map_dem}
\end{figure}

\textbf{Set-up 5.} Next, we want to study the regret behaviour under a different network on the segments. Unlike Set-ups 3 and 4, we create a network on the US states based only on the demographic variables. Based on the demographic variables, we create the similarity matrix $\bm W$ using an RBF kernel of width two and threshold the edges at $0.05$. Figure~\ref{fig:map_dem} shows the network. 

\begin{table}[!htbp]
  \centering
  \caption{Performance of PSGD in Set-up 5 relative to (a) unshrunken policy in the balanced designs and (b) unshrunken policy in the imbalanced designs (negative implies worse performance)}
     \scalebox{0.9}{\begin{tabular}{|c|r|rr|rr|rr|}
    \toprule
    \multirow{2}[4]{*}{\textbf{n}} & \multicolumn{1}{c|}{\multirow{2}[4]{*}{\textbf{T}}} & \multicolumn{2}{c|}{\textbf{0.7}} & \multicolumn{2}{c|}{\textbf{0.8}} & \multicolumn{2}{c|}{\textbf{0.9}} \\
\cmidrule{3-8}          &       & \multicolumn{1}{c|}{\textbf{Balanced }} & \multicolumn{1}{c|}{\textbf{Imbalanced}} & \multicolumn{1}{c|}{\textbf{Balanced }} & \multicolumn{1}{c|}{\textbf{Imbalanced}} & \multicolumn{1}{c|}{\textbf{Balanced }} & \multicolumn{1}{c|}{\textbf{Imbalanced}} \\
    \midrule
    \multirow{4}[8]{*}{\textbf{1000}} & \textbf{100} & -9.1\% & -12.4\% & -8.2\% & -21.7\% & -6.1\% & -34.0\% \\
\cmidrule{2-2}          & \textbf{500} & -13.5\% & -16.9\% & -18.0\% & -35.5\% & -20.8\% & -59.1\% \\
\cmidrule{2-2}          & \textbf{1000} & -1.0\% & -4.1\% & -3.2\% & -20.5\% & -4.4\% & -42.7\% \\
\cmidrule{2-2}          & \textbf{5000} & 54.8\% & 53.4\% & 54.3\% & 50.0\% & 55.7\% & 49.8\% \\
    \midrule
    \multirow{4}[8]{*}{\textbf{2500}} & \textbf{100} & -5.4\% & -8.6\% & -2.1\% & -21.1\% & 3.2\% & -32.5\% \\
\cmidrule{2-2}          & \textbf{500} & -13.8\% & -17.2\% & -10.2\% & -27.4\% & -10.8\% & -53.6\% \\
\cmidrule{2-2}          & \textbf{1000} & -3.9\% & -7.0\% & 0.5\% & -13.8\% & 0.9\% & -40.0\% \\
\cmidrule{2-2}          & \textbf{5000} & 45.1\% & 43.4\% & 48.7\% & 49.9\% & 51.5\% & 48.8\% \\
    \midrule
    \multirow{4}[8]{*}{\textbf{5000}} & \textbf{100} & -8.0\% & -11.2\% & -7.3\% & -19.3\% & -2.3\% & -31.3\% \\
\cmidrule{2-2}          & \textbf{500} & -15.6\% & -19.0\% & -15.9\% & -27.9\% & -13.2\% & -45.5\% \\
\cmidrule{2-2}          & \textbf{1000} & -3.8\% & -6.9\% & -3.3\% & -13.9\% & 0.7\% & -30.3\% \\
\cmidrule{2-2}          & \textbf{5000} & 46.8\% & 45.2\% & 48.3\% & 49.7\% & 52.2\% & 52.3\% \\
    \midrule
    \multirow{4}[8]{*}{\textbf{10000}} & \textbf{100} & -7.9\% & -11.2\% & -4.3\% & -22.1\% & 0.6\% & -30.0\% \\
\cmidrule{2-2}          & \textbf{500} & -12.4\% & -15.8\% & -12.0\% & -30.1\% & -10.4\% & -47.9\% \\
\cmidrule{2-2}          & \textbf{1000} & -0.4\% & -3.4\% & 1.3\% & -15.7\% & 3.3\% & -34.5\% \\
\cmidrule{2-2}          & \textbf{5000} & 48.5\% & 47.0\% & 51.0\% & 50.6\% & 54.0\% & 51.5\% \\
    \midrule
    \multirow{4}[8]{*}{\textbf{20000}} & \textbf{100} & -6.9\% & -10.1\% & -2.4\% & -21.0\% & 1.8\% & -31.3\% \\
\cmidrule{2-2}          & \textbf{500} & -13.9\% & -17.3\% & -10.4\% & -29.6\% & -10.8\% & -49.6\% \\
\cmidrule{2-2}          & \textbf{1000} & -2.1\% & -5.2\% & 1.9\% & -16.0\% & 2.2\% & -36.2\% \\
\cmidrule{2-2}          & \textbf{5000} & 46.2\% & 44.6\% & 49.9\% & 48.8\% & 52.5\% & 50.6\% \\
    \bottomrule
    \end{tabular}%
  \label{tab:demo_rel_reg}
  }
\end{table}%

We study the regret of the imbalanced design in this network regime. Similar to the set-up 4, we compare our policy in the imbalanced setting against (a) unshrunken policy in the balanced setup and (b) unshrunken policy in the imbalanced setting. The results are presented in Table~\ref{tab:demo_rel_reg}. At smaller $T$'s our policy performs worse than an unshrunken policy, but as $T$ grows larger, our policy performs significantly better (more than $50\%$) compared to the unshrunken policy, both with balanced and imbalanced design.

\begin{figure}[!htb]
\centering
{\includegraphics[width=0.7\textwidth]{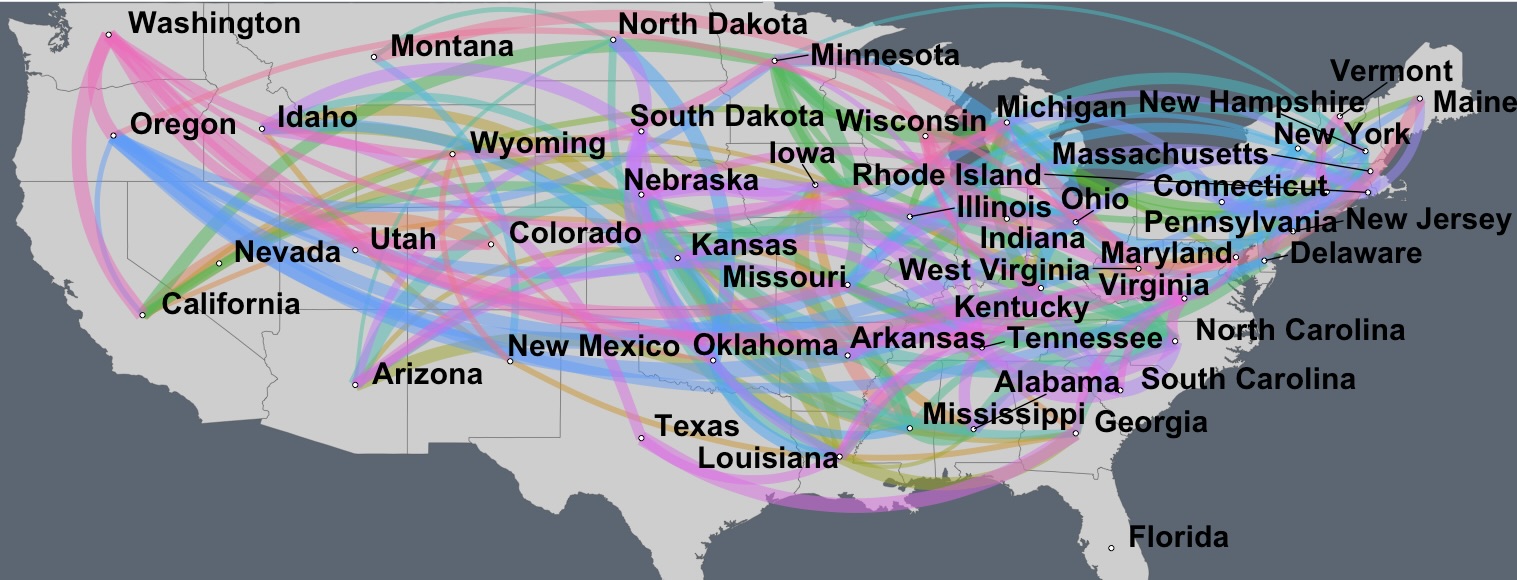}}
{\includegraphics[width=0.8\textwidth]{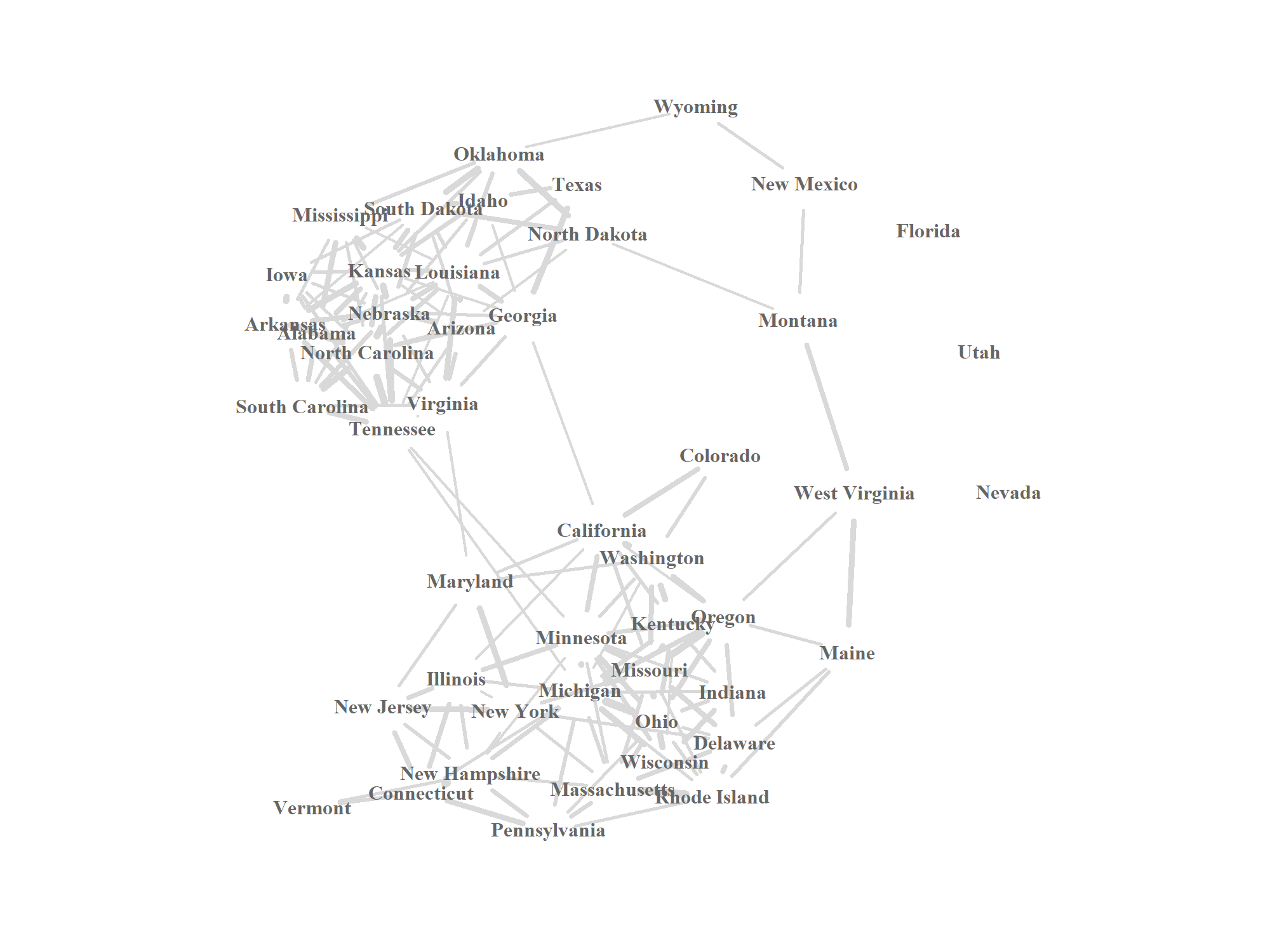}}

\caption{A network on 48 US states (barring Hawaii and Alaska) based on economic variables. It is used in set-up 6.}
    \label{fig:map_eco}
\end{figure}

\textbf{Set-up 6.} We use a network that is based on the similarity across economic variables only.  We create the network of the US states based only on the economic variables using an RBF kernel of width two and thresholding the edges at $0.05$.  Figure~\ref{fig:map_eco} shows the network.

We study the performance under this network regime in the imbalanced setting with the two scenarios as above. The results are reported in Table~\ref{tab:eco_rel_reg}. We see that overall, in all imbalanced settings our policy performs much better than the unshrunken policy in the imbalanced as well as the balanced setting.

\begin{table}[!htbp]
  \centering
  \caption{Performance of prescribed method in Set-up 6 compared to (a) unshrunken policy in the balanced setup and (b) unshrunken policy in the imbalanced setup (negative implies worse performance)}
    \scalebox{0.7}{\begin{tabular}{|c|r|rr|rr|rr|}
    \toprule
    \multirow{2}[4]{*}{\textbf{n}} & \multicolumn{1}{c|}{\multirow{2}[4]{*}{\textbf{T}}} & \multicolumn{2}{c|}{\textbf{0.7}} & \multicolumn{2}{c|}{\textbf{0.8}} & \multicolumn{2}{c|}{\textbf{0.9}} \\
\cmidrule{3-8}          &       & \multicolumn{1}{c|}{\textbf{Balanced }} & \multicolumn{1}{c|}{\textbf{Imbalanced}} & \multicolumn{1}{c|}{\textbf{Balanced }} & \multicolumn{1}{c|}{\textbf{Imbalanced}} & \multicolumn{1}{c|}{\textbf{Balanced }} & \multicolumn{1}{c|}{\textbf{Imbalanced}} \\
    \midrule
    \multirow{4}[8]{*}{\textbf{1000}} & \textbf{100} & -4.2\% & -2.1\% & -7.3\% & -2.3\% & -4.3\% & -3.8\% \\
\cmidrule{2-2}          & \textbf{500} & 4.9\% & 6.8\% & 2.0\% & 6.6\% & 3.6\% & 4.3\% \\
\cmidrule{2-2}          & \textbf{1000} & 10.7\% & 12.5\% & 8.8\% & 12.6\% & 10.3\% & 10.3\% \\
\cmidrule{2-2}          & \textbf{5000} & 29.3\% & 30.7\% & 27.7\% & 30.8\% & 29.1\% & 29.2\% \\
    \midrule
    \multirow{4}[8]{*}{\textbf{2500}} & \textbf{100} & -6.2\% & -4.0\% & -6.2\% & -3.0\% & -9.4\% & -5.9\% \\
\cmidrule{2-2}          & \textbf{500} & 4.8\% & 6.7\% & 3.4\% & 6.9\% & -1.1\% & 3.0\% \\
\cmidrule{2-2}          & \textbf{1000} & 11.7\% & 13.5\% & 10.1\% & 12.9\% & 6.4\% & 9.4\% \\
\cmidrule{2-2}          & \textbf{5000} & 30.8\% & 32.2\% & 28.9\% & 31.2\% & 25.9\% & 28.5\% \\
    \midrule
    \multirow{4}[8]{*}{\textbf{5000}} & \textbf{100} & -7.4\% & -5.2\% & -6.1\% & -2.5\% & -9.8\% & -5.6\% \\
\cmidrule{2-2}          & \textbf{500} & 3.7\% & 5.6\% & 3.4\% & 6.9\% & -1.0\% & 3.0\% \\
\cmidrule{2-2}          & \textbf{1000} & 10.2\% & 12.0\% & 10.2\% & 13.0\% & 6.2\% & 9.5\% \\
\cmidrule{2-2}          & \textbf{5000} & 29.9\% & 31.3\% & 29.1\% & 31.4\% & 25.9\% & 28.7\% \\
    \midrule
    \multirow{4}[8]{*}{\textbf{10000}} & \textbf{100} & -6.5\% & -4.4\% & -2.4\% & -2.8\% & -4.7\% & -6.1\% \\
\cmidrule{2-2}          & \textbf{500} & 4.5\% & 6.4\% & 7.0\% & 6.7\% & 3.7\% & 2.9\% \\
\cmidrule{2-2}          & \textbf{1000} & 11.0\% & 12.8\% & 13.4\% & 12.8\% & 10.8\% & 9.4\% \\
\cmidrule{2-2}          & \textbf{5000} & 30.0\% & 31.4\% & 31.4\% & 31.0\% & 29.8\% & 28.9\% \\
    \midrule
    \multirow{4}[8]{*}{\textbf{20000}} & \textbf{100} & -5.8\% & -3.7\% & -3.5\% & -2.8\% & -6.8\% & -5.7\% \\
\cmidrule{2-2}          & \textbf{500} & 4.6\% & 6.5\% & 6.1\% & 6.9\% & 1.6\% & 3.0\% \\
\cmidrule{2-2}          & \textbf{1000} & 11.0\% & 12.8\% & 12.7\% & 13.1\% & 8.7\% & 9.3\% \\
\cmidrule{2-2}          & \textbf{5000} & 30.2\% & 31.6\% & 31.0\% & 31.4\% & 28.0\% & 28.6\% \\
    \bottomrule
    \end{tabular}%
  \label{tab:eco_rel_reg}%
  }
\end{table}%

\textbf{Set-up 7.} With the $\bm{W}$ used in Set-ups $3$ and $4$, we setup an extremely unbalanced design with total number of customers fixed at $n_t=1000$. In this setting, $10$ states have $5$ leads each, while the remaining $950$ leads are distributed similarly among the remaining 38 states. We specifically study the regret from the $10$ states with very low leads.  

\begin{figure}[!htb]
    $\qquad$
    {\includegraphics[width=0.4\textwidth]{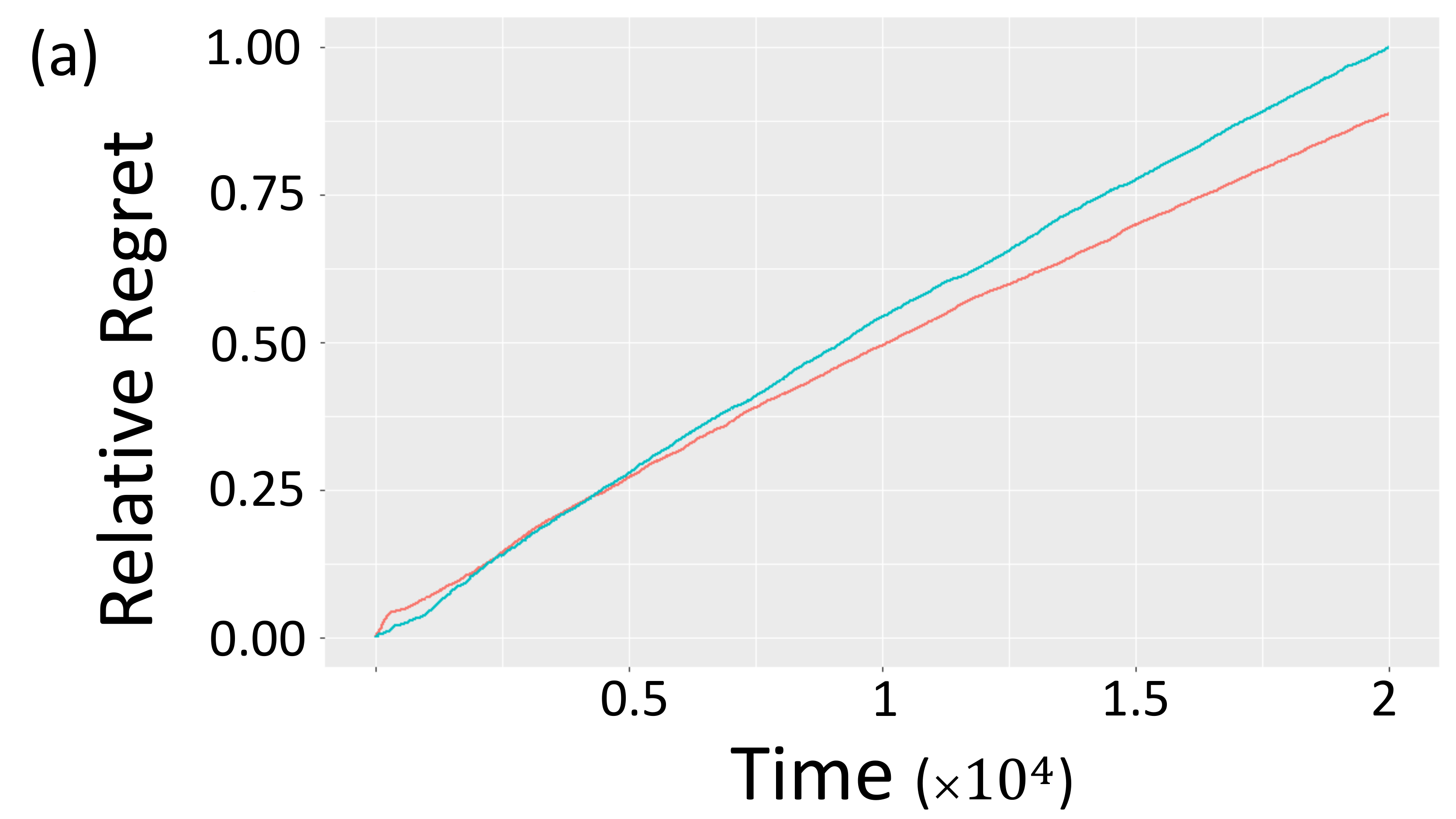}}$\qquad$
    {\includegraphics[width=0.4\textwidth]{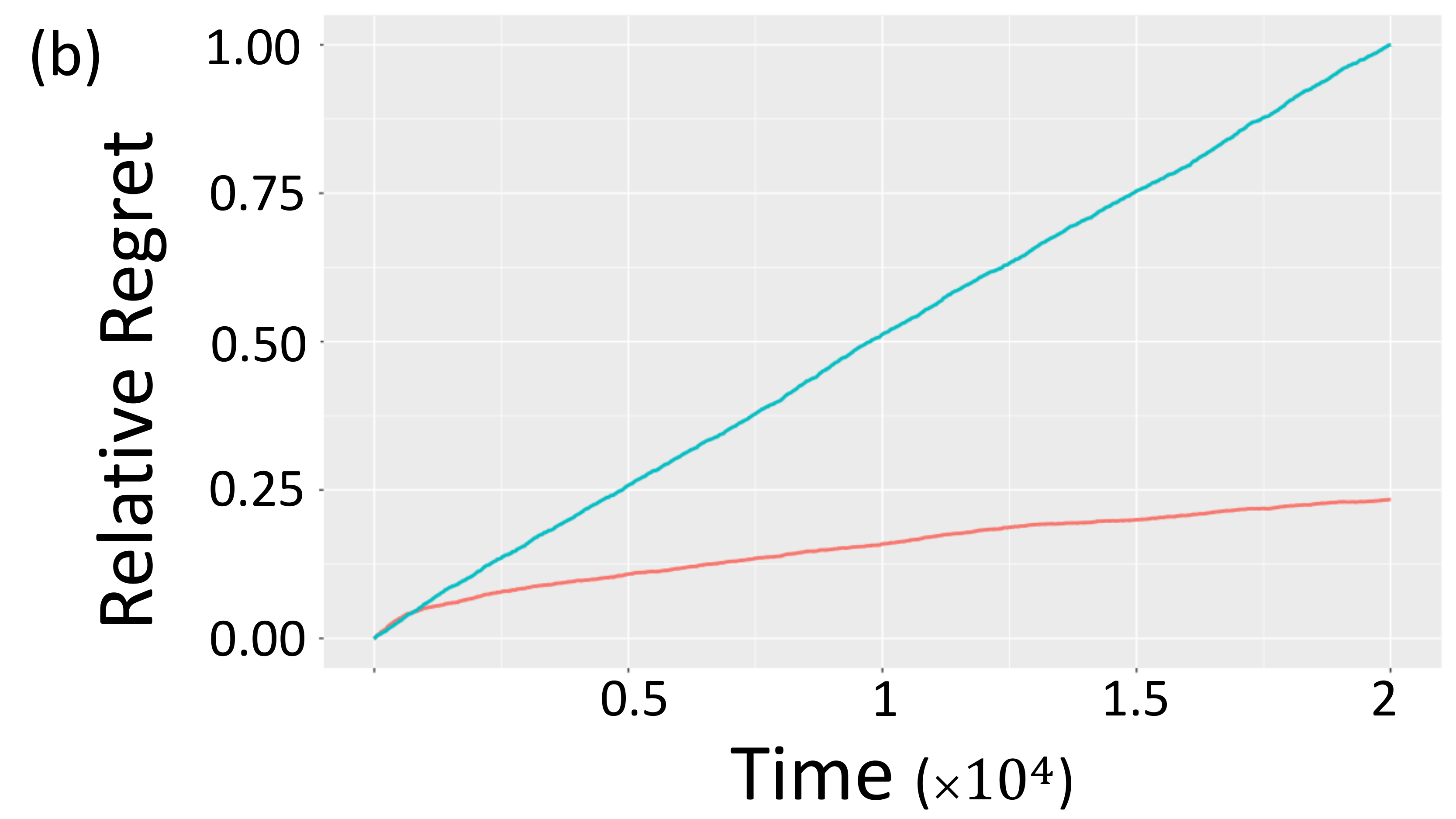}} 
    \caption{Relative regret of the prescribed policy (red) compared to unshrunken policy (green) in Set-up 7. Panel (a) corresponds to case (a) where the very low lead states are not very well connected with other states. Panel (b) corresponds to case (b) where the very low lead states are very well connected with other states.}
    \label{fig:sparse_states}
\end{figure}

Consider the two cases here (a) the ten states with very-low-leads are chosen such that they are least connected states (sum of the edge weights is least) in the network (b)  the ten states with very-low-leads are chosen such that they are are the most connected 10 states in the network. In Figure~\ref{fig:sparse_states}, we plot the relative regret of the prescribed policy from the very-low-leads states with respect to an unshrunken pricing policy. 

From the figure we see that in the case (b), the prescribed policy pools information from the other connected states and perform much better compared to an unshrunken policy. On the other hand in case (a) as the low-lead-states are not very well connected with the other states, the prescribed policy perform on a similar level compared to the unshrunken policy and does not provide any benefit. 

\textbf{Set-up 8.}
Throughout our experiments so far, we have assumed that the noises are Gaussian. Here, we study the performance of our proposed algorithm when the noise in model (2) follows Laplace distribution.

We consider the same synthetic data regime as in setup-1. We have $L=10$ segments where for any $t\geq 1$, $n_{lt}=50$ for $l=1,\ldots,5$ and $n_{lt}=200$ for $l=6,\ldots,10$. The network matrix $\bm W$ is generated using an RBF kernel of width one on independent standard Gaussian feature vectors, drawn from input space $\mathbb{R}^{10}$. 
 The covariates $\bm x_{lt}$ are generated from standard exponential distribution and set $\upsilon_t$ to $0.5$ for all $t\geq 1$. The price sensitivity and the customer preferences are assumed as $\beta_1 = -0.4$ and $\bm\mu_1 = (0.1, 0.15)$. We simply generate $Z_{lt}$ in ~\eqref{eq:2} as standard Laplace observations rather than Gaussian observations.

\begin{figure}[!htb]
    $\qquad$
    {\includegraphics[width=0.4\textwidth]{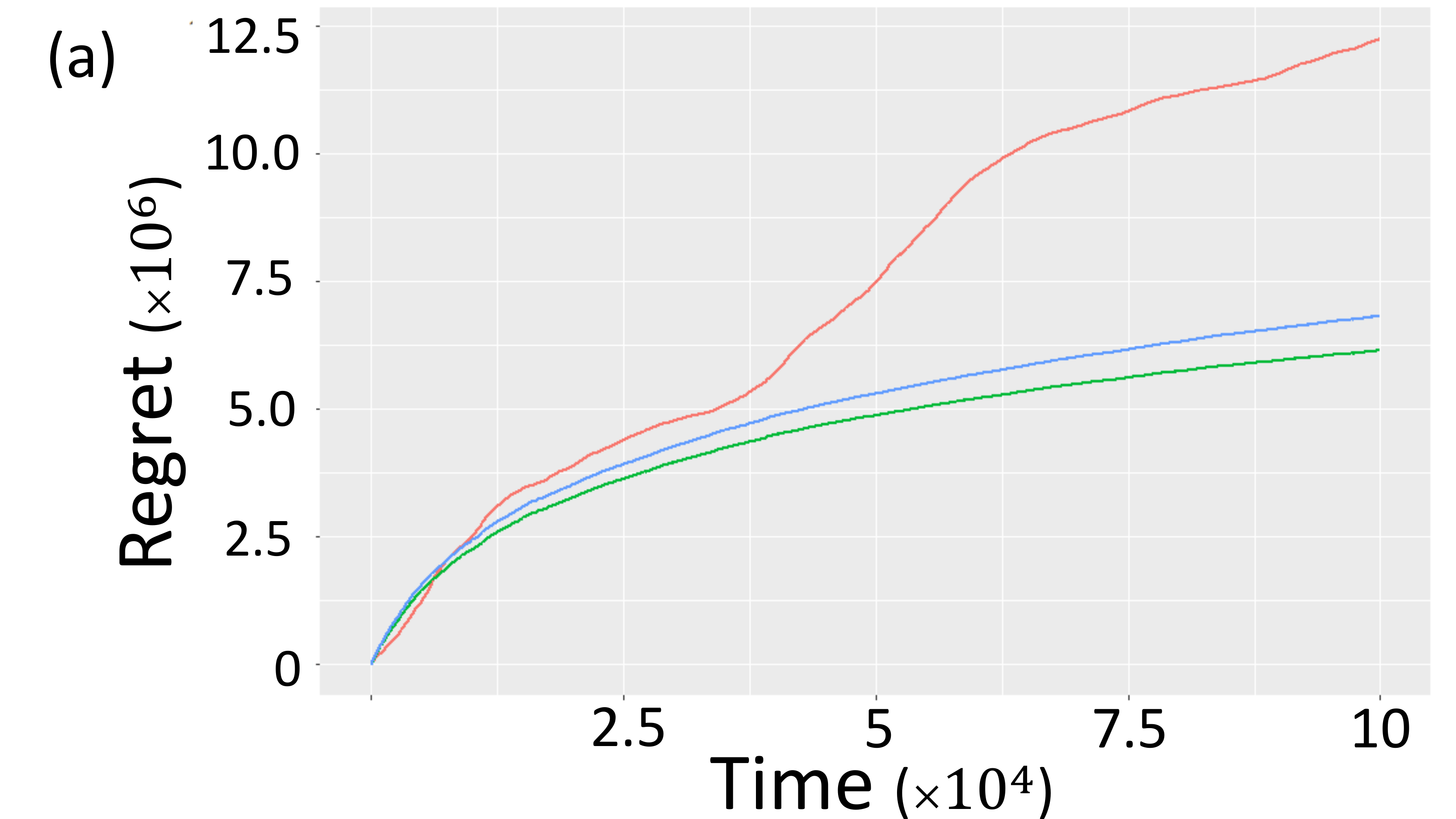}}$\qquad$
    {\includegraphics[width=0.4\textwidth]{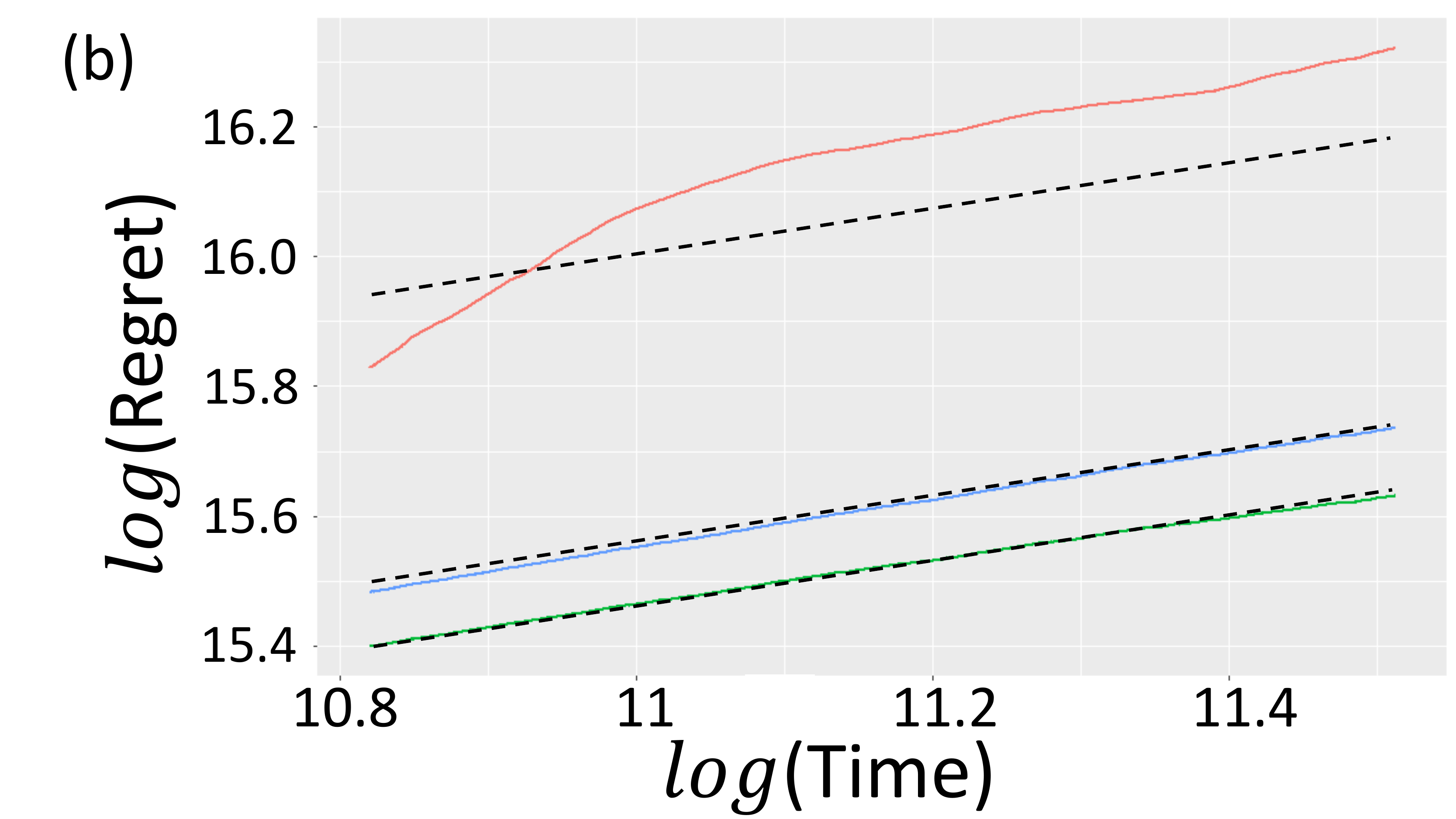}} 
    \caption{Regret plots of the proposed policy in Set-up 8 as $b$ which governs the shift in the model parameters over time changes. The plots for $b=0.5, 1, \infty$ are in red, green and blue respectively. In panels (a) Regret in original scale (b) log(Regret) vs log(T). The dotted lines in panel (b) are the best fitted line with slope 0.5.}
    \label{fig:regret_b_vary_laplace}
\end{figure}

  We study effects of the temporal variations across $\beta_t$ and $\bm \mu_t$ by varying $b$ as before. Recall that the case of $b=\infty$ corresponds to the scenario where the parameters do not change over time.  Since the noise distribution is Laplace (log-concave), following from Corollary~\ref{cor.2}, the regret for the two cases of $b = 1, \infty$ should be of order $O(\sqrt{T})$, while the regret for $b = 0.5$ should be $\mathcal{O}(T)$. In Figure~\ref{fig:regret_b_vary_laplace}, we plot the regret (cumulative revenue lost to the oracle policy) over time for the three cases. From the figures it is evident that when $b = 0.5$, is linear where as in the other two cases its is controlled at $\mathcal{O}(\sqrt{T})$.  

\begin{figure}[!htbp]
    $\qquad$
    {\includegraphics[width=0.4\textwidth]{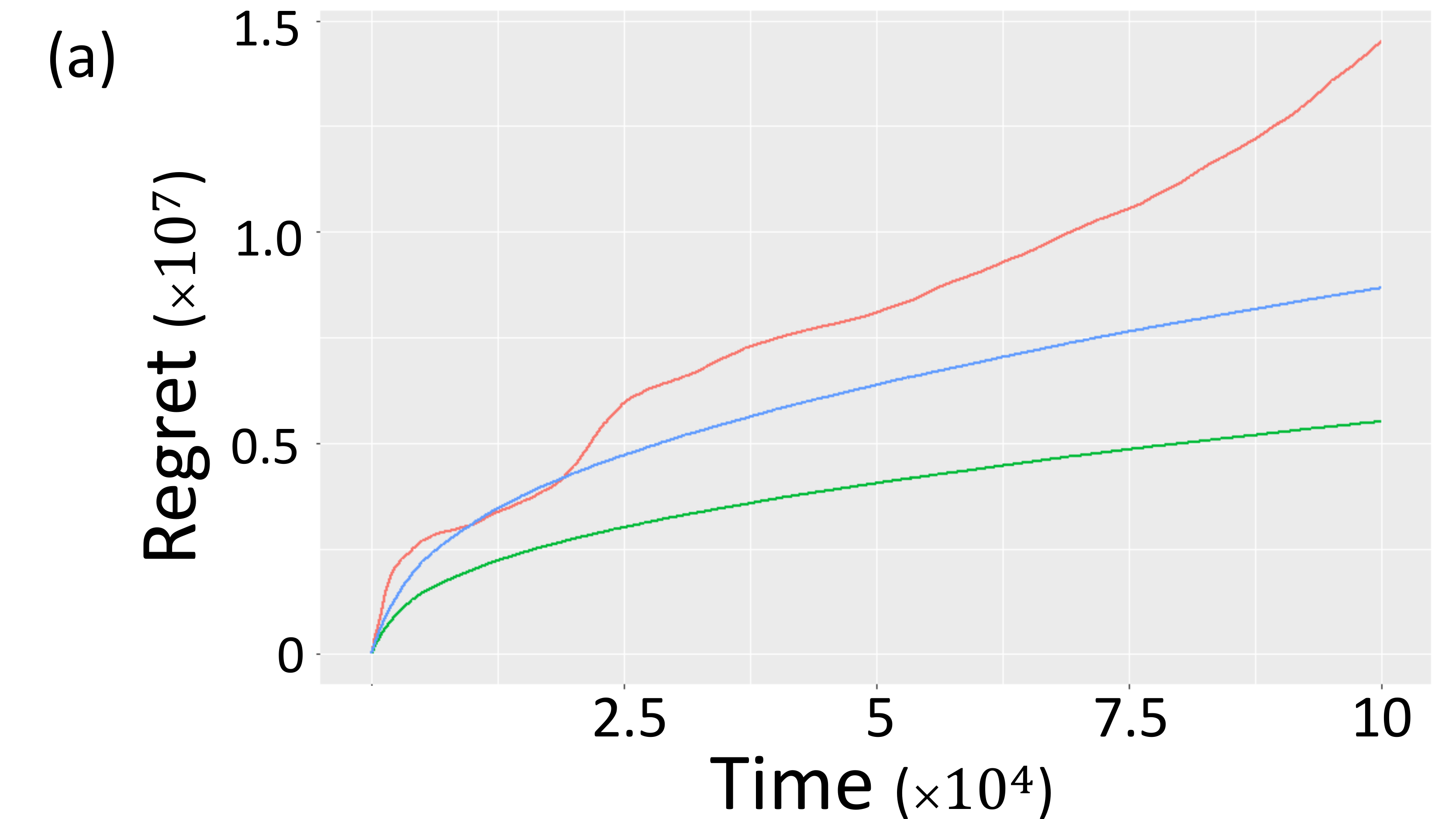}}$\qquad$
    {\includegraphics[width=0.4\textwidth]{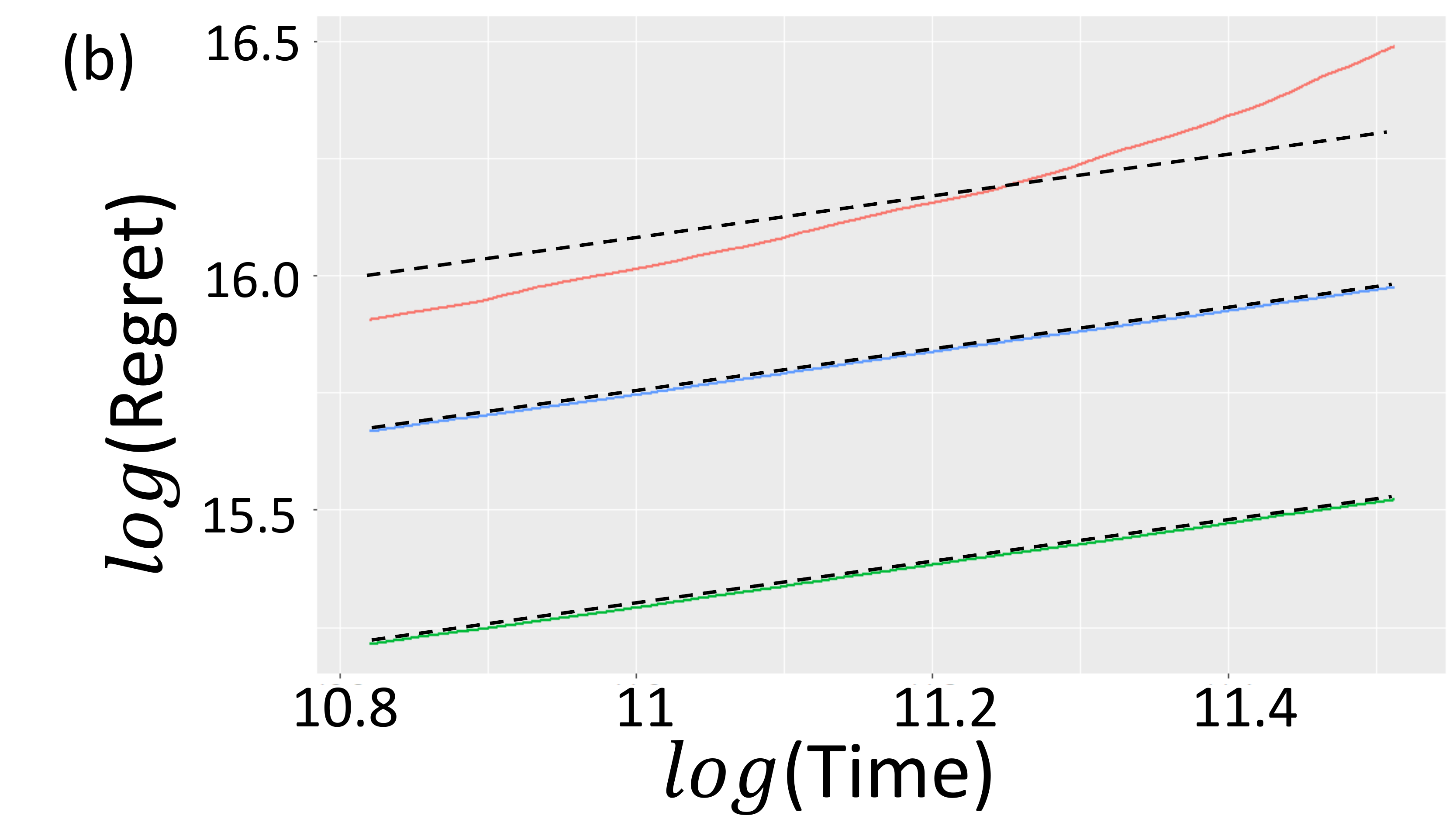}} 
    \caption{Regret plots of the proposed policy in Set-up 9 as $b$ which governs the shift in the model parameters over time changes. The plots for $b=0.5, 1, \infty$ are in red, green and blue respectively. In panels (a) Regret in original scale (b) log(Regret) vs log(T). The dotted lines in panel (b) are the best fitted line with slope 0.5.}
    \label{fig:regret_b_vary_t}
\end{figure}

\textbf{Set-up 9.}
We extend our study to a setup where the noises are i.i.d. from Student's t distribution. The setup of this experiment is same as set-up 8, with the only change being the distribution of the noise terms. Here, noises follow Student's t distribution in place of the Laplace distribution. We study the regret of our proposed policy under this setting by varying $b$. We see  the effect of the temporal variations of parameters in Figure~\ref{fig:regret_b_vary_t}. We see similar growth of regret, where for $b = 1, \infty$, the regret is $\mathcal{O}(\sqrt{T})$ and for $b = 0.5$ the regret grows linearly. While theoretically the results hold for Gaussian noises, these experiments show that our policy is applicable to a broader family of noise distributions.

\section{Discussion and Future Work}
This work studied dynamic pricing strategies in the streaming longitudinal data setting where the goal is to maximize the cumulative profit over time across a large number of customer segments.
We proposed a pricing policy based on penalized stochastic gradient descent and provided regret bounds demonstrating the asymptotic optimality of the proposed policy.
In particular, we showed that our PSGD algorithm controls the regret at the order of $\bigo(\sqrt{T})$ and that for any pricing policy, the Bayes regret cannot be of the lower order of $\bigo(\sqrt{T})$.
Hence, as $T \rightarrow \infty$, the proposed algorithm is asymptotically rate-optimal.
Our results show that for policy planning it is essential to incorporate available structural information as policies given by unshrunken models are highly sub-optimal.

There are several important future directions of our work.
In future work, it will be useful to derive the regret of the proposed algorithm when the noise is non Gaussian, e.g. heavy-tailed distributions as such noise characteristics are often associated with observed demand data. Theoretically, it will be interesting to calculate the benefits of a batched version of the proposed algorithm~1 that is equipped for price exploration within segments though it might not be practically feasible due to spill-over effects.  Also, here we have considered global spatial structure in the form of the SAR model in \eqref{eq:5}. In the future, it will be interesting to study the performance of PSGD in the presence of local shrinkage structures such as geographically weighted regression  models \citep{fotheringham2003geographically}.  Finally, it will be useful to evaluate the optimal regret in time-varying networks where the contiguity matrix $\bm{W}$ also changes over time.

\newpage
\bibliography{Bibliography-MM-MC,robust_learning_bib}
\bibliographystyle{agsm}

\newpage

\appendix

\section{Organization of the Appendix}

Here, we first present the proofs of results discussed in Section 3 of the main paper. The detailed proofs of the main results, Theorems~\ref{thm.1} and \ref{thm.2}, are provided in Section B of the appendix. All the other proofs of the intermediary results is provided in Section C of the Appendix. 
\section{Detailed Proofs of Theorem~\ref{thm.1} and \ref{thm.2}}
\subsection{Proof of Theorem~\ref{thm.1}}

The total regret $\mathcal{B}(\bm\theta,{\bm p})$ can be written down as the sum of regrets over all the segments and time period:

\[\mathcal{B}(\bm\theta,{\bm p}) = \sum_{t = 1}^T\sum_{l = 1}^L \mathcal{R}_{lt}\,.\]

Using Proposition~\ref{prop.2} and Lemma~\ref{lemma.2}, we can bound the regret as 

\[\mathcal{B}(\bm\theta,{\bm p}) = C_9C_{10}\sum_{t = 1}^T\sum_{l = 1}^L n_{lt}\langle\bm{x}_{lt}, \bm m_{lt} - \hat{\bm m}_{lt}\rangle^2 + C_9C_{10}\sum_{t = 1}^T\sum_{l = 1}^L n_{lt}\big(p_{lt}(b_{lt} - \hat{b}_{lt})\big)^2\,.\]

Taking a maximum on the constants, we can rather bound the sum of the two terms  $\sum_{t = 1}^T\sum_{l = 1}^L n_{lt}\langle\bm{x}_{lt}, \bm m_{lt} - \hat{\bm m}_{lt}\rangle^2$ and $\sum_{t = 1}^T\sum_{l = 1}^L n_{lt}\big(p_{lt}(b_{lt} - \hat{b}_{lt})\big)^2$ to get a final bound on the regret.

We do the analysis for a fixed segment $l$ first and then combine the regret across all the segments.  

\begin{lemma}\label{lemma.app.1}
     Consider model \eqref{eq:12.1}, true parameters $ \bm{m}_{lt}$, $ b_{lt}$ and the output $ \hat{\bm{m}}_{lt}$, $ \hat{b}_{lt}$ from our PSGD pricing policy, the following holds with probability at least $1 - T^{-2}$ 
     \begin{align*}
             &\sum_{t = 1}^T n_{lt} \Big(\langle\bm{x}_{lt}, \bm m_{lt} - \hat{\bm m}_{lt}\rangle^2 + \big(p_{lt}(b_{lt} - \hat{b}_{lt})\big)^2\Big)\\
             &\leq  {C}_1\sum_{t = 1}^T \frac{1}{\eta_t} \|\bm m_{l,t+1} - \bm m_{lt}\|_2 + {C}_2\sum_{t = 1}^T \frac{1}{\eta_t} |b_{l,t+1} - b_{lt}|
+ {C}_3\sum_{t = 1}^T{\eta_t} n_{lt}^2 + \frac{{C}_4}{\eta_{T+1}} + \mathcal{O}(\log T)\,.
     \end{align*}
\end{lemma}

With this lemma we have with probability at least $1 -L/T^2$, 
\begin{align*}
    &\sum_{t = 1}^T \sum_{l = 1}^L n_{lt} \Big(\langle\bm{x}_{lt}, \bm m_{lt} - \hat{\bm m}_{lt}\rangle^2 + \big(p_{lt}(b_{lt} - \hat{b}_{lt})\big)^2\Big)\\
    &\leq {C}_1\sum_{t = 1}^T \sum_{l = 1}^L\frac{1}{\eta_t} \|\bm m_{l,t+1} - \bm m_{lt}\|_2 + {C}_2\sum_{t = 1}^T \sum_{l = 1}^L\frac{1}{\eta_t} |b_{l,t+1} - b_{lt}|
    + {C}_3\sum_{t = 1}^T\sum_{l = 1}^L{\eta_t} n_{lt}^2 + \frac{{C}_4L}{\eta_{T+1}} + \mathcal{O}(\log T)\,.
\end{align*}

Define the RHS as $I$.

Consider $\mathcal{G}$ to be the probabilistic event that the above is true, then $\mathbb P (\mathcal{G}^C) = L/T^2$.

Also, since the maximum price is $M$ and we set a positive price, hence the maximum revenue lost on the event $\mathcal{G}^C$ is $\sum_{t = 1}^T n_{t}M$. Assuming that $N_T = \max_{t\leq T}n_T$, we have the maximum regret in the event $\mathcal{G}^C$ is $T M N_T$.

The total regret is thus,

\[\mathcal{B}(\bm\theta,{\bm p}) = \mathcal{B}(\bm\theta,{\bm p}|\mathcal{G}) + \mathcal{B}(\bm\theta,{\bm p}|\mathcal{G}^C)\leq I \mathbb P(\mathcal{G}) + T N_T \mathbb P(\mathcal{G}^C) \leq A + \frac{ML N_T}{T}\,.\]

Since the last term is $ \mathcal{O}(1/T)$, we have the required terms of the regret bound.

\subsubsection{Proof of Lemma~\ref{lemma.app.1}}
Let ${\bm{\psi}}_{lt} = (\bm m_{lt}, b_{lt})$ be the combined parameter space and $\bm{Q}_{lt} = (\bm x_{lt}, p_{lt})$ be the covariates and the price posted.

     By Taylor expansion of the loss function we get, for some $\Tilde{\bm{\psi}}_{l,t}$ between $\hat{\bm{\psi}}_{l,t}$ and $\bm{\psi}_{lt}$,
     \begin{equation}\label{eq:12}
         \mathcal{L}_{lt}(\hat{\bm{\psi}}_{lt}) - \mathcal{L}_{lt}(\bm{\psi}_{lt}) = \inner{\nabla \mathcal{L}_{lt}(\hat{\bm{\psi}}_{lt})}{\hat{\bm{\psi}}_{lt} - \bm{\psi}_{lt}} - \frac{1}{2}\inner{\hat{\bm{\psi}}_{lt} - \bm{\psi}_{lt}}{\nabla^2 \mathcal{L}_{lt}(\Tilde{\bm{\psi}}_{l,t})(\hat{\bm{\psi}}_{lt} - \bm{\psi}_{lt})}\,.
     \end{equation}

Simplifying the loss function in terms of $\bm{\psi}_{lt}$ and $\bm{Q}_{lt}$ gives us $$\mathcal{L}_{lt}(\bm{\psi}) = - \left( y_{lt} \log \Phi(\bm{Q}_{lt}\bm{\psi}) + \tilde{ y}_{lt}\log {\Phi}(-\bm{Q}_{lt}\bm{\psi})\right),$$ where $\tilde{ y}_{lt} = n_{lt} - y_{lt}$.
The second derivative of the loss function can thus be computed as 
     \[\nabla^2 \mathcal{L}_{lt}(\bm{\psi}) = - \left(y_{lt}\bm{Q}_{lt}\bm{Q}_{lt}^T \frac{\partial}{\partial \zeta^2}\log \Phi(\zeta)|_{\zeta = \bm{Q}_{lt}\bm{\psi}} + \tilde{ y}_{lt}\bm{Q}_{lt}\bm{Q}_{lt}^T \frac{\partial}{\partial \zeta^2}\log \Phi(\zeta)|_{\zeta = -\bm{Q}_{lt}\bm{\psi}}\right)\,.\]
     
     Let $c_\mathcal{L} = \min\left\{-\frac{\partial}{\partial \zeta^2}\log \Phi(\zeta)|_{\zeta = \bm{Q}_{lt}\bm{\psi}}, -\frac{\partial}{\partial \zeta^2}\log \Phi(\zeta)|_{\zeta = -\bm{Q}_{lt}\bm{\psi}}\right\}$.
     Based on our assumptions, $\bm{Q}_{lt}$ and $\bm{\psi}$ are bounded and so there exists $c$ such that $|\bm{Q}_{lt}\bm{\psi}|<c$. Since $\Phi$ is log-concave hence the second derivative is negative and $-\frac{\partial}{\partial \zeta^2}\log \Phi(\zeta) > 0$. Particularly since the second derivative only approaches $0$ when $\zeta$ goes to $\infty$ or $-\infty$, hence on the bounded set with $|\zeta| < c$, second derivative is bounded away from zero implying $c_{\mathcal{L}}>0$.
     
     Then,
     \begin{align*}
         \nabla^2 \mathcal{L}_{lt}(\bm{\psi}) &= - \left(y_{lt}\bm{Q}_{lt}\bm{Q}_{lt}^T \frac{\partial}{\partial \zeta^2}\log \Phi(\zeta)|_{\zeta = \bm{Q}_{lt}\bm{\psi}} + \tilde{ y}_{lt}\bm{Q}_{lt}\bm{Q}_{lt}^T\frac{\partial}{\partial \zeta^2}\log \Phi(\zeta)|_{\zeta = -\bm{Q}_{lt}\bm{\psi}}\right)\\
         &= \left(y_{lt}\bm{Q}_{lt}\bm{Q}_{lt}^T \left(-\frac{\partial}{\partial \zeta^2}\log \Phi(\zeta)|_{\zeta = \bm{Q}_{lt}\bm{\psi}}\right) + \tilde{ y}_{lt}\bm{Q}_{lt}\bm{Q}_{lt}^T \left(-\frac{\partial}{\partial \zeta^2}\log \Phi(\zeta)|_{\zeta = -\bm{Q}_{lt}\bm{\psi}}\right)\right)\\
         &\geq (y_{lt} + \tilde{ y}_{lt})\bm{Q}_{lt}\bm{Q}_{lt}^T \min\left\{-\frac{\partial}{\partial \zeta^2}\log \Phi(\zeta)|_{\zeta = \bm{Q}_{lt}\bm{\psi}}, -\frac{\partial}{\partial \zeta^2}\log \Phi(\zeta)|_{\zeta = -\bm{Q}_{lt}\bm{\psi}}\right\}\\
         &= n_{lt}\bm{Q}_{lt}\bm{Q}_{lt}^T c_{\mathcal{L}}\,.
     \end{align*}
     Where the last equality follows since  $y_{lt} + \tilde{ y}_{lt} = n_{lt}$.
    
    Using this in \eqref{eq:12}

     \begin{align}\label{eq:28}
     \begin{split}
          \mathcal{L}_{lt}(\hat{\bm{\psi}}_{lt}) - \mathcal{L}_{lt}(\bm{\psi}_t) &\leq \inner{\nabla \mathcal{L}_{lt}(\hat{\bm{\psi}}_{lt})}{\hat{\bm{\psi}}_{lt} - \bm{\psi}_{lt}} - \frac{1}{2}\inner{\hat{\bm{\psi}}_{lt} - \bm{\psi}_{lt}}{n_{lt}c_\mathcal{L} \bm{Q}_{lt}\bm{Q}_{lt}^T(\hat{\bm{\psi}}_{lt} - \bm{\psi}_{lt})}\\
          &= \inner{\nabla \mathcal{L}_{lt}(\hat{\bm{\psi}}_{lt})}{\hat{\bm{\psi}}_{lt} - \bm{\psi}_{lt}} - \frac{n_{lt}c_\mathcal{L}}{2}\inner{\hat{\bm{\psi}}_{lt} - \bm{\psi}_{lt}}{ \bm{Q}_{lt}}^2\\
          &= \inner{\nabla \mathcal{L}_{lt}(\hat{\bm{\psi}}_{lt})}{\hat{\bm{\psi}}_{l,t+1} - \bm{\psi}_{lt}} + \inner{\nabla \mathcal{L}_{lt}(\hat{\bm{\psi}}_{lt})}{\hat{\bm{\psi}}_{lt} - \hat{\bm{\psi}}_{l,t+1}} - \frac{n_{lt}c_\mathcal{L}}{2}\inner{\hat{\bm{\psi}}_{lt} - \bm{\psi}_{lt}}{ \bm{Q}_{lt}}^2\,.
     \end{split}
     \end{align}
Our update rules in \eqref{eq:19}, gives us
 \begin{align*}
    \hat{b}_{l,t+1} &= \Pi_{\Theta_{b}}(\hat{b}_{lt} - \eta_t \nabla \mathcal{L}_{lt}^{b})\,,\\
    \hat{\bm m}_{l,t+1} &= \Pi_{\Theta_{\bm m}}(\hat{\bm m}_{lt} - \eta_t \nabla \mathcal{L}_{lt}^{\bm m} )\,.
\end{align*}
     The updates defined are common OMD updates and can be rewritten as

    \[\hat{\bm{\psi}}_{l,t+1} = \arg\min_{\bm{\psi} }
 \inner{\nabla \mathcal{L}_{lt}(\hat{\bm{\psi}}_{lt})}{\bm{\psi}}  + \frac{1}{2 \eta_{t}}\|\bm{\psi} - \hat{\bm{\psi}}_{lt}\|^2\,.\]

Since the above loss function is convex and $\hat{\bm{\psi}}_{l,t+1}$ is the minimizer, we get

\[\inner{\bm{\psi}-\hat{\bm{\psi}}_{l,t+1}}{\eta_t\nabla\mathcal{L}_{lt}(\hat{\bm{\psi}}_{lt}) + \hat{\bm{\psi}}_{l,t+1} - \hat{\bm{\psi}}_{lt}} \geq 0\,.\]
 
 Putting $\bm{\psi} = \bm{\psi}_{lt}$ above, we get $\inner{\hat{\bm{\psi}}_{l,t+1} - \bm{\psi}_{lt}}{\eta_{t}\nabla \mathcal{L}_{lt}(\hat{\bm{\psi}}_{lt})} \leq \inner{\bm{\psi}_{lt} - \hat{\bm{\psi}}_{l,t+1}}{\hat{\bm{\psi}}_{l,t+1} - \hat{\bm{\psi}}_{lt}}$.

Also, note that 
 \begin{equation*}\
   \inner{\bm{\psi}_{lt} - \hat{\bm{\psi}}_{l,t+1}}{\hat{\bm{\psi}}_{l,t+1} - \hat{\bm{\psi}}_{lt}} = \frac{1}{2}\left(\|\bm{\psi}_{lt} - \hat{\bm{\psi}}_{lt}\|^2 - \|\bm{\psi}_{lt} -  \hat{\bm{\psi}}_{l,t+1}\|^2 - \| \hat{\bm{\psi}}_{l,t+1}- \hat{\bm{\psi}}_{lt}\|^2\right) \,. 
 \end{equation*}

With the above two equations the first term in \eqref{eq:28} is bounded as:
\[\inner{\nabla \mathcal{L}_{lt}(\hat{\bm{\psi}}_{lt})}{\hat{\bm{\psi}}_{l,t+1} - \bm{\psi}_{lt}} \leq \frac{1}{2\eta_t}\left(\|\bm{\psi}_{lt} - \hat{\bm{\psi}}_{lt}\|^2 - \|\bm{\psi}_{lt} -  \hat{\bm{\psi}}_{l,t+1}\|^2 - \| \hat{\bm{\psi}}_{l,t+1}- \hat{\bm{\psi}}_{lt}\|^2\right)\,. \]

Using the inequality $ab\le (a^2+b^2)/2$, the second term in \eqref{eq:28} can be bounded as,
     \begin{equation}\label{eq:14}
         \inner{\nabla \mathcal{L}_{lt}(\hat{\bm{\psi}}_{lt})}{\hat{\bm{\psi}}_{lt} - \hat{\bm{\psi}}_{l,t+1}}  \leq \frac{1}{2\eta_{t}}\|\hat{\bm{\psi}}_{lt} - \hat{\bm{\psi}}_{l,t+1}\|^2 + \frac{\eta_{t}}{2}\|\nabla \mathcal{L}_{lt}(\hat{\bm{\psi}}_{lt})\|^2\,.
     \end{equation}

     Also, $\nabla \mathcal{L}_{lt}(\bm{\psi}) = -\left(y_{lt}\bm{Q}_{lt}\frac{\partial}{\partial \zeta^2}\log \Phi(\zeta)|_{\zeta = \bm{Q}_{lt}\bm{\psi}} - \Tilde{y}_{lt}\bm{Q}_{lt}\frac{\partial}{\partial \zeta^2}\log \Phi(\zeta)|_{\zeta = -\bm{Q}_{lt}\bm{\psi}}\right)$. 
     
     Let $C_{\mathcal{L}} = \max\{-\frac{\partial}{\partial \zeta^2}\log \Phi(\zeta)|_{\zeta = \bm{Q}_{lt}\bm{\psi}}, \frac{\partial}{\partial \zeta^2}\log \Phi(\zeta)|_{\zeta = -\bm{Q}_{lt}\bm{\psi}}\}$ in the restricted space. Hence, $\|\nabla \mathcal{L}_{lt}(\hat{\bm{\psi}}_{lt})\|^2 \leq {C_{\mathcal{L}}^2 n_{lt}^2}\|\bm{Q}_{lt}\|^2$.

Combining all the parts we have,

\[\mathcal{L}_{lt}(\hat{\bm{\psi}}_{lt}) - \mathcal{L}_{lt}(\bm{\psi}_t) \leq \frac{1}{2\eta_t}\|\bm{\psi}_{lt} - \hat{\bm{\psi}}_{lt}\|^2 - \frac{1}{2\eta_t}\|\bm{\psi}_{lt} -  \hat{\bm{\psi}}_{l,t+1}\|^2 + \frac{\eta_t}{2}{C_{\mathcal{L}}^2 n_{lt}^2}\|\bm{Q}_{lt}\|^2- \frac{n_{lt}c_\mathcal{L}}{2}\inner{\hat{\bm{\psi}}_{lt} - \bm{\psi}_{lt}}{ \bm{Q}_{lt}}^2\,.\]

Adding and subtracting $\|\bm{\psi}_{l,t+1} - \hat{\bm{\psi}}_{l,t+1}\|^2$ to above we get
\begin{align}
    \mathcal{L}_{lt}(\hat{\bm{\psi}}_{lt}) - \mathcal{L}_{lt}(\bm{\psi}_t) &\leq \frac{1}{2\eta_t}\left(\|\bm{\psi}_{lt} - \hat{\bm{\psi}}_{lt}\|^2 - \|\bm{\psi}_{l,t+1} - \hat{\bm{\psi}}_{l,t+1}\|^2\right)\nonumber\\ &\quad+ \frac{1}{2\eta_t}\left(\|\bm{\psi}_{l,t+1} - \hat{\bm{\psi}}_{l,t+1}\|^2 - \|\bm{\psi}_{lt} -  \hat{\bm{\psi}}_{l,t+1}\|^2\right) \nonumber\\
& \quad+ \frac{\eta_t}{2}{C_{\mathcal{L}}^2 n_{lt}^2}\|\bm{Q}_{lt}\|^2- \frac{n_{lt}c_\mathcal{L}}{2}\inner{\hat{\bm{\psi}}_{lt} - \bm{\psi}_{lt}}{ \bm{Q}_{lt}}^2\,.\label{eq:telescope}
\end{align}
The second term can be simplified as 
\[\|\bm{\psi}_{l,t+1} - \hat{\bm{\psi}}_{l,t+1}\|^2 - \|\bm{\psi}_{lt} -  \hat{\bm{\psi}}_{l,t+1}\|^2 = \inner{\bm{\psi}_{l,t+1}+ \bm{\psi}_{lt} - 2\hat{\bm{\psi}}_{l,t+1}}{\bm{\psi}_{l,t+1} - \bm{\psi}_{lt}}\leq 4C_{\bm{\psi}} \|\bm{\psi}_{l,t+1} - \bm{\psi}_{lt}\|_2\,,\]
where $C_{\bm{\psi}}$ is $\max\|\bm{\psi}\|$ and $C_{\bm{\psi}} \leq 2C_b + 2C_m$, since $\bm{\psi} = (\bm m, b)$.

Summing both sides of~\eqref{eq:telescope} over $t=1,\dotsc,T$, we get  $\sum_{t = 1}^T\left(\mathcal{L}_{lt}(\hat{\bm{\psi}}_{lt}) - \mathcal{L}_{lt}(\bm{\psi}_t)\right)$
is bounded above by:
\begin{align*}
    &\frac{\|\bm{\psi}_{l1} - \hat{\bm{\psi}}_{l1}\|^2}{2\eta_1} + \sum_{t = 2}^T\|\bm{\psi}_{lt} - \hat{\bm{\psi}}_{lt}\|^2\left(\frac{1}{2\eta_{t+1}} - \frac{1}{2\eta_t}\right) + 4C_{\bm{\psi}}\sum_{t = 1}^T \frac{1}{2\eta_t} \|\bm{\psi}_{l,t+1} - \bm{\psi}_{lt}\|_2 \\
&+ \sum_{t = 1}^T \frac{\eta_t}{2}{C_{\mathcal{L}}^2 n_{lt}^2}\|\bm{Q}_{lt}\|^2- \sum_{t = 1}^T \frac{n_{lt}c_\mathcal{L}}{2}\inner{\hat{\bm{\psi}}_{lt} - \bm{\psi}_{lt}}{ \bm{Q}_{lt}}^2\,.
\end{align*}

 Under the assumption that $\eta_t$ are non-decreasing, 
\[\frac{\|\bm{\psi}_{l1} - \hat{\bm{\psi}}_{l1}\|^2}{2\eta_1} + \sum_{t = 2}^T\|\bm{\psi}_{lt} - \hat{\bm{\psi}}_{lt}\|^2\left(\frac{1}{2\eta_{t+1}} - \frac{1}{2\eta_t}\right) \leq \frac{4C_{\bm{\psi}}^2}{2\eta_1} + 4C_{\bm{\psi}}^2\sum_{t = 2}^T\left(\frac{1}{2\eta_{t+1}} - \frac{1}{2\eta_t}\right) = \frac{4C_{\bm{\psi}}^2}{2\eta_{T+1}}\,.\]

Hence, we finally have 
\begin{align*}
    \sum_{t = 1}^T\left(\mathcal{L}_{lt}(\hat{\bm{\psi}}_{lt}) - \mathcal{L}_{lt}(\bm{\psi}_t)\right) \leq  &\frac{4C_{\bm{\psi}}^2}{2\eta_{T+1}} + 4C_{\bm{\psi}}\sum_{t = 1}^T \frac{1}{2\eta_t} \|\bm{\psi}_{l,t+1} - \bm{\psi}_{lt}\|_2 \nonumber\\
&+ \sum_{t = 1}^T \frac{\eta_t}{2}{C_{\mathcal{L}}^2 n_{lt}^2}\|\bm{Q}_{lt}\|^2- \sum_{t = 1}^T \frac{n_{lt}c_\mathcal{L}}{2}\inner{\hat{\bm{\psi}}_{lt} - \bm{\psi}_{lt}}{ \bm{Q}_{lt}}^2\,.
\end{align*}

Define 
\begin{align}\label{eq:Adef}
A:= \frac{4C_{\bm{\psi}}^2}{2\eta_{T+1}} + 4C_{\bm{\psi}}\sum_{t = 1}^T \frac{1}{2\eta_t} \|\bm{\psi}_{l,t+1} - \bm{\psi}_{lt}\|_2 
+ \sum_{t = 1}^T \frac{\eta_t}{2}{C_{\mathcal{L}}^2 n_{lt}^2}\|\bm{Q}_{lt}\|^2.
\end{align}


Since, $\|\bm{\psi}_{l,t+1} - \bm{\psi}_{lt}\|_2 \leq 2 (\|\bm m_{l,t+1} - \bm m_{lt}\|_2 + |b_{l,t+1} - b_{lt}|)$ and $\|\bm{Q}_{lt}\|^2$ is bounded, we can simplify $A$ as 
\[A:= \Tilde{C}_1\sum_{t = 1}^T \frac{1}{\eta_t} \|\bm m_{l,t+1} - \bm m_{lt}\|_2 + \Tilde{C}_2\sum_{t = 1}^T \frac{1}{\eta_t} |b_{l,t+1} - b_{lt}|
+ \Tilde{C}_3\sum_{t = 1}^T{\eta_t} n_{lt}^2 + \frac{\Tilde{C}_4}{\eta_{T+1}}\,. \]

Note that in order to prove the lemma we need to show a bound on  $\sum_{t = 1}^T n_{lt} \Big(\langle\bm{x}_{lt}, \bm m_{lt} - \hat{\bm m}_{lt}\rangle^2 + \big(p_{lt}(b_{lt} - \hat{b}_{lt})\big)^2\Big)$ which is same as showing a bound on $\sum_{t = 1}^T {n_{lt}}\inner{\hat{\bm{\psi}}_{lt} - \bm{\psi}_{lt}}{ \bm{Q}_{lt}}^2$, since $\bm{\psi}_{lt} = (\bm m_{lt},b_{lt})$ and $\bm{Q}_{lt} = (\bm x_{lt},p_{lt})$.

We next provide a lower bound on the cumulative difference $\sum_{t = 1}^T\left(\mathcal{L}_{lt}(\hat{\bm{\psi}}_{lt}) - \mathcal{L}_{lt}(\bm{\psi}_t)\right) $. Write
\begin{align}\label{eq:Dtk}
 \mathcal{L}_{ltk}({\bm{\psi}}_{lt}) - \mathcal{L}_{ltk}(\hat{\bm{\psi}}_{lt}) \leq \inner{\nabla\mathcal{L}_{ltk}(\bm{\psi}_{lt})}{\hat{\bm{\psi}}_{lt} - \bm{\psi}_{lt}}:=D_{tk}\,,
 \end{align}
using convexity of the loss $\mathcal{L}_{ltk}$. We also have
\begin{align*}
    \nabla \mathcal{L}_{ltk}(\bm{\psi}) &= -(y_{ltk}\frac{\partial}{\partial\bm{\psi}}\log \Phi(\bm{Q}_{lt}\bm{\psi}) - (1 - y_{ltk})\frac{\partial}{\partial\bm{\psi}}\log \Phi(-\bm{Q}_{lt}\bm{\psi}))\\ 
    &= \bm{Q}_{lt}\left(-y_{ltk}\frac{\phi(\bm{Q}_{lt}\bm{\psi})}{\Phi(\bm{Q}_{lt}\bm{\psi})}+ (1-{y}_{ltk})\frac{\phi(-\bm{Q}_{lt}\bm{\psi})}{\Phi(-\bm{Q}_{lt}\bm{\psi})}\right)\,.
\end{align*}

Let $\mathcal{F}_t$ be the $\sigma$-field generated by the noise till time $t$. Then, since $\hat{\bm{\psi}}_{lt}$ only depends on noise till time $t$,
$\mathbb{E}[D_{tk}|\mathcal{F}_{t-1}] =  \inner{\mathbb{E}[\nabla\mathcal{L}_{ltk}|\mathcal{F}_{t-1}]}{\hat{\bm{\psi}}_{lt} - \bm{\psi}_{lt}}$. In addition, $E[\nabla\mathcal{L}_{ltk}|\mathcal{F}_{t-1}]=0$ using the fact that $\mathbb P(Y_{ltk} = 1) = \Phi(\bm{Q}_{lt}\bm{\psi})$ and $\mathbb P(Y_{ltk} = 0) = \Phi(-\bm{Q}_{lt}\bm{\psi}))$. Therefore, the partial sums of $D_{tk}$ is a martingale with respect to the filtration $\mathcal{F}_t$.

Also, as described above $\left(-y_{ltk}\frac{\phi(\bm{Q}_{lt}\bm{\psi})}{\Phi(\bm{Q}_{lt}\bm{\psi})}+ (1-{y}_{ltk})\frac{\phi(-\bm{Q}_{lt}\bm{\psi})}{\Phi(-\bm{Q}_{lt}\bm{\psi})}\right)$ is bounded above with $C_\mathcal{L}$. Hence $|D_{tk}| \leq \beta_t:=C_\mathcal{L}|\inner{\bm{Q}_{lt}}{\hat{\bm{\psi}}_{lt} - \bm{\psi}_{lt}}|$. Using convexity of $e^{\lambda z}$, for any $\lambda \in \mathbb R$ we have

\begin{align*}
\mathbb{E}\left[e^{\lambda D_{tk}} \mid \mathcal{F}_{t-1}\right] & \leq \mathbb{E}\left[\frac{\beta_t-D_{tk}}{2 \beta_t} e^{-\lambda \beta_t}+\frac{\beta_t+D_{tk}}{2 \beta_t} e^{\lambda \beta_t} \mid \mathcal{F}_{t-1}\right] \\
& =\mathbb{E}\left[\frac{e^{-\lambda \beta_t}+e^{\lambda \beta_t}}{2}\right]+\mathbb{E}\left[D_{tk} \mid \mathcal{F}_{t-1}\right]\left(\frac{e^{-\lambda \beta_t}+e^{\lambda \beta_t}}{2 \beta_t}\right)=\cosh \left(\lambda \beta_t\right) \leq e^{\lambda^2 \beta_t^2 / 2}\,.
\end{align*}

where $\beta_t = C_\mathcal{L}|\inner{\bm{Q}_{lt}}{\hat{\bm{\psi}}_{lt} - \bm{\psi}_{lt}}|$. We next use the following result from \citep[Proposition C.1]{javanmard2017perishability}.

\begin{proposition} \citep[Proposition C.1]{javanmard2017perishability}
    Consider a martingale difference sequence $D_t$ adapted to a filtration $\mathcal{F}_t$, such that for any $\lambda \geq 0, \mathbb{E}\left[e^{\lambda D_t} \mid \mathcal{F}_{t-1}\right] \leq e^{\lambda^2 \sigma_t^2 / 2}$. Then, for $D(T)=\sum_{t=1}^T D_t$, the following holds true:
$$
\mathbb{P}(D(T) \geq \xi) \leq e^{-\xi^2 /\left(2 \sum_{t=1}^T \sigma_t^2\right)}\,.
$$
\end{proposition}
We apply the above theorem with $D(T) = \sum_{t = 1}^T\sum_{k = 1}^{n_{lt}}D_{tk}$. Invoking~\eqref{eq:Dtk}, this gives us, 
\[\mathbb{P}\left( \sum_{t = 1}^T\left(\mathcal{L}_{lt}(\hat{\bm{\psi}}_{lt}) - \mathcal{L}_{lt}(\bm{\psi}_t)\right) \leq -2C_\mathcal{L} \sqrt{\log T}\left\{\sum_{t=1}^Tn_{lt}\left\langle \bm{Q}_{lt}, \bm{\psi}_{lt}-\hat{\bm{\psi}}_{lt}\right\rangle^2\right\}^{1 / 2}\right)\leq \frac{1}{T^2}\,.\]

Hence with probability at least $1 - 1/T^2$, 
\[\sum_{t = 1}^T\left(\mathcal{L}_{lt}(\hat{\bm{\psi}}_{lt}) - \mathcal{L}_{lt}(\bm{\psi}_t)\right) \geq -2C_\mathcal{L} \sqrt{\log T}\left\{\sum_{t=1}^Tn_{lt}\left\langle \bm{Q}_{lt}, \bm{\psi}_{lt}-\hat{\bm{\psi}}_{lt}\right\rangle^2\right\}^{1 / 2}\,.\]

Let $B = \sum_{t = 1}^T {n_{lt}}\inner{\hat{\bm{\psi}}_{lt} - \bm{\psi}_{lt}}{ \bm{Q}_{lt}}^2$, then with the complete analysis till now we have

\[-2C_\mathcal{L} \sqrt{B \log T} \leq \sum_{t = 1}^T\left(\mathcal{L}_{lt}(\hat{\bm{\psi}}_{lt}) - \mathcal{L}_{lt}(\bm{\psi}_t)\right) \leq A - \frac{c_\mathcal{L}}{2}B\,.\]

Hence, $B - (4 C_\mathcal{L}/c_\mathcal{L})\sqrt{B\log T}\leq (2/c_\mathcal{L})A$. Consider two cases:

Case 1: $\sqrt{B\log T} \leq (c_\mathcal{L}/8C_\mathcal{L}) B$, then $B \leq 4A/c_\mathcal{L}$.

Case 2: $\sqrt{B\log T} \geq (c_\mathcal{L}/8C_\mathcal{L}) B$, then $B \leq (8 C_\mathcal{L}/c_{\mathcal{L}})^2\log T$.

Combining the two cases, we have 
\[B \leq \frac{4A}{c_\mathcal{L}} + \mathcal{O}(\log T)\,.\]

Substituting for 
\[B = \sum_{t = 1}^T {n_{lt}}\inner{\hat{\bm{\psi}}_{lt} - \bm{\psi}_{lt}}{ \bm{Q}_{lt}}^2 = \sum_{t = 1}^T {n_{lt}} \left(\inner{\bm x_{lt}}{\bm m_{lt} - \hat{\bm m}_{lt}}^2+ p_{lt}^2(b_{lt} - \hat b_{lt})^2\right)\,,\]
 and $A$ from~\eqref{eq:Adef} we obtain the desired result.

\subsection{Proof of Theorem~\ref{thm.2}}\label{lemma.c2}
Recall the varaince of segment $l$ in the utility model given by $V_{lt}^2 = \|(\bm I - \rho_t \bm W)^{-1} \bm e_l\|^2\tau^2+\sigma^2$. We assume that we have a known fixed $\rho$, the auto-correlation parameter, and so the variances do not change over time. 

Indicate the variances of segments by $V_{1}, V_{2}, \cdots, V_{L}$. Our utility model is thus 
\[\Tilde{U}_{ltk}= \frac{\beta_{t}}{V_{l}}\,p_{lt}+\bm{x}_{lt}' \frac{\bm\mu_{t}}{V_{l}} + Z_{ltk}\,.\]

Without loss of generality, assume that $\bm x_{lt}$ is of dimension one. We would give a small variation that would work for any dimension as well. Let $v_{1}, v_{2}, \cdots, v_{L}$ be the inverse of the fixed variances. The model is thus, 

\[\Tilde{U}_{ltk}= {\beta_{t}}{v_{l}}\,p_{lt}+{x}_{lt} {\mu_{t}}{v_{l}} + Z_{ltk\,.}\]

Assume that $-\beta_t = \mu_t = \gamma$, i.e. the parameters do not change over time and are negative of each other. In this setup $U^0_{ltk} = v_l\gamma(x_{lt} - p_{lt} )$, the noiseless utility. Here setting $p_{lt} = x_{lt}$ would be uninformative since we would just observe noise, and we cannot get any information about the unknown parameter $\gamma$. In addition, in our model a price $p^*_{lt}$ is optimum if it satisfies
\[p^*_{lt} = -\frac{1}{\beta_t v_l}\frac{\Phi({\beta_{t}}{v_{l}}\,p^*_{lt}+{x}_{lt} {\mu_{t}}{v_{l}})}{\phi({\beta_{t}}{v_{l}}\,p^*_{lt}+{x}_{lt} {\mu_{t}}{v_{l}})}\,.\]
Under the assumption that $-\beta_t = \mu_t = \gamma$, this reduces to
\[p^*_{lt} = \frac{1}{\gamma v_l}\frac{\Phi(v_l\gamma(x_{lt} - p^*_{lt}))}{\phi(v_l\gamma(x_{lt} - p^*_{lt}))}\,.\]

Therefore, for $\gamma_0:= (v_l x_{lt})^{-1} \Phi(0)/\phi(0)$, the uninformative price is optimal prices, i.e., $p^*_{lt}(\gamma_0) = x_{lt}$. 

Note that if $x_{lt}$ was of higher dimension, we could set $\bm x_{lt} = (a/d, a/d, a/d, \cdots, a/d)$ with $ d$ the dimension of $\bm x_{lt}$ and set $\bm \mu_t = (\gamma, \gamma, \gamma, \cdots, \gamma)$ to get the exactly same result: $p^*_{lt}(\gamma_0) = a$ for $\gamma_0= (v_l a)^{-1}\Phi(0)/\phi(0)$ is an uniformative price. 

Now that we know the existence of a setting where the uninformative prices are optimal prices, we can show that the regret is at least of the order of $\sqrt{T}$.


We construct a problem class $(\Gamma, \{\mathcal{P}_{lt}\})$, for $l=1,\dotsc, L$, $t=1,\dotsc, T$ as follows. Recall $\gamma_0$ the parameter for which the optimal price is uninformative. We use the shorthand $r_{lt}(p,\gamma)$ to denote the expected revenue obtained from a typical customer from segment $l$ at time $t$, if the model parameter is $\gamma$. Therefore, recalling our utility model $U_{ltk} = v_l\gamma(x_{lt}-p_{lt})+Z_{ltk}$, we have $r_{lt}(p,\gamma) = p(\Phi(v_l\gamma(x_{lt} - p)))$. By optimality of $p^*_{lt}(\gamma_0)$, we have $r''_{lt}(p^*_{lt}(\gamma_0),\gamma_0) <-2c$ for some constant $c>0$, and by continuity of $r''_{lt}$ we can find a neighborhood $\mathcal{P}_{lt}$ around $p^*_{lt}(\gamma_0)$ such that $r''_{lt}(p,\gamma_0)<-c$ for all $p\in\mathcal{P}_{lt}$. 
We next consider the mapping $\gamma\mapsto p^*_{lt}(\gamma)$. By continuity of this mapping, we can find a small enough neighborhood $\Gamma_{lt}$ around $\gamma_0$ such that the  optimal prices $p^*_{lt}(\gamma)\in \mathcal{P}_{lt}$ for all $\gamma\in \Gamma_{lt}$. Finally, we take $\Gamma: = \cap_{t=1}^T \cap_{l=1}^L \Gamma_{lt}$. Note that $\Gamma$ is non-empty because $\gamma_0\in \Gamma$. Furthermore, by our construction we have the following properties for the problem class $(\Gamma, \{\mathcal{P}\}_{lt})$, for $l=1,\dotsc, L$ and $t=1,\dotsc, T$:
\begin{itemize}
\item For all $\gamma\in \Gamma$, we have $p^*_{lt}(\gamma)\in \mathcal{P}_{lt}$.
\item For all prices $p\in\mathcal{P}_{lt}$, we have $r''_{lt}(p,\gamma_0)<-c$\,. 
\end{itemize}





For any pricing policy $\pi$ and a parameter $\gamma \in \Gamma$, let $f_{t}^{\pi, \gamma} : \{0,1\}^{N_t} \rightarrow [0,1]$ be the probability distribution function for all the consumers purchase responses $\bm Y= (Y_{ljk}, \ell = 1,\dotsc, L, j=1,\dotsc, t, k=1,\dotsc, n_{lj})$ until time $t$. Here, $N_t = \sum_{j = 1}^t\sum_{l = 1}^L n_{lj}$, under policy $\pi$ and model parameter $\gamma$. 
The pricing policy uses all the sales data till time $t-1$ to give a price $p_{lt}^*$. We use $\bm y_{t}\in\{0,1\}^{N_t}$ to denote all sales data till time $t$. So, if the pricing policy gives the prices $p_{lt} := \pi(\bm y_{t-1})$ for all the time periods, then 
\[f_t^{\pi, \gamma}(\bm y_t) = \prod_{j = 1}^t\prod_{l = 1}^L\prod_{k = 1}^{n_{lj}} q_{lj}(p_{lj}, \gamma)^{y_{ljk}}(1 - q_{lj}(p_{lj}, \gamma))^{1 - y_{ljk}}\,,\]
where $q_{lj}(p_{lj}, \gamma) = \Phi(v_l\gamma(x_{lj} - p_{lj} ))$. We next want to show that for $\gamma_0$, the  parameter for which the uninformative price is optimal, any policy incurs a large regret if it tries to learn $\gamma_0$. Formally, we aim to show that

    \[\mathcal{R}^\pi_t(\gamma_0) \geq C\frac{1}{(\gamma_0 - \gamma)^2} \KL(f_t^{\pi, \gamma_0},f_t^{\pi, \gamma})\,.\]

We employ the chain rule for KL divergence \citep{cover1991information},

\begin{align*}
&\KL\left(f_t^{\pi, \gamma_0} ; f_t^{\pi, \gamma}\right)  =\sum_{s=1}^t \KL\left(f_t^{\pi, \gamma_0} ; f_t^{\pi, \gamma} | \bm {Y}_{s-1}\right) \\
& =\sum_{s=1}^t \sum_{\mathbf{y}_s \in\{0,1\}^{N_s}} f_s^{\pi, \gamma_0}\left(\mathbf{y}_s\right) \log \left(\frac{f_s^{\pi, \gamma_0}\left(y_{lsk} \mid \mathbf{y}_{s-1}\right)}{f_s^{\pi, \gamma}\left(y_{lsk} \mid \mathbf{y}_{s-1}\right)}\right) \\
& =\sum_{s=1}^t \sum_{\mathbf{y}_{s-1} \in\{0,1\}^{N_{s-1}}} f_{s-1}^{\pi, \gamma_0}\left(\mathbf{y}_{s-1}\right) \sum_{l = 1}^L \sum_{k = 1}^{n_{ls}}\sum_{y_{lsk} \in\{0,1\}} f_s^{\pi, \gamma_0}\left(y_{lsk} \mid \mathbf{y}_{s-1}\right) \log \left(\frac{f_s^{\pi, \gamma_0}\left(y_{lsk} \mid \mathbf{y}_{s-1}\right)}{f_s^{\pi, \gamma}\left(y_{lsk} \mid \mathbf{y}_{s-1}\right)}\right) \\
& =\sum_{s=1}^t \sum_{\mathbf{y}_{s-1} \in\{0,1\}^{N_{s-1}}} f_{s-1}^{\pi, \gamma_0}\left(\mathbf{y}_{s-1}\right) \sum_{l = 1}^L \sum_{k = 1}^{n_{ls}}\KL\Big(f_s^{\pi, \gamma_0}\left(y_{lsk} \mid \mathbf{y}_{s-1}\right) ; f_s^{\pi, \gamma}\left(y_{lsk} \mid \mathbf{y}_{s-1}\right)\Big)\,.
\end{align*}

Based on the definition of $f_t^{\pi, \gamma}$, $f_s^{\pi,\gamma_0}(y_{lsk})$ is distributed as Bernoulli $q_{ls}(p_{ls}, \gamma_0)$ and  $f_s^{\pi,\gamma}(y_{lsk})$ is distributed as Bernoulli $q_{ls}(p_{ls}, \gamma)$. Using the fact that for Bernoulli random variables $B_1\sim {\sf Bern}(q_1), B_2\sim {\sf Bern}(q_2)$, we have $\KL(B_1, B_2) \leq \frac{(q_1 - q_2)^2}{q_2(1 - q_2)}$, we get

\begin{align}\label{eq:KL}
\KL\left(f_t^{\pi, \gamma_0} ; f_t^{\pi, \gamma}\right) \leq \sum_{s=1}^t \sum_{\mathbf{y}_{s-1} \in\{0,1\}^{N_{s-1}}} f_{s-1}^{\pi, \gamma_0}\left(\mathbf{y}_{s-1}\right) \sum_{l = 1}^L \sum_{k = 1}^{n_{ls}}\frac{(q_{ls}(p_{ls}, \gamma_0) - q_{ls}(p_{ls}, \gamma))^2}{q_{ls}(p_{ls}, \gamma)(1 - q_{ls}(p_{ls}, \gamma))}\,.
\end{align}

Since the prices and the parameters are bounded, and $q_{lt}$ is the normal distribution function, $q_{lt}$ is bounded away from zero. Hence, there exists constant $C$ such that $q_{ls}(1- q_{ls}) \geq C$.

Also, $q_{ls}(p_{ls}^*, \gamma) = \Phi(v_l\gamma(x_{ls} - p_{ls}^* ))$ and $q_{ls}(p_{ls}^*, \gamma_0) = \Phi(v_l\gamma_0(x_{ls} - p_{ls}^* ))$. Since we are working on a bounded set, the distribution function $\Phi$ is Lipschitz as well. Hence,
\begin{align*}
    q_{ls}(p_{ls}, \gamma_0) - q_{ls}(p_{ls}, \gamma) &= \Phi(v_l\gamma_0(x_{ls} - p_{ls} )) - \Phi(v_l\gamma(x_{ls} - p_{ls} ))\\
    &\leq C (v_l\gamma_0(x_{ls} - p_{ls} ) - v_l\gamma(x_{ls} - p_{ls}))\\
    &=C v_l (\gamma_0 - \gamma)(x_{ls} - p_{ls} )\\
    &= C v_l (\gamma_0 - \gamma)(p_{ls}^*(\gamma_0) - p_{ls} )\,,
\end{align*}

where $p^*_{ls}(\gamma_0)$ is the optimal price for when the parameter is $\gamma_0$. Recall that by the definition of $\gamma_0$, the optimal price for $\gamma_0$ is $x_{lt}$. We thus have \[(q_{ls}(p_{ls}, \gamma_0) - q_{ls}(p_{ls}, \gamma))^2 \leq C(\gamma_0 - \gamma)^2(p^*_{ls}(\gamma_0) - p_{ls} )^2\,.\]

Using the above bound in~\eqref{eq:KL}, we get 
\[\KL\left(f_t^{\pi, \gamma_0} ; f_t^{\pi, \gamma}\right) \leq C (\gamma - \gamma_0)^2\sum_{s=1}^t   \sum_{l = 1}^L \sum_{k = 1}^{n_{ls}}\sum_{\mathbf{y}_{s-1} \in\{0,1\}^{N_{s-1}}}f_{s-1}^{\pi, \gamma_0}\left(\mathbf{y}_{s-1}\right)(p_{ls}^*(\gamma_0) - p_{ls} )^2\,.\]

The inner summation is indeed the expectation with respect to $\gamma_0$, by noting that $p_{ls}$ is a measurable function of $\bm y_{s-1}$. Hence, we have 
\begin{align}
\KL\left(f_t^{\pi, \gamma_0} ; f_t^{\pi, \gamma}\right) &\leq C (\gamma - \gamma_0)^2\sum_{s=1}^t   \sum_{l = 1}^L \sum_{k = 1}^{n_{ls}}\mathbb{E}_{\gamma_0}(p_{ls}^* (\gamma_0)-p_{ls})^2\nonumber\\
&= C (\gamma - \gamma_0)^2\sum_{s=1}^t   \sum_{l = 1}^Ln_{ls}\mathbb{E}_{\gamma_0}(p_{ls}^*(\gamma_0) - p_{ls} )^2\,.\label{eq:KL0}
\end{align}

By the construction of problem class $(\Gamma, \{\mathcal{P}_{lt}\})$, we have $r''_{lt}(p,\gamma_0)\le -c$, for $\gamma\in\Gamma$ and $p\in\mathcal{P}_{lt}$. Therefore, by Taylor expansion of $r_{ls}(p,\gamma)$ around $p^*_{ls}$, we obtain
\[
r_{ls}(p_{ls},\gamma_0) = r_{ls}(p^*_{ls}(\gamma_0),\gamma_0) + r'_{ls}(p^*_{ls}(\gamma_0),\gamma_0)(p_{ls} - p^*_{ls}(\gamma_0))+\frac{1}{2} r''_{ls}(\tilde{p},\gamma_0)(p_{ls} - p^*_{ls}(\gamma_0))^2\,,
\]
for some $\tilde{p}$ between $p_{ls}$ and $p^*_{ls}$.
By optimality of $p^*_{ls}$ we have $r'_{ls}(p^*_{ls}(\gamma_0),\gamma_0)=0$. In addition, since $\tilde{p}\in\mathcal{P}_{ls}$, we have $r''_{ls}(\tilde{p},\gamma_0)<-c$, which implies that
\[
(p_{ls} - p^*_{ls}(\gamma_0))^2\le \frac{2}{c}\Big(r_{ls}(p^*_{ls}(\gamma_0),\gamma_0) - r_{ls}(p_{ls},\gamma_0)\Big)\,.
\]
Using the above bound in~\eqref{eq:KL0}, we arrive at
\begin{align}\label{eq:KL-Bound1}
\KL\left(f_t^{\pi, \gamma_0} ; f_t^{\pi, \gamma}\right) \le  C (\gamma - \gamma_0)^2\sum_{s=1}^t   \sum_{l = 1}^L  n_{ls}\mathbb{E}_{\gamma_0}[r_{ls}(p_{ls}^*(\gamma_0),\gamma_0) - r_{ls}(p_{ls},\gamma_0)] \leq C (\gamma - \gamma_0)^2\text{ Reg}_t\,,
\end{align}
which completes the proof~\eqref{eq:KL-LB}.


We next proceed with our proof for bound~\eqref{eq:KL-LB2}. Recall the optimality condition

\[v_l \gamma p^*_{lt}(\gamma) = \frac{\Phi(v_l\gamma(x_{lt} - p^*_{lt}(\gamma)))}{\phi(v_l\gamma(x_{lt} - p^*_{lt}(\gamma)))}\,.\]

Differentiating with respect to $\gamma$ on both sides we get,

\[v_l p^*_{lt}(\gamma) + v_l \gamma \frac{d}{d\gamma}p^*_{lt}(\gamma) = v_l \left(x_{lt} - p^*_{lt}(\gamma) - \gamma\frac{d}{d\gamma}p^*_{lt}(\gamma)\right) \kappa(\gamma)\,, \]
where 
\[
\kappa(\gamma) = \frac{\phi^2(v_l\gamma(x_{lt} - p^*_{lt}(\gamma))) - \Phi(v_l\gamma(x_{lt} - p^*_{lt}(\gamma)))\phi'(v_l\gamma(x_{lt} - p^*_{lt}(\gamma))) }{\phi^2(v_l\gamma(x_{lt} - p^*_{lt}(\gamma)))}\,.
\]
By rearranging the terms we have

\[
\frac{d}{d\gamma}p^*_{lt}(\gamma) = 
\frac{1}{\gamma}\left(-p^*_{lt}(\gamma) + \frac{k(\gamma)}{1 + k(\gamma)}\right).\] Since we are working on finite sets, we can restrict the problem class $\Gamma$, such that $|\frac{d}{d\gamma}p(\gamma)| > C$, for some constant $C$ and all $\gamma\in\Gamma$. 
Therefore, by an application of the Mean Value Theorem, we have
\[
|p^*_{lt}(\gamma) - p^*_{lt}(\gamma_0)|
\geq C|\gamma - \gamma_0|\,.
\]
Let $\gamma_1 := \gamma_0 + 1/(4T^{1/4})$. Using the above bound, the optimal prices for $\gamma_0$ and $\gamma_1$ are apart by at least $C/(4T^{1/4})$. 

Consider two disjoint sets $D_1$ and $D_0$ of prices, as follows:
\[D_{\gamma_0} := \left\{p: |p - p^*_{lt}(\gamma_0)| \leq \frac{C}{10T^{1/4}}\right\}\,, \quad \quad D_{\gamma_1} := \left\{p: |p - p^*_{lt}(\gamma_1)| \leq \frac{C}{10T^{1/4}}\right\}.\]
Note that $D_{\gamma_0}$ and $D_{\gamma_1}$ are disjoint since $|p^*_{lt}(\gamma_1)-p^*_{lt}(\gamma_0)|\ge C/(4T^{1/4})$.

For $\gamma\in\{\gamma_0,\gamma_1\}$, if the posted price $p_{lt}$ is not in the set $D_{\gamma}$, then the instantaneous regret is at least 
\[
r_{lt}(p^*_{lt}(\gamma),\gamma) - r_{lt}(p_{lt},\gamma) \ge \frac{c}{2} (p^*_{lt}(\gamma) - p_{lt})\ge \left(\frac{cC}{20}\right)^2\frac{1}{\sqrt{T}}\,.
\]
 Hence following a similar proof strategy as in \citep[Lemma 3.4]{broder2012dynamic},
we have
\[\text{Reg}_T^{\pi,\gamma_0} + \text{Reg}_T^{\pi,\gamma_1} \geq \left(\frac{cC}{20}\right)^2\frac{1}{\sqrt{T}}\sum_{t = 1}^T\sum_{l = 1}^L n_{lt}\left(\mathbb P_{\gamma_0}(p_{lt} \notin D_{\gamma_0}) + \mathbb P_{\gamma_1}(p_{lt} \notin D_{\gamma_1})\right)\,.\]
Note that $p_{lt}$ is measurable with respect to measure $f^{\pi,\gamma}_{t-1}$ under the model $\gamma$. Therefore, by using
a standard result on the minimum error in a simple hypothesis test~\citep[Theorem 2.2]{tsybakov2004introduction}, we have
\begin{align}
\text{Reg}_T^{\pi,\gamma_0} + \text{Reg}_T^{\pi,\gamma_1} &\geq \frac{C_1}{\sqrt{T}}\sum_{t = 1}^T\sum_{l = 1}^L n_{lt} e^{-\KL\left(f_{t-1}^{\pi, \gamma_0} ; f_{t-1}^{\pi, \gamma_1}\right) }\nonumber\\
&\ge \frac{C_1}{\sqrt{T}}N_T e^{-\KL\left(f_{T}^{\pi, \gamma_0} ; f_{T}^{\pi, \gamma_1}\right) }\,,\label{eq:KL-Bound2}
\end{align}
where in the second step we used the fact that $\KL\left(f_{t}^{\pi, \gamma_0} ; f_{t}^{\pi, \gamma_1}\right)$ is non-decreasing in $t$ and $N_T:= \sum_{t=1}^T \sum_{l=1}^L n_{lt}$. 
Previously we established the lower bound~\eqref{eq:KL-Bound1}, which reads as \[\text{Reg}_T^{\pi,\gamma_0} \geq \frac{C_2}{(\gamma_0 - \gamma)^2}\KL\left(f_t^{\pi, \gamma_0} ; f_t^{\pi, \gamma}\right)\,.\]

Putting, $\gamma = \gamma_1 = \gamma_0 + 1/4T^{1/4}$ we get $\text{Reg}_T^{\pi,\gamma_0} \geq {C_2}\sqrt{T}\KL\left(f_t^{\pi, \gamma_0} ; f_t^{\pi, \gamma_1}\right)$. Combining this bound with~\eqref{eq:KL-Bound2}, we get
\begin{align*}
\max_{\gamma\in\{\gamma_0,\gamma_1\}}\text{Reg}_T^{\pi,\gamma} &\ge \frac{1}{2} \left(\text{Reg}_T^{\pi,\gamma_0}+ \text{Reg}_T^{\pi,\gamma_1}\right) \\
&\ge C\sqrt{T} \left( \KL\left(f_t^{\pi, \gamma_0} ; f_t^{\pi, \gamma_1}\right) + \frac{N_T}{T} e^{-\KL\left(f_{T}^{\pi, \gamma_0} ; f_{T}^{\pi, \gamma_1}\right) }\right)\nonumber\\
&\ge C\sqrt{T} \left(1+\log(N_T/T)\right)\,,
\end{align*}
where in the last step we used the inequality $ae^{-b} + b \ge 1+\log(a)$.


\section{Proofs of all other results and intermediate steps}
\subsection{Proof of Proposition~\ref{prop.1}}\label{proof.prop.1}
Recall that $V^2_{lt} = \| (\bm I - \rho_t \bm W)^{-1}\bm e_l\|^2\tau^2 + \sigma^2$. Since $W\succeq \bm 0$, and $\rho_t\ge0$, we have $\bm I -\rho_t \bm W\preceq \bm I$. In addition, by Assumption~2.2, we have $\bm I - \rho_t\bm W \succeq \varepsilon \bm I$.
Therefore, since $\|\bm e_l\| = 1$, 
\[
1\le \|(I-\rho_t \bm W)^{-1}\bm e_l\| \le \frac{1}{\varepsilon}\,,
\]
from which we obtain the result.

Next, to prove the upper bound on the optimal prices, we recall that 
$$0\le \bm x'_{lt} \bm m_{lt} \le \|\bm x_{lt}\| \|\bm{m}_{lt}\|\le \frac{\|\bm \mu_t\|}{V_{lt}}\le \frac{C_{\mu}}{c_V}\,.$$

Invoking relation~\eqref{eq:16}, and noting that $\beta_t$ and so $b_t$ are negative, we arrive at
\[
p^*_{lt} = \frac{1}{-b_{lt}} \left(\varphi^{-1}(-\bm x_{lt}'\bm m_{lt}) + \bm x_{lt}' \bm m_{lt}\right) \le c_{\beta}^{-1} C_V (C_\mu c_V^{-1} -0.5 \phi(0))\,, 
\]
where we used that $\varphi^{-1}$ is increasing, $\bm x_{lt} \ge 0$, and $\varphi^{-1}(0) = -0.5/\phi(0)$.

\subsection{Proof of Lemma~\ref{lemma.1}}

By definition, $V^2_{lt} = \| (\bm I - \rho_t \bm W)^{-1}\bm e_l\|^2\tau^2 + \sigma^2$. Hence, if $\omega_*$ is the smallest eigenvalue of $\bm W$, then $V^2_{lt} \geq \tau^2/(1 - \rho_t \omega_{*})^2$.

We want to bound the $|b_{l,t+1} - b_{l,t}|$ and $\|\bm m_{l,t+1} - \bm m_{l t}\|_2$.

\begin{align*}
   \|\bm m_{l,t+1} - \bm m_{l t}\|_2 &= \left\|\frac{\bm \mu_{t+1}}{V_{l,t+1}} - \frac{\bm \mu_{t}}{V_{lt}}\right\|_2\\
   &\leq \left\|\frac{\bm \mu_{t+1} - \bm \mu_{t}}{V_{l,t+1}}\right\|_2 + \bm\mu_{t}\left\{\frac{1}{V_{l,t+1}} - \frac{1}{V_{lt}}\right\}\\
   &\leq \frac{\delta_{t\mu}}{\tau/(1 - \rho_t \omega_{*})} + C_\mu \left\{\frac{1}{V_{l,t+1}} - \frac{1}{V_{lt}}\right\}\,.
\end{align*}

Further, the second term can be simplified as 
\begin{align*}
\left\{\frac{1}{V_{l,t+1}} - \frac{1}{V_{lt}}\right\} = \frac{V_{lt} - V_{l,t+1}}{V_{l,t+1}V_{lt}} &= \frac{V_{lt}^2 - V_{l,t+1}^2}{V_{l,t+1}V_{lt}(V_{lt}+V_{l,t+1})}\nonumber\\
&\leq \frac{1}{2c_V^3} (V_{lt}^2 - V_{l,t+1}^2)\nonumber\\
&\le\frac{\tau^2}{2c_V^3} (\| (\bm I - \rho_t \bm W)^{-1}\bm e_l\|^2-\| (\bm I - \rho_{t+1} \bm W)^{-1}\bm e_l\|^2)\le C\delta_{t\rho}\,.
%
\end{align*}

The same analysis can be done for $|b_{l,t+1} - b_{lt}|$ as well.
\subsection{Proof of Corollary~\ref{cor.2}}

The corollary follows directly by applying the results from Lemma~\ref{lemma.1} in Theorem~\ref{thm.1}. Since $\rho_t = \rho$ for all $t$, hence $\delta_{t\rho} = 0$.

Since $\eta_t\propto 1/\sqrt{t}$, we get 
\begin{itemize}
    \item $\mathcal{R}_1 = LC_1C\tau^{-1}(1 - \rho \omega_*)\sum_{t = 1}^T \sqrt{t}\delta_{t\beta}$,
    \item $\mathcal{R}_2 = LC_1C\tau^{-1}(1 - \rho \omega_*)\sum_{t = 1}^T \sqrt{t}\delta_{t\mu}$,
    \item $\mathcal{R}_3 = C_3C\sum_{t = 1}^Tn_t^2/\sqrt{t} = \mathcal{O}(\sqrt{T})$,
    \item $\mathcal{R}_4 = C_4CL(C_b + C_m)\sqrt{T+1} = \mathcal{O}(\sqrt{T})$\,.
\end{itemize}

Changing the constants appropriately gives us the corollary.

\subsection{Proof of Lemma~\ref{cor.1}}\label{append.cor.1}
Consider the particular set-up when $\bm{\mu}_t=0$, $\beta_t=\beta$, $\rho_t=\rho$ and $\sigma=1$ in \eqref{eq:2}. 
Further, assume $n_{lt}=n$ for all $l, t$ and $n \to \infty$.  
The proof can easily be extended to the generic set-up. Under these parametric assumptions, first note that, $\text{Rev}(\bm{\lambda},l,t,p_{lt})=n p_{lt}\Phi(\alpha_{lt}+\beta p_{lt})$.
The optimal pricing strategy $p_{lt}^*$ maximizes $\text{Rev}(\bm{\lambda},l,t,p_{lt})$ over $p_{lt}$ for any fixed $\bm{\lambda}$. 

Now, consider an arbitrary pricing policy ${\bm{p}}$ based on the unpenalized likelihood $\text{PL}(\bm{\lambda},0)$. Such a policy will be dominated by its oracle counter-part ${\bm{p}}^{\text{or}}$ which already knows the price coefficient $\beta$ and also, knows the latent utility $U_{lt}$. Note that, the revenue of any pricing policy ${\bm{p}}$ based on the unpenalized likelihood is always dominated by the revenue of this oracle strategy, i.e., 
$\text{Rev}(\bm{\lambda},{\bm{p}})\leq \text{Rev}(\bm{\lambda},{{\bm{p}}}^{\text{or}})$. Subsequently,  the oracle strategy ${{\bm{p}}}^{\text{or}}$ will have a lower regret.    
Next, we concentrate on the regret of ${\bm{p}}^{\text{or}}$. 

For this calculation note that based on model \eqref{eq:2}, the only unknown parameters for the oracle strategy 
$\bm{p}^{\text{or}}$ are the $\alpha_{lt}$s. Under this framework consider $\alpha_{lt}$s being best estimated by $\hat{\alpha}_{lt}^{\text{or}}$. 

Now, note that as we do not have any structural assumption between $\bm{\alpha}_t$ and $\bm{\alpha}_{t+1}$ over $t=1,\ldots,T$, for any $t$,  $\hat{\alpha}_{lt}^{\text{or}}$ will be estimated based on $\{U_{ltk}:l=1,\ldots,L; k=1,\ldots,n\}$. 
As the prices $p_{lt}$ are known (based on the filtration $\mathcal{F}_{t-1}$ which contains all information upto time $t-1$) this further reduces to estimating the the $L$ means $\bm{\alpha}_{t}$ from  uncorrelated $L$ dimensional Gaussian location model where we observe $n^{-1}\sum_{k=1}^n U_{ltk}-\beta p_{lt}$ for $l=1,\ldots,L$. From the Cramer-Rao lower bound for Gaussian family, it follows that for all $l=1,\ldots,L$, we will have the following error bound on any estimate $\hat{\alpha}_{lt}$: 
$$\mathbb{E}_{\bm{\lambda}}(\hat{\alpha}_{lt}-\alpha_{lt})^2 \geq n^{-1}.$$
As such consider the $\alpha_{lt}$s under the oracle framework to be estimated by the MLE. Let $\hat{\delta}_{lt}=\hat{\alpha}_{lt}^{\text{or}}-\alpha_{lt}$. Then, noting that the MLE is asymptotically rotation invariant in this case, we have for any $\bm{\lambda}$: 
\begin{align}\label{temp.11}
\mathbb{E}_{\bm{\lambda}}\,\hat{\bm{\delta}}_{t}\hat{\bm{\delta}}_{t}^T= n^{-1} I_L~ \text{ and } \mathbb{E}_{\bm{\lambda}}\,\hat{\bm{\delta}}_{t} \to \bm{0} \text{ as } n \to \infty.
\end{align}
Now, note that for the oracle strategy, 
$$\text{Rev}(\bm{\lambda},l,t,p_{lt}^{\text{or}})= \max_{p\geq 0} n p\, \Phi(\hat{\alpha}_{lt}^{\text{or}}+\beta p)= \max_{p\geq 0} n p\, \Phi(\hat{\delta}_{lt} +{\alpha}_{lt}+\beta p)~.$$
Let $f(l,t,p)= n p\, \Phi(\hat{\delta}_{lt} +{\alpha}_{lt}+\beta p)$. Consider Taylor-Series expansion: 
$$ f(l,t,p)= n p\, \Phi({\alpha}_{lt}+\beta p) + n \,\hat{\delta}_{lt}\, p\, \phi({\alpha}_{lt}+\beta p) + 2^{-1} n p\, \hat{\delta}_{lt}^2 \phi'({\alpha}_{lt}+\beta p)+r(l,t,p),$$
where $r(l,t,p)$ contains third and higher order terms. Now, we have ${L}^{-1}\sum_l f(l,t,p_{lt})$ converges in probability to 
$$\frac{1}{L}\sum_l n p_{lt}\, \Phi({\alpha}_{lt}+\beta p_{lt}) + \frac{1}{L}\sum_l n \, p_{lt} \phi({\alpha}_{lt}+\beta p_{lt}) \mathbb{E}_{\bm{\lambda}} \hat{\delta}_{lt} + \frac{1}{2L}\sum_l   n p_{lt}  \phi'({\alpha}_{lt}+\beta p_{lt}) \mathbb{E}_{\bm{\lambda}} \hat{\delta}_{lt}^2,$$
as $L^{-1}\sum_l r(l,t,p_{lt}) \to 0$ in probability as $n\, \mathbb{E}_{\bm{\lambda}} \hat{\delta}_{lt}^{2+m}=O(n^{-m/2})$, for $m\geq 1$. Using \eqref{temp.11}, the second term in the above expression vanishes and the third term gets further simplified, resulting in the following asymptotic result: 
$$\frac{1}{L}\sum_l f(l,t,p_{lt})=n \bigg[\frac{1}{L}\sum_l  p_{lt}\, \Phi({\alpha}_{lt}+\beta p_{lt})\bigg]
+ \frac{1}{2L}\sum_l   p_{lt}  \phi'({\alpha}_{lt}+\beta p_{lt})+{o}(1)~.$$
Thus, the regret of ${\bm{p}}^{\text{or}}$ at time $t$ is given by
 \begin{align}\label{temp12}
 L^{-1}\sum_{l=1}^L\mathcal{R}_{lt}(\bm{\lambda},{\bm{p}}^{\text{or}})\geq (\mathcal{A} -\mathcal{B})/L +o(1),
 \end{align}
where, 
\begin{align*}
\mathcal{A}&=\max_{p_{lt}: l=1,\ldots,L} \bigg[\sum_l n p_{lt}\, \Phi({\alpha}_{lt}+\beta p_{lt})\bigg], \text{ and } \\
\mathcal{B}&=\max_{p_{lt}: l=1,\ldots,L} \bigg[ \sum_l n p_{lt}\, \Phi({\alpha}_{lt}+\beta p_{lt})
- 2^{-1}\sum_l   p_{lt} ({\alpha}_{lt}+\beta p_{lt}) \phi({\alpha}_{lt}+\beta p_{lt})\bigg].
\end{align*}
Note that, the expression in $\mathcal{B}$ is simplified using $\phi'(u) = -u\phi(u)$. Now recall that $\beta$, being the price sensitivity, is negative. 
Based on model \eqref{eq:2}, for the utilities to be positive we have the following assumption of the price: ${\alpha}_{lt}+\beta p_{lt}>0$ for all $l$ and $t$. Let the prices be selected such that $\inf_{l} {\alpha}_{lt}+\beta p_{lt}>\epsilon_0$ for some prefixed small $\epsilon_0 >0$. By Proposition~\ref{prop.1}, the optimal prices are bounded and so are $\sup_l \alpha_{lt}+\beta p_{lt}<M_0$.  Then, 
$$\mathcal{A}-\mathcal{B}\geq 2^{-1} \epsilon \sum_l p_{lt}^*,$$
where $p_{lt}^*$ is the optimal price based on criterion $\mathcal{B}$, and $\epsilon = \min_{\epsilon_0<|u|<M'} u\phi(u)$. 
Thus, the cumulative regret of ${\bm{p}}^{\text{or}}$ over time is given by 
 \begin{align*}
 \mathcal{R}(\bm{\lambda},{\bm{p}}^{\text{or}})=\sum_{t=1}^T\sum_{l=1}^L\mathcal{R}_{lt}(\bm{\lambda},{\bm{p}}^{\text{or}})= \Omega(LT).
 \end{align*}
 Thus, we have, 
 \begin{align*}
 \mathcal{B}(\bm{\theta},{\bm{p}}_U)= \Omega(LT). 
 \end{align*}
Now, consider the regret from the proposed strategy. Based on \eqref{eq:23.1}, we have
\begin{align*}
   \mathcal{B}(\bm{\theta},\bm{p})\leq  &{C}_6 {\tau}^{-1}({1 - \rho_{*}\omega_*})\sum_{t = 1}^T\sqrt{t}({\delta_{t\beta} + \delta_{t\mu}})  + C_7 \sum_{t = 1}^T \sqrt{t}\delta_{t\rho} + \mathcal{O}(\sqrt{T})=\mathcal{O}(\sqrt{T}),
\end{align*}
where, the second asymptotic result follows as 
$\sum_{t = 1}^T \sqrt{t}\delta_{t\beta}$, $\sum_{t = 1}^T \sqrt{t}\delta_{t\mu}$ and
$\sum_{t = 1}^T \sqrt{t}\delta_{t\rho}$ are all bounded above by $\mathcal{O}(\sqrt{T})$. Comparing the above two displays the result follows. 
\subsection{Proof of Proposition~\ref{prop.2}}

    Consider the revenue function given by~\eqref{eq:Rev-mod}, $\text{Rev}_{lt}(p) = n_{lt}p\Phi(b_{lt}p + \bm x_{lt}'\bm m_{lt})$. By definition, $p_{lt}^*$ is the maximizer of $\text{Rev}_{lt}(p)$ and hence $\text{Rev}_{lt}'(p_{lt}^*) = 0$. Using Taylor series expansion around $p_{lt}^*$, we get
    \[\mathcal{R}_{lt} = \text{Rev}_{lt}(p_{lt}^*) - \text{Rev}_{lt}(p_{lt}) = \frac{1}{2}\text{Rev}''_{lt}(p)(p_{lt} - p_{lt}^*)^2,\]
    for some $p$ between $p_{lt}$ and $p_{lt}^*$. The second derivative can be bounded as 
    
     \[\frac{1}{2}\text{Rev}_{lt}''(p) = n_{lt} \frac{2b_{lt} \phi(b_{lt}p + \bm x_{lt}'\bm m_{lt}) + p b_{lt}^2 \phi'(b_{lt}p + \bm x_{lt}'\bm m_{lt})}{2} \leq \left(C_b \phi(0) + M C_b^2 \frac{\phi(0)}{\sqrt{2}}\right) n_{lt}.\]

Setting $C_9 =  C_b \phi(0) + M C_b^2 \frac{\phi(0)}{\sqrt{2}}$ completes the proof.


\subsection{Proof of Lemma~\ref{lemma.2}}

We have the true parameters $b_{lt}, \bm{m}_{lt}$ and the output $\hat{b}_{lt}, \hat{\bm{m}}_{lt}$ from our PSGD pricing policy. Also, $p_{lt}$ and $p_{lt}^*$ are the price based on our policy and the optimal price based on the true parameters, respectively. 

As discussed in section~\ref{sec.2}, we can write the prices in terms of the utility model parameters using the function $g(\cdot,\cdot)$, as follows: 
\begin{align*}
    p_{lt}^* := g(b_{lt}, \bm m_{lt}) =  -\frac{\varphi^{-1}(-\bm{x}_{lt}' {\bm m}_{lt}) + \bm{x}_{lt}' {\bm m}_{lt}}{b_{lt}} ~,\\
    p_{lt} := g(\hat{b}_{lt}, \hat{\bm m}_{lt}) =  -\frac{\varphi^{-1}(-\bm{x}_{lt}' \hat{\bm m}_{lt}) + \bm{x}_{lt}' \hat{\bm m}_{lt}}{\hat{b}_{lt}} ~.
 \end{align*}
    Now, note that,
    \begin{align*}
        &(p_{lt} - p_{lt}^*)^2 = \left\{{(\varphi^{-1}(-\bm{x}_{lt}' {\bm m}_{lt}) + \bm{x}_{lt}' {\bm m}_{lt}})\, {b_{lt}}^{-1}-{(\varphi^{-1}(-\bm{x}_{lt}' \hat{\bm m}_{lt}) + \bm{x}_{lt}' \hat{\bm m}_{lt})}\,{\hat{b}_{lt}}^{-1}\right\}^2=\{A+B\}^2,
        \end{align*}
        where, the right side above is decomposed as
        \begin{align*}
        &A=(\varphi^{-1}(-\bm{x}_{lt}' {\bm m}_{lt}) + \bm{x}_{lt}' {\bm m}_{lt})\,{b_{lt}}^{-1} -(\varphi^{-1}(-\bm{x}_{lt}' \hat{\bm m}_{lt}) + \bm{x}_{lt}' \hat{\bm m}_{lt})\,{b_{lt}}^{-1}, \text{ and, }\\
        &B=\left(\varphi^{-1}(-\bm{x}_{lt}' \hat{\bm m}_{lt}) + \bm{x}_{lt}' \hat{\bm m}_{lt}\right)\,(1/{b_{lt}} - 1/{\hat{b}_{lt}}).
    \end{align*}
    Using the naive bound $\{A+B\}^2\leq 2(A^2+B^2)$ first and then the bound $|b_{lt}|= |\beta_t|/V_{lt}\ge c_\beta/C_V$ and the policy rule $p_{lt} = g(\hat{b}_{lt}, \hat{\bm m}_{lt})$, it follows that $(p_{lt} - p_{lt}^*)^2$ is bounded above by 
    \begin{align*}
        {2 C_V^2}{c_\beta^{-2}}\left\{({\varphi^{-1}(-\bm{x}_{lt}' {\bm m}_{lt}) + \bm{x}_{lt}' {\bm m}_{lt}}) - ({\varphi^{-1}(-\bm{x}_{lt}' \hat{\bm m}_{lt}) + \bm{x}_{lt}' \hat{\bm m}_{lt}})\right\}^2 + 2p_{lt}^2 {b_{lt}}^{-2} ({b_{lt}-\hat{b}_{lt}})^2\,,
    \end{align*}
    Since, $\varphi^{-1}(-v)+v$ is 1-Lipschitz, we have 
  \[\left(({\varphi^{-1}(-\bm{x}_{lt}' {\bm m}_{lt}) + \bm{x}_{lt}' {\bm m}_{lt}}) - ({\varphi^{-1}(-\bm{x}_{lt}' \hat{\bm m}_{lt}) + \bm{x}_{lt}' \hat{\bm m}_{lt})}\right)^2  \leq \inner{\bm x_{lt}} {\bm m_{lt} - \hat{\bm m}_{lt}}^2\,. \]

Therefore, we have
\begin{align*}
(p_{lt} - p_{lt}^*)^2 \le 
{2C_V^2} c_\beta^{-2} \inner{\bm x_{lt}}{\bm m_{lt}-\hat{\bm{m}}_{lt}}^2 
+{2C_V^2}c_\beta^{-2}p_{lt}^2(b_{lt}-\hat{b}_{lt})^2\,.
\end{align*}


Setting $C_{10} = {2C_V^2}{c_\beta^{-2}}$ proves the lemma.

\end{document}